%% file: main.tex
\documentclass[journal]{IEEEtran}
\usepackage{cite}
\usepackage{amsmath,amssymb,amsfonts}
\usepackage{algorithmic}
\usepackage{graphicx}
\usepackage{textcomp}
\def\BibTeX{{\rm B\kern-.05em{\sc i\kern-.025em b}\kern-.08em
    T\kern-.1667em\lower.7ex\hbox{E}\kern-.125emX}}

\usepackage{textcomp}
\usepackage{caption}
\usepackage{subcaption}

\usepackage{pgfplots}
\usepackage{pgfplotstable}
\pgfplotsset{compat=1.7}
\usepackage{tikz}

\usepackage{array}
\usepackage{multirow, makecell}

\usepackage{booktabs}
\usepackage{tabularx}

\usepackage[bookmarks=false]{hyperref}

\usepackage[style=base]{caption}
\usepackage{subcaption}

\usepackage[figuresright]{rotating}
\usepackage[nolist,nohyperlinks]{acronym}
\usepackage[ancient]{flushend}

\usepackage{pifont}
\newcommand{\cmark}{\ding{51}}%
\usetikzlibrary{shapes.geometric, arrows, calc}
\tikzstyle{mainprocess} = [rectangle, minimum width=7.2cm, minimum height=1.23cm, text centered, draw=black, fill=white!96!black]
\tikzstyle{subprocess} = [rectangle, minimum width=4.2cm, minimum height=1.23cm, text centered, draw=black, fill=white!96!black]
\tikzstyle{subsubprocess} = [rectangle, minimum width=3cm, minimum height=1.23cm, text centered, draw=black, fill=white!96!black]
\tikzstyle{arrow} = [dashed,->,>=stealth]

\begin{document}

\title{Collaborative Multi-Robot Systems for Search and Rescue: Coordination and Perception}

\author{

    \IEEEauthorblockN{
        Jorge Pe\~{n}a Queralta\textsuperscript{1},
        Jussi Taipalmaa\textsuperscript{2},
        Bilge Can Pullinen\textsuperscript{2},
        Victor Kathan Sarker\textsuperscript{1},
        Tuan Nguyen Gia\textsuperscript{1},\\
        Hannu Tenhunen\textsuperscript{1},
        Moncef Gabbouj\textsuperscript{2},
        Jenni Raitoharju\textsuperscript{2},
        Tomi Westerlund\textsuperscript{1}
    }\\[6pt]
    \IEEEauthorblockA{%
        \textsuperscript{1}\href{https://tiers.utu.fi}{Turku Intelligent Embedded and Robotic Systems, University of Turku, Finland} \\
        Email: \textsuperscript{1}\{jopequ, vikasar, tunggi, tovewe\}@utu.fi\\[2pt]
        \textsuperscript{2}Department of Computing Sciences, Tampere University, Finland\\[-1pt]
        Email: \textsuperscript{2}\{jussi.taipalmaa, bilge.canpullinen, moncef.gabbouj, jenni.raitoharju\}@tuni.fi\\
    }
    \vspace{-1em}
}


%
%
%

\maketitle
\input{sections/00_Abstract}
\IEEEpeerreviewmaketitle

\input{sections/01_Introduction}

\input{sections/01_Projects}
\input{sections/02_System}

\input{sections/03_Coordination_Planning}   
\input{sections/04_Perception}

\input{sections/05_ActivePerception}

\input{sections/06_Discussion}

\input{sections/07_Conclusion}


\section*{Acknowledgment}
This research work is supported by the Academy of Finland's AutoSOS project (Grant No. 328755).

\bibliographystyle{IEEEtran}
\bibliography{reference.bib}




\end{document}

%% file: sections/00_Abstract.tex
\begin{abstract}

Autonomous or teleoperated robots have been playing increasingly important roles in civil applications in recent years. Across the different civil domains where robots can support human operators, one of the areas where they can have more impact is in search and rescue (SAR) operations. In particular, multi-robot systems have the potential to significantly improve the efficiency of SAR personnel with faster search of victims, initial assessment and mapping of the environment, real-time monitoring and surveillance of SAR operations, or establishing emergency communication networks, among other possibilities. SAR operations encompass a wide variety of environments and situations, and therefore heterogeneous and collaborative multi-robot systems can provide the most advantages. In this paper, we review and analyze the existing approaches to multi-robot SAR support, from an algorithmic perspective and putting an emphasis on the methods enabling collaboration among the robots as well as advanced perception through machine vision and multi-agent active perception. Furthermore, we put these algorithms in the context of the different challenges and constraints that various types of robots (ground, aerial, surface or underwater) encounter in different SAR environments (maritime, urban, wilderness or other post-disaster scenarios). This is, to the best of our knowledge, the first review considering heterogeneous SAR robots across different environments, while giving two complimentary points of view: control mechanisms and machine perception. Based on our review of the state-of-the-art, we discuss the main open research questions, and outline our insights on the current approaches that have potential to improve the real-world performance of multi-robot SAR systems.

\end{abstract}

\begin{IEEEkeywords}
    Robotics, search and rescue (SAR), multi-robot systems (MRS), machine learning (ML), active perception, active vision, multi-agent perception, autonomous robots.
\end{IEEEkeywords}


%% file: sections/01_Introduction.tex
\begin{figure}[ht!]
    \centering
    \begin{subfigure}[t]{0.46\textwidth}
        \includegraphics[width=\textwidth]{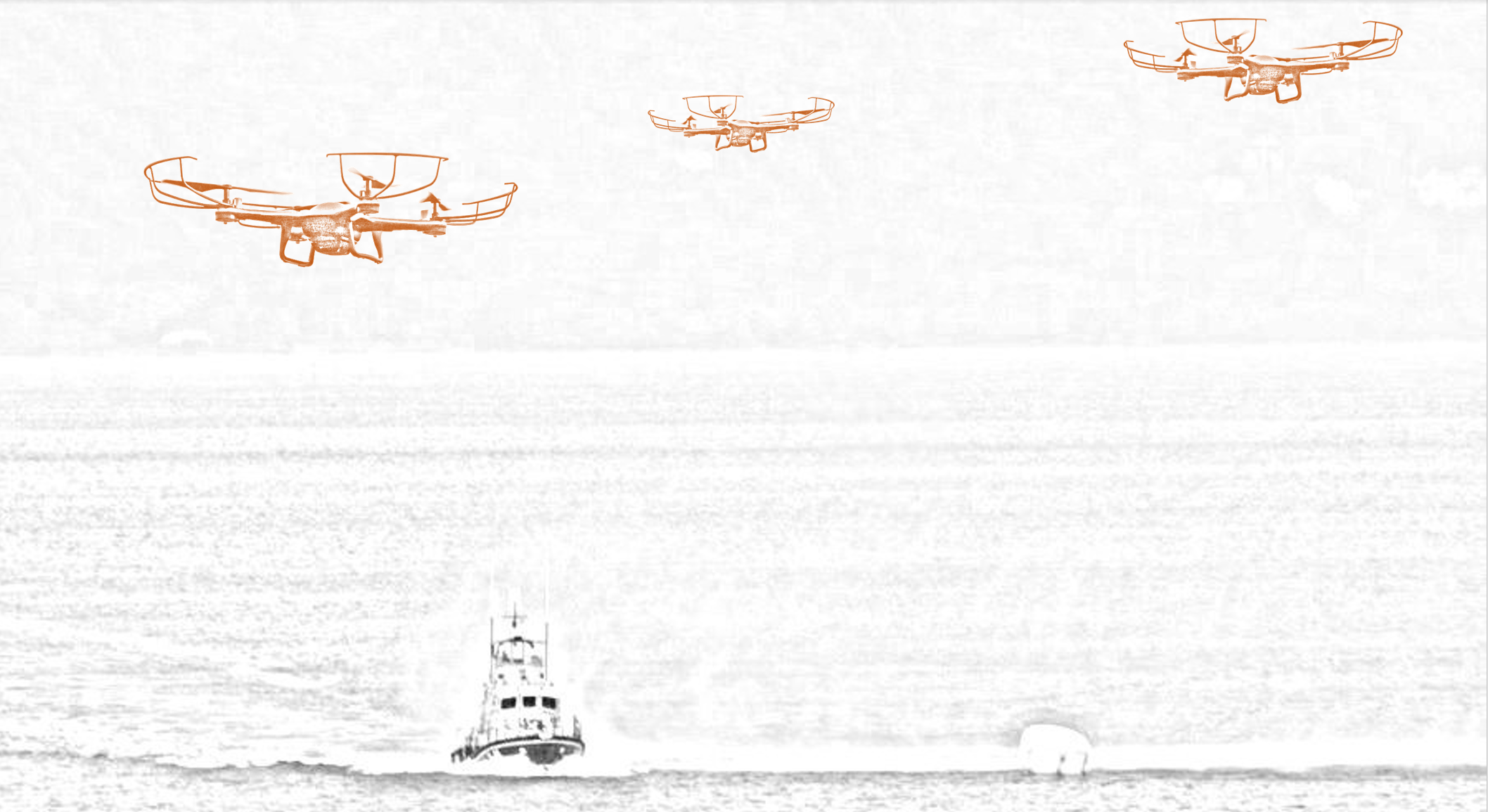}
        \caption{Maritime search and rescue with UAVs and USVs}
        \label{subfig:scenario_maritime}
        \vspace{1em}
    \end{subfigure}
    \begin{subfigure}[t]{0.46\textwidth}
        \includegraphics[width=\textwidth]{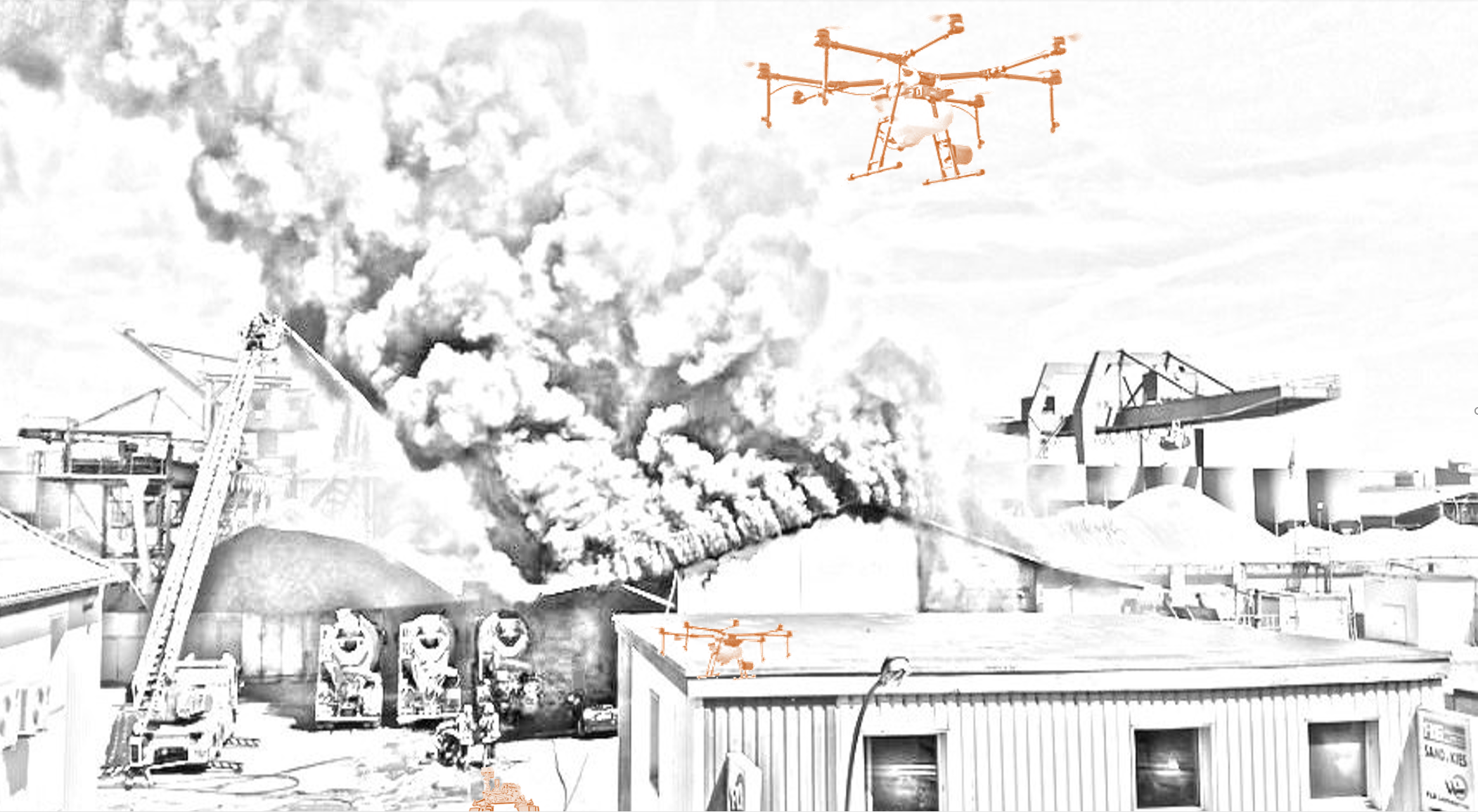}
        \caption{Urban search and rescue with UAVs and UGVs}
        \label{subfig:scenario_fire}
        \vspace{1em}
    \end{subfigure}
    \begin{subfigure}[t]{0.46\textwidth}
        \includegraphics[width=\textwidth]{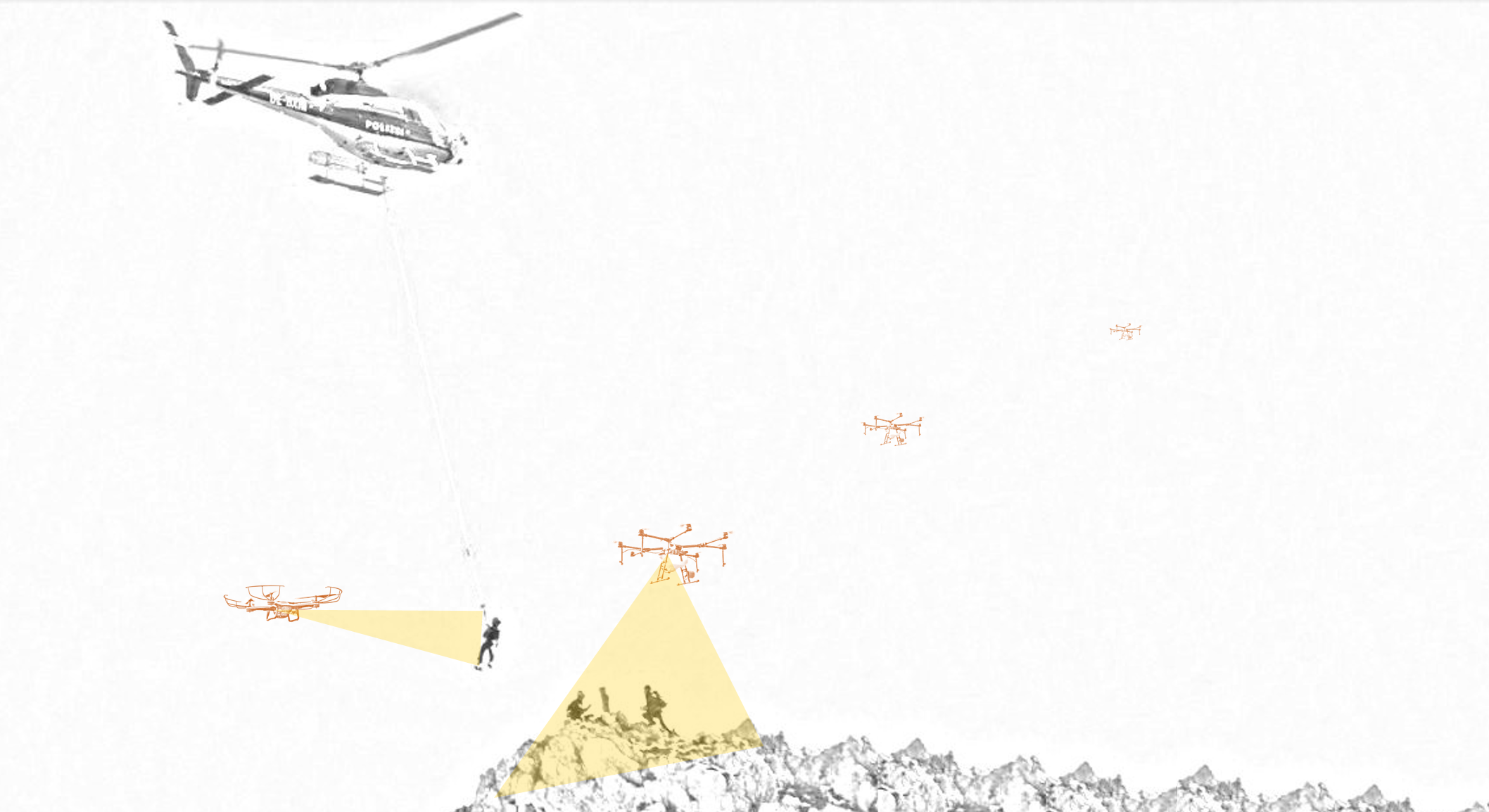}
        \caption{Wilderness search and rescue with support UAVs}
        \label{subfig:scenario_mountain}
    \end{subfigure}
    \caption{Different search and rescue scenarios where heterogeneous multi-robot systems can assist SAR taskforces.}
    \vspace{-2em}
    \label{fig:scenarios}
\end{figure}

\begin{figure*}
    \centering
    \begin{subfigure}[t]{0.99\textwidth}
        \small{\input{fig/structure_system}}
        \caption{Aspects of multi-robot SAR systems discussed in Section~\ref{sec:system} of this paper.}
        \label{subfig:structure_system}
    \end{subfigure}
    \begin{subfigure}[t]{0.99\textwidth}
        \vspace{2em}
        \small{\input{fig/structure_algorithms}}
        \caption{Division of multi-robot SAR systems into separate components from an algorithmic point of view. Control, planning and coordination algorithms are described in Section~\ref{sec:coverage}, while Section~\ref{sec:vision} reviews perception algorithms from a machine learning perspective. Section~\ref{sec:active_perception} then puts these two views together by reviewing the works in single and multi-agent active perception. }
        \label{subfig:structure_algorithms}
        \vspace{1em}
    \end{subfigure}
    \caption{Summary of the different aspects of multi-robot SAR systems considered in this survey, where we have separated (a) system-level perspective, and (b) planning and perception algorithmic perspective.}
    \label{fig:SAR_system}
\end{figure*}
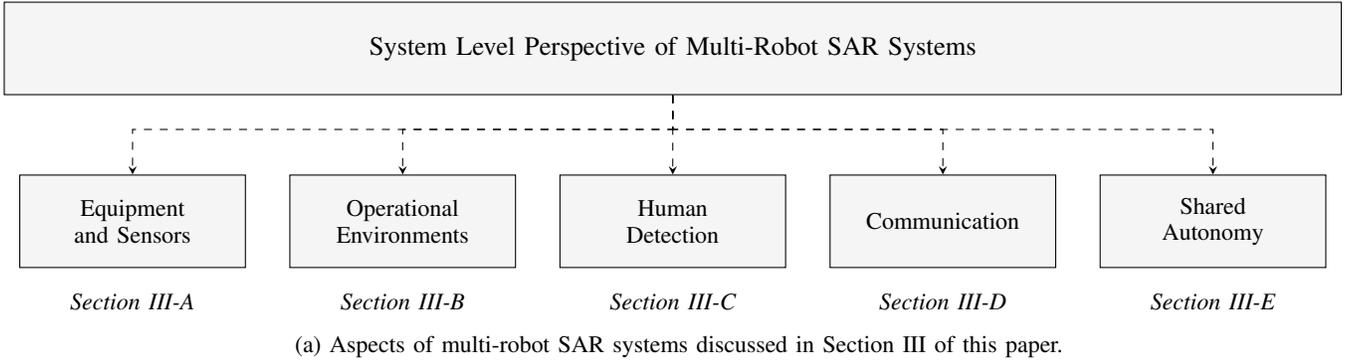
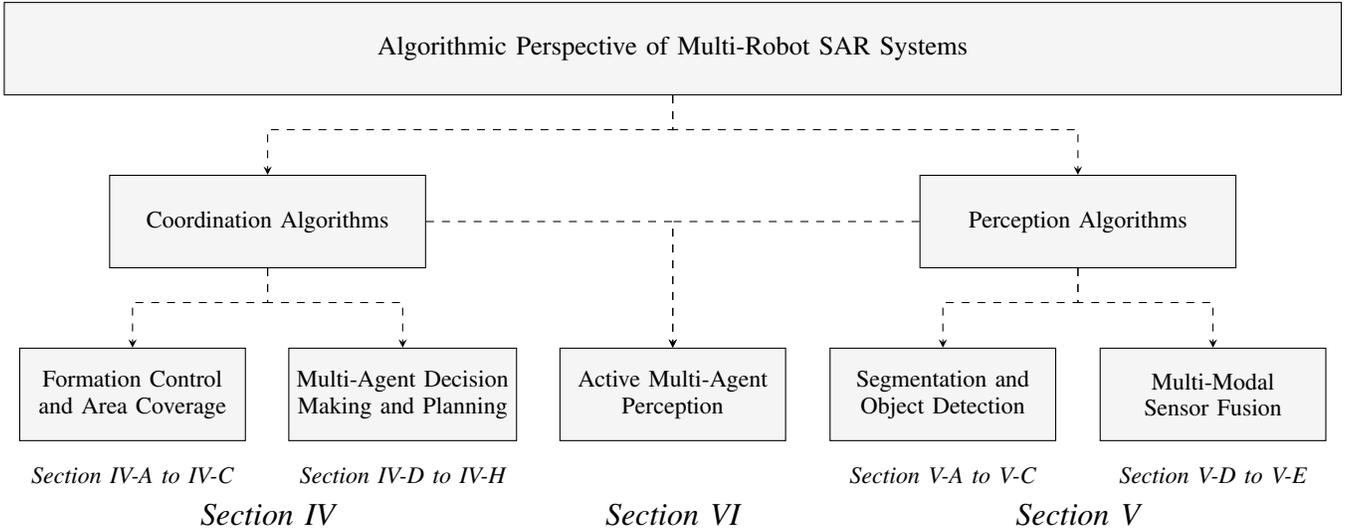

\section{Introduction}\label{sec:intro}

Autonomous robots have seen an increasing penetration across multiple domains in the last decade. In industrial environments, collaborative robots are being utilized in the manufacturing sector, and fleets of mobile robots are swarming in logistics warehouses. Nonetheless, their utilization within civil applications presents additional challenges owing to the interaction with humans and their deployment in potentially unknown environments~\cite{INGRAND201710, deng2018survey, shakhatreh2019unmanned}. Among civil applications, search and rescue (SAR) operations present a key scenario where autonomous robots have the potential to save lives by enabling faster response time~\cite{mehmood2018multi, roberts2016unmanned}, supporting in hazardous environments~\cite{luk2005intelligent, lunghi2019multimodal, sung2019multi}, or providing real-time mapping and monitoring of the area where an incident has occurred~\cite{merino2005cooperative, brenner2017new}, among other possibilities. In this paper, we perform a literature review of multi-robot systems for SAR scenarios. These systems involve SAR operations in a variety of environments, some of which are illustrated in Fig.~\ref{fig:scenarios}. With the wide variability of SAR scenarios, different situations require robots to be able to operate in different environments. In this document, we utilize the following standard notation to refer to the different types of robots: unmanned aerial vehicles (UAVs), unmanned ground vehicles (UGVs), unmanned surface vehicles (USVs), and unmanmed underwater vehicles (UUVs). These can be either autonomous or teleoperated, and very often a combination of both modalities exists when considering heterogeneous multi-robot systems. In maritime SAR, autonomous UAVs and USVs can support in finding victims (Fig.~\ref{subfig:scenario_maritime}). In urban scenarios, UAVs can provide real-time information for assessing the situation and UGVs can access hazardous areas (Fig.~\ref{subfig:scenario_fire}). In mountain scenarios, UAVs can help in monitoring and getting closer to the victims that are later rescued by a helicopter (Fig.~\ref{subfig:scenario_mountain}). 

In recent years, multiple survey papers addressing the utilization of multi-UAV systems for civil applications have been published. In~\cite{hayat2016survey}, the authors perform an exhaustive review of multi-UAV systems from the point of view of communication, and for a wide range of applications from construction or delivery to SAR missions. An extensive classification of previous works is done taking into account the mission and network requirements in terms of data type, frequency, throughput and quality of service (latency and reliability). In comparison to~\cite{hayat2016survey}, our focus is on multi-robot systems including also ground, surface, or underwater robots. Another recent review related to civil applications for UAVs was carried out in~\cite{shakhatreh2019unmanned}.In~\cite{shakhatreh2019unmanned}, the authors provide a classification in terms of technological trends and algorithm modalities utilized in research papers: collision avoidance, mmWave communication and radars, cloud-based offloading, machine learning, image processing and software-defined networking, among others. 

A recent survey~\cite{grogan2018use} focused on UAVs for SAR operations, with an extensive classification of research papers based on (i) sensors utilized onboard the UAVs, (ii) robot systems (single or multi-robot systems, and operational mediums), and (iii) environment where the system is meant to be deployed. In a study from Grayson et al.~\cite{grayson2014search}, the focus is on using multi-robot systems for SAR operations, with an emphasis on task allocation algorithms, communication modalities, and human-robot interaction for both homogeneous and heterogeneous multi-robot systems. In this work, we review also heterogeneous multi-robot systems. However, rather than focusing on describing the existing solutions at a system level, we put an emphasis on the algorithms that are being used for multi-robot coordination and perception. Moreover, we describe the role of machine learning in single and multi-agent perception, and discuss how active perception can play a key role towards the development of more intelligent robots supporting SAR operations. The survey is further divided into three main sub-categories: 1) planning and area coverage algorithms, 2) machine perception, and 3) active perception algorithms combining the previous two concepts (Fig. \ref{fig:SAR_system}). 

While autonomous robots are being increasingly adopted for SAR missions, current levels of autonomy and safety of robotic systems only allow for full autonomy in the \textit{search} part, but not for \textit{rescue}, where human operators need to intervene~\cite{queralta2020autosos}. This leads to the design of shared autonomy interfaces and research in human-swarm interaction, which will be discussed in Section~\ref{sec:system}. In general, the literature on multi-robot SAR operations with some degree of autonomy is rather sparse, with most results being based on simulations or simplified scenarios~\cite{nazarova2020application}. At the same time, the literature on both robots for SAR (autonomous or teleoperated) and multi-robot coordination and perception is too vast to be reviewed here in a comprehensive manner. Therefore, we review the most significant multi-robot coordination and perception algorithms that can be applied to autonomous or semi-autonomous SAR operations, while also providing an overview of existing technologies in use in SAR robotics that can be naturally extended to multi-robot systems. In the areas where the literature allows, we compare the different solutions involving individual robots or multi-robot systems.

In summary, in this survey we provide an overview and analysis of multi-robot SAR from an algorithmic point of view, exploring various degrees of autonomy. We focus on studying the different methods for efficient multi-robot collaboration and control and real-time machine learning models for multi-modal sensor fusion. From the point of view of collaboration, we categorize previous works based on decision-making modalities (e.g., centralized or distributed, or requiring local or global interactions) for role and task assignment. In particular, the main types of tasks that we review are collaborative search and multi-robot area coverage, being the most relevant to SAR operations. From the machine learning perspective, we review novel multi-modal sensor fusion algorithms and we put a focus on active vision techniques. In this direction, we review the current trends in single- and multi-agent perception, mainly for object detection and tracking, and segmentation algorithms. Our objective is therefore to characterize the literature from two different yet complementary points of view: single- and multi-robot control and coordination, on one side, and machine learning for single- and multi-agent perception, on the other. This is, to the best of our knowledge, the first survey to cover these two aspects simultaneously, as well as describing SAR operations with heterogeneous multi-robot systems. In summary, our contribution focuses on reviewing the different aspects of multi-robot SAR operations with
\begin{enumerate}
    \item a system-level perspective for designing autonomous SAR robots considering the operational environment, communication, level of autonomy, and the interaction with human operators,
    \item an algorithmic point of view of multi-robot coordination, multi-robot search and area coverage, and distributed task allocation and planning applied to SAR operations, 
    \item a deep learning viewpoint to single- and multi-agent perception, with a focus on object detection and tracking and segmentation, and a description of the challenges and opportunities of active perception in multi-robot systems for SAR scenarios.
\end{enumerate}

The remainder of this paper is organized as follows: Section~\ref{sec:projects} describes some of the most relevant projects in SAR robotics, with an emphasis on those considering multi-robot systems. In Section~\ref{sec:system}, we present a system view on SAR robotic systems, describing the different types of robots being utilized, particularities of SAR environments, and different aspects for multi-robot SAR including communication and shared autonomy. Section~\ref{sec:coverage} follows with the description of the main algorithms in multi-agent planning and multi-robot coordination that can be applied to SAR scenarios, with an emphasis on area coverage algorithms. In Section~\ref{sec:vision}, we focus on machine vision and multi-agent perception from a deep learning perspective. Then, Section~\ref{sec:active_perception} goes through the concept of active vision and delves into the integration of both coordination and planning algorithms with robotic vision towards active perception algorithms where the latter provides additional feedback to the control loops of the former. In Section~\ref{sec:discussion}, we discuss open research questions in the field of autonomous heterogeneous multi-robot systems for SAR operations, outlining the main aspects that current research efforts are being directed to. Finally, Section~\ref{sec:conclusion} concludes this work.

%% file: fig/structure_system.tex
\begin{tikzpicture}[node distance=2.3cm]

    %
    %
    \node (start) [mainprocess, minimum width=0.99\textwidth] {\normalsize{System Level Perspective of Multi-Robot SAR Systems}};
    
    \node (projt) [subsubprocess, below of=start, xshift=-0.4\textwidth, align=center] {Equipment \\ and Sensors};
    \node (openv) [subsubprocess, below of=start, xshift=-0.2\textwidth, align=center] {Operational \\ Environments};
    \node (human) [subsubprocess, below of=start, xshift=+0\textwidth, align=center] {Human \\ Detection};
    \node (objct) [subsubprocess, below of=start, xshift=+0.2\textwidth, align=center] {Communication};
    \node (shaut) [subsubprocess, below of=start, xshift=+0.4\textwidth, align=center] {Shared\\Autonomy};
    
    %
    %
    \draw [arrow] (start) |- ($(projt.north) + (0,6mm)$) -| (projt);
    \draw [arrow] (start) |- ($(openv.north) + (0,6mm)$) -| (openv);
    \draw [arrow] (start) |- ($(human.north) + (0,6mm)$) -| (human);
    \draw [arrow] (start) |- ($(objct.north) + (0,6mm)$) -| (objct);
    \draw [arrow] (start) |- ($(shaut.north) + (0,6mm)$) -| (shaut);
    
    %
    %
    \node (sec_cov) [below of=projt, yshift=1.23cm] {\small{\textit{Section~\ref{subsect:system}}}};
    \node (sec_cov) [below of=openv, yshift=1.23cm] {\small{\textit{Section~\ref{subsect:environment}}}};
    \node (sec_cov) [below of=human, yshift=1.23cm] {\small{\textit{Section~\ref{subsect:human}}}};
    \node (sec_cov) [below of=objct, yshift=1.23cm] {\small{\textit{Section~\ref{subsect:shared}}}};
    \node (sec_cov) [below of=shaut, yshift=1.23cm] {\small{\textit{Section~\ref{subsect:communication}}}};
    
\end{tikzpicture}

%% file: fig/structure_algorithms.tex
\begin{tikzpicture}[node distance=2.3cm]

    %
    %
    \node (start) [mainprocess, minimum width=0.99\textwidth] {\normalsize{Algorithmic Perspective of Multi-Robot SAR Systems}};
    
    \node (coord) [subprocess, below of=start, xshift=-0.3\textwidth] {Coordination Algorithms};
    \node (perct) [subprocess, below of=start, xshift=+0.3\textwidth] {Perception Algorithms};
    
    \node (formt) [subsubprocess, below of=coord, xshift=-0.1\textwidth, align=center] {Formation Control \\ and Area Coverage};
    \node (roall) [subsubprocess, below of=coord, xshift=+0.1\textwidth, align=center] {Multi-Agent Decision \\ Making and Planning};
    \node (objct) [subsubprocess, below of=perct, xshift=-0.1\textwidth, align=center] {Segmentation and \\ Object Detection};
    \node (segmt) [subsubprocess, below of=perct, xshift=+0.1\textwidth, align=center] {Multi-Modal \\ Sensor Fusion};
    
    \node (activ) [subsubprocess, below of=start, yshift=-2.3cm, align=center] {Active Multi-Agent \\ Perception};

    %
    %
    \draw [arrow] (start) |- ($(coord.north) + (0,6mm)$) -| (coord);
    \draw [arrow] (start) |- ($(perct.north) + (0,6mm)$) -| (perct);
    
    \draw [arrow] (coord) |- ($(formt.north) + (0,6mm)$) -| (formt);
    \draw [arrow] (coord) |- ($(roall.north) + (0,6mm)$) -| (roall);
    
    \draw [arrow] (perct) |- ($(objct.north) + (0,6mm)$) -| (objct);
    \draw [arrow] (perct) |- ($(segmt.north) + (0,6mm)$) -| (segmt);
    
    \draw [arrow] (coord) -| (activ);
    \draw [arrow] (perct) -| (activ);
    
    %
    %
    \node (ch4a) [below of=formt, yshift=+1.23cm] {\small{\textit{Section~\ref{subsect:task} to~\ref{subsect:coverage}}}};
    \node (ch4b) [below of=roall, yshift=+1.23cm] {\small{\textit{Section~\ref{subsect:single_planning} to~\ref{subsect:planning_heterogeneous}}}};
    
    \node (ch5a) [below of=objct, yshift=+1.23cm] {\small{\textit{Section~\ref{subsec:segmentation} to~\ref{subsec:light}}}};
    \node (ch5b) [below of=segmt, yshift=+1.23cm] {\small{\textit{Section~\ref{subsec:multimodal} to~\ref{subsec:multi_perception}}}};
    
    \node (sec_cov) [below of=coord, yshift=-1.6cm] {\large{\textit{Section~\ref{sec:coverage}}}};
    \node (sec_vis) [below of=perct, yshift=-1.6cm] {\large{\textit{Section~\ref{sec:vision}}}};
    \node (sec_act) [below of=activ, yshift=+0.7cm] {\large{\textit{Section~\ref{sec:active_perception}}}};
    
\end{tikzpicture}

%% file: sections/01_Projects.tex
\begin{sidewaystable*}
    \centering
    \caption{\footnotesize{Selection of international projects and competitions in SAR robotics. We describe the utilization of different types of robots (UAV, USV, UGV), whether heterogeneous robots are employed, where the data is processed, and the characterization of networking and control strategies. The latter two aspects are only classified from a topological point of view in this table: centralized/predefined versus mesh/ad-hoc networks, and centralized versus distributed control. The application scenarios refer to either the specific objective of the project, or the scenarios utilized for testing. In the competitions section, each parameter defines the possibilities but not necessarily the characterization for all systems participating the challenges.}}
    \label{tab:projects}
    \footnotesize
    \renewcommand{\arraystretch}{1.42}
    \begin{tabular}{@{}p{1em}p{0.05\textwidth}p{0.16\textwidth}cccccccccc@{}}
        \toprule
        & & \multirow{2}{*}{Description} & Multi-robot & Auton. & Auton. & Auton. & Multi-UAV & Heterogeneous & Sensor Data & Ad-Hoc & Distributed & \multirow{2}{*}{Scenario} \\
        & &  & system & UGV & USV & UAV & system & robots & Processing & Network & control \\
        \midrule
        \parbox[t]{2mm}{\multirow{18}{*}{\rotatebox[origin=c]{90}{International Projects in SAR Robotics}}} & \multirow{2}{*}{\textbf{COMETS}} & \scriptsize{Real-time coordination and control of multiple heterogeneous UAVs.} & \multirow{2}{*}{\cmark} & \multirow{2}{*}{\cmark} & \multirow{2}{*}{-} & \multirow{2}{*}{\cmark} & \multirow{2}{*}{\cmark} & \multirow{2}{*}{\shortstack{3xUAVs\\Heli+Airship}} & \multirow{2}{*}{Offboard} & \multirow{2}{*}{-} & \multirow{2}{*}{-} & \multirow{2}{*}{Forest fire} \\
        & \textbf{Guardians} & \scriptsize{Swarm of autonomous robots applied to navigate and search an urban ground.} & \multirow{2}{*}{\cmark} & \multirow{2}{*}{\cmark} & \multirow{2}{*}{-} & \multirow{2}{*}{-} & \multirow{2}{*}{-} & \multirow{2}{*}{-} & \multirow{2}{*}{Onboard} & \multirow{2}{*}{\cmark} & \multirow{2}{*}{\cmark} & Firefighting \\
        & \textbf{ICARUS} & \scriptsize{Development of robotic tools which can assist human SAR operators.} & \multirow{2}{*}{\cmark} & \multirow{2}{*}{\cmark} & \multirow{2}{*}{\cmark} & \multirow{2}{*}{\cmark} & \multirow{2}{*}{\cmark} & \multirow{2}{*}{\shortstack{UGV+UAVs\\+USV+UUV}} & \multirow{2}{*}{Offboard} & \multirow{2}{*}{-} & \multirow{2}{*}{-} & \multirow{2}{*}{\shortstack{Practical \\ SAR integration}} \\
        & \textbf{NIFTi} & \scriptsize{Natural human-robot cooperation in dynamic environments for urban SAR.} & \multirow{2}{*}{\cmark} & \multirow{2}{*}{\cmark} & \multirow{2}{*}{-} & \multirow{2}{*}{\cmark} & \multirow{2}{*}{-} & \multirow{2}{*}{\cmark} & \multirow{2}{*}{-} & \multirow{2}{*}{-} & \multirow{2}{*}{-} & \multirow{2}{*}{\shortstack{Urban \\ disasters}} \\
        & \textbf{Darius} & \scriptsize{Integrated unmanned systems for urban, forest fires and maritime SAR.} & \multirow{2}{*}{\cmark} & \multirow{2}{*}{\cmark} & \multirow{2}{*}{\cmark} & \multirow{2}{*}{\cmark} & \multirow{2}{*}{\cmark} & \multirow{2}{*}{\shortstack{UAVs+UGVs\\+USV+UUV}} & Offboard & \multirow{2}{*}{\cmark} & - & \multirow{2}{*}{\shortstack{Forest, urban \\ and maritime}} \\
        & \textbf{SEAGULL} & UAVs to support maritime situational awareness. & \multirow{2}{*}{\cmark} & \multirow{2}{*}{-} & \multirow{2}{*}{-} & \cmark & \cmark & \multirow{2}{*}{-} & \multirow{2}{*}{Offboard} & \multirow{2}{*}{-} & \multirow{2}{*}{-} & \multirow{2}{*}{Maritime} \\
        & \textbf{TRADR} & \scriptsize{Long-term human-robot teaming for response in industrial accidents.} & \multirow{2}{*}{\cmark} & \multirow{2}{*}{\cmark} & \multirow{2}{*}{-} & \multirow{2}{*}{\cmark} & \multirow{2}{*}{\cmark} & \multirow{2}{*}{\cmark} & \multirow{2}{*}{-} & \multirow{2}{*}{-} & \multirow{2}{*}{-} & \multirow{2}{*}{Industrial} \\
        & \textbf{SmokeBot} & \scriptsize{Robots with environmental sensors for disaster sites with low visibility.} & \multirow{2}{*}{-} & \multirow{2}{*}{\cmark} & \multirow{2}{*}{-} & \multirow{2}{*}{\cmark} & \multirow{2}{*}{-} & \multirow{2}{*}{-} & \multirow{2}{*}{\cmark} & \multirow{2}{*}{-} & \multirow{2}{*}{-} & \multirow{2}{*}{\shortstack{Fires and \\ low visibility}} \\
        & \textbf{Centauro} & \scriptsize{Mobility and dexterous manipulation in SAR by full-body telepresence.} & \multirow{2}{*}{\cmark} & \multirow{2}{*}{-} & \multirow{2}{*}{-} & \multirow{2}{*}{-} & \multirow{2}{*}{-} & \multirow{2}{*}{-} & \multirow{2}{*}{-} & \multirow{2}{*}{-} & \multirow{2}{*}{-} & \multirow{2}{*}{\shortstack{Harsh \\ environments}} \\
        & \textbf{MEXT DDT} & \scriptsize{Aero, on-rubble/underground, and in-rubble robots for urban earthquakes.} & \multirow{2}{*}{\cmark} & \multirow{2}{*}{\cmark} & \multirow{2}{*}{-} & \multirow{2}{*}{\cmark} & \multirow{2}{*}{\cmark} & \multirow{2}{*}{UAVs+UGVs} & \multirow{2}{*}{\cmark} & \multirow{2}{*}{\cmark} & \multirow{2}{*}{-} & \multirow{2}{*}{Earthquake} \\
        & \textbf{AutoSOS} & \scriptsize{Multi-UAV system supporting maritime SAR with lightweight AI at the edge.} & \multirow{2}{*}{\cmark} & \multirow{2}{*}{-} & \multirow{2}{*}{\cmark} & \multirow{2}{*}{\cmark} & \multirow{2}{*}{\cmark} & \multirow{2}{*}{USV+UAVs} & \multirow{2}{*}{\cmark} & \multirow{2}{*}{\cmark} & \multirow{2}{*}{\cmark} & \multirow{2}{*}{Maritime} \\
        \midrule
        \parbox[t]{2mm}{\multirow{13}{*}{\rotatebox[origin=c]{90}{Competitions}}} & \textbf{Robocup\newline Rescue} & \scriptsize{Challenges involved in SAR applications and promoting research collaboration.} & \multirow{3}{*}{\cmark} & \multirow{3}{*}{\cmark} & \multirow{3}{*}{\cmark} & \multirow{3}{*}{\cmark} & \multirow{3}{*}{\cmark} & \multirow{3}{*}{\cmark} & \multirow{3}{*}{\shortstack{Onboard+\\offboard\\allowed}} & \multirow{3}{*}{\cmark} & \multirow{3}{*}{\cmark} & \multirow{3}{*}{\shortstack{\\[-6pt]Multiple\\environments}} \\
        & \textbf{ImPACT-TRC} & \scriptsize{Tough Robotics Challenge for technologies to aid in disaster response, recovery and preparedness.} & \multirow{3}{*}{\cmark} & \multirow{3}{*}{\cmark} & \multirow{3}{*}{-} & \multirow{3}{*}{\cmark} & \multirow{3}{*}{-} & \multirow{3}{*}{UAVs+UGVs} & \multirow{3}{*}{-} & \multirow{3}{*}{-} & \multirow{3}{*}{-} & \multirow{3}{*}{\shortstack{Earthquakes,\\nuclear disasters,\\tsunamis}} \\ 
        & \textbf{ERL-ESR} & \scriptsize{European Robotics League (ERL) Emergency Service Robots} & \multirow{3}{*}{\cmark} & \multirow{3}{*}{\cmark} & \multirow{3}{*}{\cmark} & \multirow{3}{*}{\cmark} & \multirow{3}{*}{\cmark} & \multirow{3}{*}{\cmark} & \multirow{3}{*}{\cmark} & \multirow{3}{*}{\cmark} & \multirow{3}{*}{\cmark} & \multirow{3}{*}{\shortstack{\\[-6pt]Urban\\multi-domain}} \\
        & \textbf{RESCON} & \scriptsize{Rescue Robot Contest for large-scale urban disasters} & \multirow{3}{*}{-} & \multirow{3}{*}{\cmark} & \multirow{3}{*}{-} & \multirow{3}{*}{-} & \multirow{3}{*}{-} & \multirow{3}{*}{-} & \multirow{3}{*}{\cmark} & \multirow{3}{*}{-} & \multirow{3}{*}{-} & \multirow{3}{*}{\shortstack{\\[-6pt]Earthquake\\body recovery}} \\
        & \textbf{DARPA} & \scriptsize{Robotics Challenge in human-supervised robotic technology for disaster-response operations.} & \multirow{3}{*}{-} & \multirow{3}{*}{\cmark} & \multirow{3}{*}{-} & \multirow{3}{*}{-} & \multirow{3}{*}{-} & \multirow{3}{*}{-} & \multirow{3}{*}{\cmark} & \multirow{3}{*}{-} & \multirow{3}{*}{-} & \multirow{3}{*}{Urban SAR} \\
        & \textbf{OnShape} & \scriptsize{Robots to the Rescue student design challenge. Teleoperated robots for disaster response.} & \multirow{3}{*}{-} & \multirow{3}{*}{\cmark} & \multirow{3}{*}{-} & \multirow{3}{*}{-} & \multirow{3}{*}{-} & \multirow{3}{*}{-} & \multirow{3}{*}{-} & \multirow{3}{*}{-} & \multirow{3}{*}{-} & \multirow{3}{*}{Simulation} \\
        \bottomrule
    \end{tabular}
\end{sidewaystable*}

\section{International Projects and Competitions}
\label{sec:projects}

Over the past two decades, multiple international projects have been devoted to SAR robotics, often with the aim of working towards multi-robot solutions and the development of multi-modal sensor fusion algorithms. In this section, we review the most relevant international projects and international competitions in SAR robotics, which are listed in Table~\ref{tab:projects}. Some of the projects focus in the development of complex robotic systems that can be remotely controlled~\cite{klamt2019flexible}. However, the majority of the projects consider multi-robot systems~\cite{ollero2005multiple, kruijff2014designing, de2018persistent, cubber2017search}, and over half of the projects consider collaborative heterogeneous robots. In Table~\ref{tab:projects}, we have described these projects from a system-level point of view, without considering the degree of autonomy or the control and perception algorithms. These latter two aspects are described in Sections~\ref{sec:system} through~\ref{sec:active_perception}, where not only these projects but also other relevant works are put into a more appropriate context.

An early approach to the design and development of heterogeneous multi-UAV systems for cooperative activities was presented within the COMETS project (real-time coordination and control of multiple heterogeneous unmanned aerial vehicles)~\cite{ollero2005multiple}. More recently, other international projects designing and developing autonomous multi-robot systems for SAR operations include the NIFTi EU project (natural human-robot cooperation in dynamic environments)~\cite{kruijff2014designing}, ICARUS (unmanned SAR)~\cite{matos2016multiple, cubber2017search}, TRADR (long-term human-robot teaming for disaster response)~\cite{de2018persistent, gawel2018x, freda20193d}, or SmokeBot (mobile robots with novel environmental sensors for inspection of disaster sites with low visibility)~\cite{fritsche2016radar, wei2016multi}. Other projects, such as CENTAURO (robust mobility and dexterous manipulation in disaster response by fullbody telepresence in a centaur-like robot), have focused on the development of more advanced robots that are not fully autonomous but controlled in real-time~\cite{klamt2019flexible}.

In COMETS, the aim of the project was to design and implement a distributed control system for cooperative activities using heterogeneous UAVs. To that end, the project researchers developed a remote-controlled airship and an autonomous helicopter and worked towards cooperative perception in real-time~\cite{ollero2005multiple, gancet2005task, merino2005cooperative}. In NIFTi, both UGVs and UAVs were utilized for autonomous navigation and mapping in harsh environments~\cite{kruijff2014designing}. The focus of the project was mostly on human-robot interaction and on distributing information for human operators at different layers. Similarly, in the TRADR project, the focus was on collaborative efforts towards disaster response of both humans and robots~\cite{de2018persistent}, as well as on multi-robot planning~\cite{gawel2018x, freda20193d}. In particular, the results of TRARD include a framework for the integration of UAVs in SAR missions, from path planning to a global 3D point cloud generator~\cite{surmann2019integration}. The project continued with the foundation of the German Rescue Robotics Center at Fraunhofer FKIE, where broader research is conduced, for example, in maritime SAR~\cite{guldenring2019heterogeneous}. In ICARUS, project researchers developed an unmanned maritime capsule acting as a UUV, USVs, a large UGV, and a group of UAVs for rapid deployment, as well as mapping tools, middleware software for tactical communications, and a multi-domain robot command and control station~\cite{cubber2017search}. While these projects focused on the algorithmic aspects of SAR operation, and on the design of multi-robot systems, in Smokebot the focus was on developing sensors and sensor fusion methods for harsh environments~\cite{fritsche2016radar, wei2016multi}. A more detailed description of some of these projects, specially those that started before 2017, is available in~\cite{cubber2017introduction}.

In terms of international competition and tournaments, a relevant precedent in autonomous SAR operations is the European Robotics League Emergency Tournament. In~\cite{winfield2016eurathlon}, the authors describe the details of what was the world's first multi-domain (air, land and sea) multi-robot SAR competition. A total of 16 international teams competed with tasks including (i) environment reconnaissance and mapping (merging ground and aerial data), (ii) search for missing workers outside and inside an old building, and (iii) pipe inspection with localization of leaks (on land and underwater). This and other competitions in the field are listed in Table~\ref{tab:projects}.

%% file: sections/02_System.tex
\section{Multi-Robot SAR: System-Level Perspective}\label{sec:system}

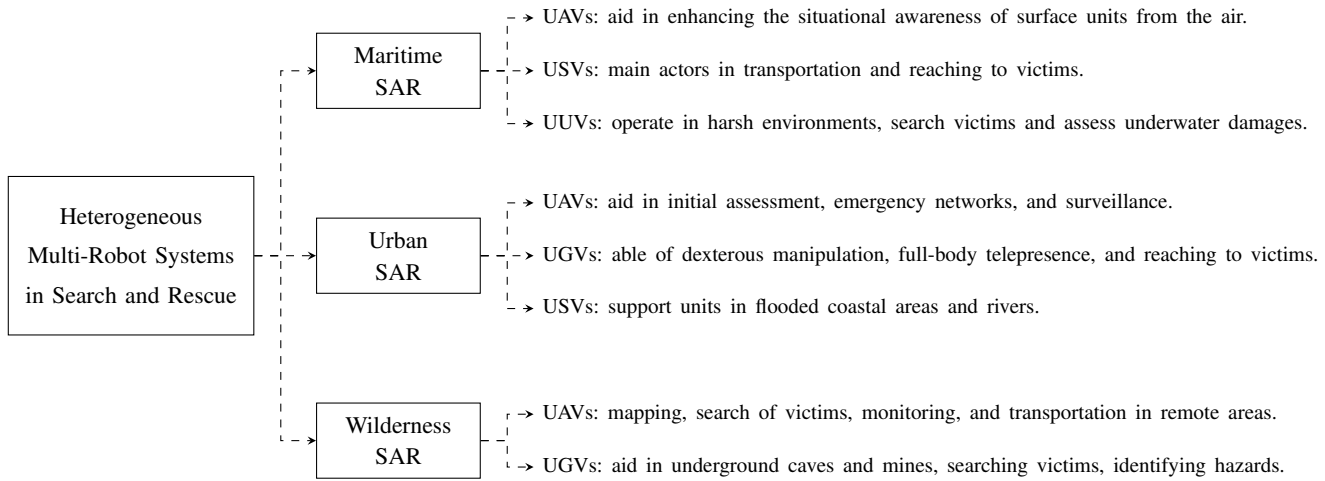
\begin{figure*}
    \centering
    \input{fig/heterogeneous}
    \caption{Types of autonomous robots utilized in different SAR scenarios.}
    \label{fig:vehicles}
\end{figure*}

Robotic SAR systems can differ in multiple ways: their environment (urban, maritime, wilderness), the amount and type of robots involved (USVs, UAVs, UGVs, UUVs), their level of autonomy, and the ways in which humans control the robotic systems, among other factors. Compared to robots utilized in other situations, SAR robots often need to be deployed in more hazardous environmental conditions, or in remote areas limiting communication and sensing. Through this section, we describe the main aspects involved in the design and deployment of SAR robots, with a focus, wherever the literature is extensive enough, in multi-robot SAR. The main subsections are listed in Fig.~\ref{subfig:structure_system}. We describe these different aspects of robotic SAR systems and the challenges and opportunities in terms of coordination, control, and perception. This section therefore includes a discussion on multiple system-level topics: from how different environments affect the way robotic perception is designed, to how inter-robot communication systems and human-robot interfaces affect multi-robot coordination and multi-agent perception algorithms.

\subsection{System Requirements and Equipment Used}
\label{subsect:system}

Owing to the wide variety of situations and natural disasters requiring SAR operations, and taking into account the extreme and challenging environments in which robots often need to operate during SAR missions, most of the existing literature has focused towards describing the design, development and deployment of specialized robots. Here we give a global overview of the main equipment and sensors utilized in ground, maritime and aerial SAR robots.

%
%
Two complimentary examples of ground robots for SAR operations are introduced in~\cite{berns2017unmanned}, where both large and small robots are described. Ground robots for SAR missions can be characterized among those with dexterous manipulation capabilities and robust mobility on uneven terrain, such as the robot developed within the CENTAURO project~\cite{klamt2019flexible}, smaller robots with the ability of moving through tight spaces~\cite{berns2017unmanned}, or serpentine-like robots able of tethered operation across complex environments~\cite{konyo2019impact}. Some typical sensors in ground robots for SAR operations, as described in~\cite{berns2017unmanned}, are inertial and GNSS sensors, RGB-D cameras, infrared and thermal cameras, laser scanners, gas discrimination sensors, and microphones and speakers to offer a communication channel between SAR personnel and victims.

%
%
In terms of aerial robots, a representative and recent example is available in~\cite{surmann2019integration}, where the authors introduce a platform for instantaneous UAV-based 3D mapping during SAR missions. The platform offers a complete sensor suite. The main sensors are a 16-channel laser scanner, an infrared camera for thermal measurements, an RGB camera, and inertial/positional sensors for GNSS and altitude estimation. The UAV, a DJI S1000+ octocopter, is connected to a ground station on-board a fire fighter command vehicle with a custom radio link capable of over 300\,Mbps downlink speed at distances up to 300\,m. The system is able to produce point clouds colored both by reflectance (from the laser measurements) and temperature (from the infrared camera). This suite of sensors is one of the most complete for UAVs, except for the lack of ultrasonic sensors. In general, however, cameras are the predominant sensors owing to their flexibility, size and weight. Examples of autonomous quadrotors, fixed-wing and rotatory-wing vehicles equipped with GNSS sensors and RGB cameras for search of people in emergency scenarios are available in~\cite{goodrich2008supporting, qi2016search, sun2016camera}. A description of different types of aerial SAR robots utilized within the ICARIUS project is available in~\cite{konrad2017unmanned}, and a recent survey on UAVs for SAR operations by Grogan et al. shows the predominance of RGB cameras as the main or only sensor in use, without considering inertial and GNSS units~\cite{grogan2018use}.  

%
%
Maritime SAR operations often involve both surface and underwater robots as well as support UAVs. Descriptions of different surface robots offering an overview of existing solutions are available in~\cite{jorge2019survey} and~\cite{matos2017unmanned}. Some particularities of maritime SAR robots include the use of seafloor pressure sensors, seismometers, and hydrophone for the detection of tsunamis and earthquakes, or sensors for measuring meteorological variables as well as water conditions (e.g., temperature, salinity, depth, pH balance and concentrations of different chemicals). Other examples include sensors for various liquids and substances for robots utilized in oil spills or contaminated waters (e.g., laser fluorosensors).

%
%
A significant challenge in SAR robotics, owing to the specialization of robots in specific tasks, is interoperability. The ICARUS and DARIUS projects have both worked towards the integration of different unmanned vehicles or robots for SAR operations~\cite{serrano2015icarus, cubber2017introduction}. Interoperability is particularly important in heterogeneous multi-robot systems, where data from different sources needs to be aggregated in real-time for efficient operation and fast actuation. Furthermore, because robots in SAR operations are mostly supervised or partly teleoperated, the design of a ground station is an essential piece in a complete SAR robotics system. This is even more critical when involving the control of multi-robot systems. The design of a generic ground station able to accommodate a wide variety of unmanned vehicles has been one of the focuses of the DARIUS project~\cite{chrobocinski2012darius}. The approach to interoperability taken within the ICARIUS project is described in detail in~\cite{lopez2017interoperability}. The project outcomes included a library for multi-robot cooperation in SAR missions that assumes that the Robot Operating System (ROS) is the middleware utilized across all robots involved in the mission. ROS is the de facto standard in robotics industry and research~\cite{quigley2009ros}. In~\cite{lopez2017interoperability}, the authors also characterize typical robot roles, levels of autonomy for different types of robots, levels of interoperability, and robot capabilities.

\subsection{Operational Environment}
\label{subsect:environment}

In this subsection, we characterize the main SAR environments (urban, maritime and wilderness) and discuss how the different challenges in each of these types of scenario have been addressed in the literature.

\begin{table*}
    \centering
    \caption{Challenges and Opportunities for Autonomous Robots in different types of environments: Urban SAR~\cite{davids2002urban, shah2004survey, liu2013robotic, ganz2015urban, chen2017robust, surmann2019integration, konyo2019impact, lewis2019developing}, Maritime SAR~\cite{murphy2012marine, mendoncca2016cooperative, matos2016multiple, zhao2019search, queralta2020autosos}, and Wilderness SAR~\cite{silvagni2017multipurpose, bryant2019autonomous, chikwanha2012survey, zhao2017search}.}
    \label{tab:environments}
    \renewcommand{\arraystretch}{2}
    \begin{tabular}{@{}p{0.07\textwidth}p{0.5\textwidth}p{0.37\textwidth}@{}}
        \toprule
         & \textbf{Challenges} & \textbf{Opportunities}\\
        \midrule
        \textbf{Maritime} \newline \textbf{SAR} & (i) Visual detection of people at sea, with potentially vast areas to search and comparatively small targets to detect.  \newline (ii) The need for long-distance operation, with either high levels of autonomy or real-time communication in remote environments. \newline  (iii) Underwater robots often rely on tethered communication or need to resurface to share their findings. \newline (iv) Localization and mapping underwater presents significant challenges owing to the transmission characteristics in water of light and other electromagnetic waves used in more traditional sensing methods. \newline (v) Motion affected by marine currents, waves and limited water depths. \newline & (i) UAVs can provide a significant improvement at sea in term of situational awareness from the air, and can be deployed on-site even from small ships. \newline (ii) Heterogeneous multi-robot systems can aid in multi-modal coordinated search aggregating information from the different perspectives (aerial, surface, underwater). \newline (iii) Disposable or well-adapted USVs and UUVs can be utilized in harsh environments or bad weather conditions when SAR operations at sea are interrupted for safety reasons. \\
        \textbf{Urban} \newline \textbf{SAR} & (i) The presence of hazardous materials, radiation areas, or high temperatures. \newline (ii) Localization and mapping of unknown, unstructured, dense and hazardous environments that result from disasters such as earthquakes or explosions, and in which robots are meant to operate. \newline (iii) Navigation in narrow spaces and uneven terrain, being able to traverse small apertures and navigate over unstable debris.\newline (iv) Close cooperation with human operators in a potentially shared operation space, requiring for well defined human-robot interaction models. & (i) Relieving human personnel from emotional stress and physical threats (e.g., radiation, debris). \newline (ii) Reducing the time for locating survivors. Mortality in USAR scenarios raises significantly after 48\,h. \newline (iii) Assessing the structural parameters of the site and assisting on remote or semi-autonomous triage. \newline (iv) Detecting and locating survivors and analyzing the surrounding structures. \newline (v) Establishing a communication link to survivors. \newline \\
        \textbf{Wilderness} \newline \textbf{SAR} & (i) In avalanche events, robots often need to access remote areas (long-term operation) while in harsh weather conditions (e.g., low temperatures, low air pressure, high wind speeds). \newline (ii) Exploration of underground mines and caves presents significant challenges from the point of view of long-term localization and communication.  \newline  (iii) SAR operations to find people lost while hiking or climbing mountains often occur in the evening or at night, when visibility conditions make it more challenging for UAVs or other robots to identify objects and people. \newline (iv) WiSAR operations often involve tracking of a moving target, with a search area that expands through time. & (i) After an avalanche, areas that are hard to reach by land can be quickly surveyed with UAVs. \newline (ii) SAR personnel in mines or caves can rely on robots for environmental monitoring, mainly toxic gases, and avoid hazardous areas. \newline (iii) UAVs equipped with thermal cameras can aid in the search of lost hikers or climbers at night, as well as relay communication from SAR personnel. \newline (iv) Multi-robot systems can build probabilistic maps for movable targets and revisit locations more optimally. \newline \\
        \bottomrule
    \end{tabular}
\end{table*}

%
%
\paragraph*{Maritime SAR}

Search and rescue operations at sea were characterized by Zhao et al. in~\cite{zhao2019search}. The authors differentiated five stages in sea SAR operations: (i) distress alert received, (ii) organizing and planning, (iii) maritime search, (iv) maritime rescue, and (v) rapid evacuation. Influential factors affecting planning and design of operations in each of these stages are also provided. Some of the most significant factors influence the planning across all stages. These include injury condition, possession of location devices and rescue equipment, and environmental factors such as geographic position, wave height, water temperature, wind speed and visibility. The paper also emphasizes that maritime accidents tend to happen suddenly. In particular, a considerable amount of accidents happen near the shoreline with favorable weather conditions, such as beaches during the summer. In these areas, robotic SAR systems can be ready to act fast. For instance, Xian et al. designed a life-ring drone delivery system for rescuing people caught in a rip current near the shore, showing a 39\% reduction in response time compared to the time lifeguards need to reach their victims in a beach~\cite{xiangRipRescue2016}.

From these and other works, we have summarized the main challenges for autonomous robots in marine SAR environments, and listed them in Table~\ref{tab:environments}. The most important from the point of view of robotic design and planning are the limited sensing and communication ranges (both on the surface and underwater), together with the difficulties of operating in maritime environments. Winds, marine currents and waves complicate positioning and navigation of robots, limiting their controlability.

The main types of autonomous robots utilized in maritime SAR operations are USVs and UUVs~\cite{matos2016multiple}, but also support UAVs~\cite{yeong2015review}. Owing to the important advantages in terms of situational awareness that these different robots enable together, sea SAR operations are one of the scenarios where heterogeneous multi-robot systems have been already widely adopted~\cite{yeong2015review}. A representative work on the area, showing a heterogeneous and cooperative multi-robot system for SAR operations after ship accidents, was presented by Mendo\c{c}a et al.~\cite{mendoncca2016cooperative}. The authors proposed the utilization of both a USV and UAV to find shipwreck survivors at sea, where the USV would carry the UAV until it arrives near the shipwreck location. By utilizing these two types of robots, the system is able to leverage the longer range, larger payload capability, and extended operational time of the USV with the better visibility and situational awareness that the UAV provides on-site. 

The combination of USVs and UUVs has also been widely studied, with or without UAVs. Some of the most prominent examples in this direction come from the euRathlon competition and include solutions from the ICARUS project~\cite{matos2017unmanned}. The surface robot was first utilized to perform an autonomous assessment, mapping and survey of the area, identifying points of interest. Then, the underwater vehicle was deployed and the path planning automatically adjusted to the points of interest previously identified. In particular, the task of the UUV was to detect pipe leaks and find victims underwater.

%
%
\paragraph*{Urban SAR}

Early works on urban SAR (USAR) are included in a survey by Shah et al. describing the most important benefits of incorporating robots in USAR operations~\cite{shah2004survey}, and a paper by Davids describing the state of the field at  the start of the twenty-first century~\cite{davids2002urban}. In both cases, part of the motivation for research in the area had a direct source at the World Trade Center disaster in New York City. More recently, Liu et al. have presented a survey of USAR robotics from the perspective of designing robotic controllers~\cite{liu2013robotic}.

Urban SAR scenarios include (i) natural disasters such as earthquakes or hurricanes, (ii) large fires, (iii) hazardous accidents (e.g., gas explosions or traffic accidents involving trucks transporting hazardous chemicals), (iv) airplane crashes in urban areas, or any other type of event resulting in trapped or missing people, or collapsed buildings and inaccessible areas. The main benefits of involving autonomous robots in USAR scenarios are clear, including first and foremost to increase the safety of rescue personnel by reducing their exposure to potential hazards in the site and providing an initial assessment of the situation. From some of the most significant works in the area~\cite{davids2002urban, shah2004survey, liu2013robotic, ganz2015urban, chen2017robust, surmann2019integration, konyo2019impact, lewis2019developing}, we have summarized the main opportunities and challenges in the development of robotic systems to support USAR operations. The main challenges for autonomous robots in USAR operations can be encapsulated in the points listed in Table~\ref{tab:environments}, together with the potential opportunities that the literature in the topic describes.

In recent years, multiple research efforts have been put towards solving some of the aforementioned challenges. In terms of navigation for autonomous robots, Mittal et al. presented a novel method for UAV navigation and landing in urban post-disaster scenarios, where existing maps have been rendered unusable and different hazard factors must be taken into account~\cite{mittal2019vision}. In a similar direction, Chen et al. developed a robust SLAM method fusing monocular visual and lidar data, for a rescue robot able to climb stairs and operate in uneven terrain~\cite{chen2017robust}. With the significant advances that UAVs have seen over the past two decades and the increasingly fast adoption of UAVs in civil applications, commercial off-the-shelf UAVs have already the potential to support SAR operations. In~\cite{tomic2012toward}, a research platform aiming at fully autonomous SAR UAVs was defined. In~\cite{surmann2019integration}, the authors describe a heterogeneous multi-UAV system focused at providing an initial assessment of the environment through mapping, object detection and annotation, and scene classifier. 

Novel types of robotic systems have also been developed to better adapt to the challenges of USAR environments. For instance, taking inspiration from video scopes and fiber scopes utilized to obtain imagery from confined spaces, robots that extend this concept have been developed~\cite{hatazaki2007active, namari2012tube}. In~\cite{konyo2019impact}, researchers participating in the ImPACT-TRC challenge presented a thin serpentine robot platform, a long and flexible continuum robot with a length of up to 10\,m and a diameter of just 50\,mm, able to localize itself with visual SLAM techniques and access collapsed buildings. Other environments posing significant challenges to the deployment of autonomous robots are urban fires. To be able to utilize UAVs near fires, Myeong et al. presented FAROS, a fireproof drone for USAR operations~\cite{myeongFireproofDrone2017}.

%
%
\paragraph*{Wilderness SAR}

In wilderness SAR (WiSAR) operations, the literature often includes SAR in mountains~\cite{silvagni2017multipurpose}, underground mines and caves~\cite{chikwanha2012survey, zhao2017search, ranjan2019wireless}, and forests and other rural or remote environments~\cite{goodrich2008supporting, macwanMultirobot2015, wangVortexSearch2018}. While mostly describing individual robots or homogeneous multi-robot systems, the potential of heterogeneous robotic systems for WiSAR operations has been shown in recent years~\cite{kashino2019aerial}.

One of the most common SAR operations in mountain environments occurs in a post-avalanche scenario. In areas with a risk of avalanches, mountaineers often carry avalanche transmitters (AT), a robust technology that has been in use for decades. UAVs prepared for harsh conditions (strong winds, high altitude and low temperatures) have been utilized for searching ATs~\cite{silvagni2017multipurpose}. In~\cite{bryant2019autonomous}, an autonomous multi-UAV system for localizing avalanche victims was developed. The authors identify two key benefits of using multiple agents: robustness through redundancy and minimization of search time. In~\cite{kinugasa2016validation}, the authors study the potential of using ATs in urban scenarios for SAR missions with autonomous robots, where the victims are assumed to have ATs with them before an accident happens. 

Forest environments also present significant challenges from the perception point of view, due to the density of the environments and lack of structure for path planning. Forest or rural trails can be utilized in this regard but are not always available~\cite{zhilenkovForest2018}. Moreover, WiSAR operations in forests and rural areas often involve tracking a moving target (a lost person). This implies that the search area increases through time, and path planning algorithms need to adapt accordingly. This issue is addressed in~\cite{macwanMultirobot2015} for a multi-robot system, with an initial search path that is monitored and re-optimized in real-time if the search conditions need to be modified.

Another specific scenario that has attracted research attention is SAR for mining applications~\cite{chikwanha2012survey}. Two specific challenges in SAR operations in underground environments are the limitations of wireless communication and the existence of potentially toxic gases. Regarding the former one, Ranjan et al. have presented an overview of wireless robotic communication networks for underground mines~\cite{ranjan2019wireless}. To support SAR personnel in the latter challenge, Zhao et al. presented a SAR multi-robot system for remotely sensing environmental variables, including the concentration of known gases, in underground coal mine environments~\cite{zhao2017search}. More recently, Fan et al. integrated gas discrimination with underground mapping for emergency response, while not restricted to underground environments~\cite{fan2019towards}. Robots able to map the concentration of different gases in an emergency response scenario can aid firefighters and other personnel in both detecting hazardous areas and understanding the development of the situation. 

In summary, robots for WiSAR need to be equipped to operate in rough meteorological conditions, but also to deal with some specific parameters, such as lower air pressure hindering UAV flight at high altitudes or challenges inherent to autonomous navigation in underground mines. In forest or other dense and unstructured environments, robots need to have a robust situational awareness to enable autonomous navigation. Moreover, these scenarios involve specific sensors or environmental variables to be monitored, including snow depth and ATs in avalanches or gas discrimination sensors in underground mines.

%
%
\subsection{Triage}
\label{subsect:human}

In a scene of an accident or a natural disaster, an essential step once victims are found is to follow a triage protocol. Triage is the process through which victims are pre-assessed and their treatment prioritized based on the degree of their injuries. This is particularly important when the mediums to transport or treat the injured are limited, such as in remote locations or when mass casualties occur. 

In~\cite{murphy2013interacting}, the authors explored from the perspective of medical specialists how robots could interact with victims and perform an autonomous triage, as well as which procedures could be used for trapped victims found by robots to interact with medical specialists via the robot's communication interface. This issue had already been studied in a early work by Chang et al.~\cite{chang2007towards}, in which a simple triage and rapid treatment (START) protocol was described. The START protocol is meant to provide first responders with information regarding four key vital signs of encountered victims: blood perfusion, mobility, respiration, and mental state. In~\cite{chang2007towards}, the focus was on analyzing the potential benefits and challenges in robotics technology to assess those vital signs in an autonomous manner. More recently, Ganz et al. presented a triage support method for enhancing the situational awareness of emergency responders in urban SAR operations with the DIORAMA disaster management system~\cite{ganz2015urban}.

The concept of triage has been extended in~\cite{nordstrom2016vessel} to Vessel Triage, a method for assessing distress situations onboard ships in maritime SAR operations.

%
%
\subsection{Shared Autonomy and Human-Swarm Interaction}
\label{subsect:shared}

Depending on the SAR target area, mission objective and rescue strategy, the mode of the operation can be segregated into semi-autonomous and autonomous SAR with varying amount of human supervision. In multi-robot systems and robots involving complex  manipulation (e.g., humanoids) with a high number of degrees of freedom, such as humanoids, the concept of shared autonomy gains importance. Shared autonomy refers to the autonomous control of the majority of degrees of freedom in a system, while designing a control interface for human operators to control a reduced number of parameters defining the global behavior of the system~\cite{sa2014inspection}. For instance, in~\cite{marion2018director} the authors describe the design principles followed to give the operators of a humanoid robot enough situational awareness while simplifying the actual control of the robot via predefined task sequences. This results in automated perception and control, avoiding catastrophic errors due to and exceeding amount of unnecessary information overwhelming the operator, while still enabling timely reaction and operation flexibility for unknown environments. In~\cite{masone2014semi}, a semi-autonomous trajectory generation system for mobile multi-robot systems with integral haptic shared control was presented. The main novelty was not only in providing an interface for controlling a system with a high number of degrees of freedom through a reduced number of parameters defining the path shape, but also in providing real-time haptic feedback. The feedback provides information about the mismatch between the operator's input and the actual behavior of the robots, which is affected by an algorithm performing autonomous collision avoidance, path regularity checks, and attraction to points of interest. The authors performed experiments with a UAV, where different levels of control were given to the operator. Other works by the same authors defining multi-UAV shared control strategies are available in~\cite{franchi2012shared, lee2013semiautonomous}. Taking into account that most existing SAR robotic systems are semi-autonomous or teleoperated~\cite{de2018persistent, freda20193d, cubber2017search}, an appropriate shared autonomy approach is a cornerstone towards mission efficiency and success.

Another research direction in the control of multi-robot systems is human-swarm interaction. An early study comparing two types of interaction was presented by Kolling et al. in~\cite{kolling2013human}, where the authors acknowledge the applicability of such models in SAR operations. The two different types of human-swarm interaction defined in the study were intermittent interaction (through selection of a fixed subset of swarm members) and environmental interaction (through beacons that control swarm members in their vicinity), which also differentiate in being actively or passively influencing robots, respectively.

Within the EU Guardians project, researchers explored the possibilities of human-swarm interaction for firefighting, and defined the main design ideas in~\cite{naghsh2008analysis}. Two different interfaces were designed: a tactile interface situated in the firefighters torso for transmitting the location of hazardous locations, and a visual interface for the swarm of robots to indicate directions to firefighters. The project combined a base station and a swarm of robots designed to collaborate with firefighters on the scene.

%
%
\subsection{Communication}
\label{subsect:communication}

Communication plays an vital role in multi-robot systems due to the need of coordination and information sharing necessary to carry out collaborative tasks. In multi-robot systems, a mobile ad-hoc network (MANET) is often formed for wireless communication and routing messages between the robots. The topology of MANETs and quality of service can vary significantly when robots move. This becomes particularly challenging in remote and unknown environments. Therefore, strategies must be defined to overcome the limitations in terms of communication reliability and bandwidth~\cite{marcotte2019adaptive}. A relatively common approach is to switch between different communication channels or systems to adapt to the environment and requirements of the mission~\cite{jawhar2018networking}. In particular, in heterogeneous multi-robot systems, multiple communication technologies might be utilized to create data links of variable bandwidth and range. Nonetheless, some research efforts have also considered multi-robot exploration and search in communication-limited environments~\cite{de2010autonomous}. In general, owing to the changing characteristics in terms or wireless transmission in different physical mediums, different communication technologies are utilized for various types of robots. An overview of the main communication technologies utilized in multi-robot systems is available in~\cite{jawhar2018networking}, while a review on MANET-based communication for SAR operations is available in~\cite{anjum2017review}.

In terms of standards for safety and management of shared space, Ship Security Alerts System (SSAS) and Automatic Dependent Surveillance-Broadcast (ADS-B) are the main technologies utilized in ships and aircraft, respectively. These technologies enable a better awareness of other vehicles in the vicinity, which is a significant aspect to consider in multi-robot systems. In recent years, UAVs and other robots have seen an increasing adoption of the MAVLink protocol for teleoperation of unmanned vehicles~\cite{atoev2017data}. Both in commercial or industrial UAVs, as well as in research, these technologies are being put together for increased security when multiple UAVs are operating in the same environment~\cite{curtis2017uav, sherman2017autonomous}. 

Collaborative multi-robot systems need to be able to communicate to keep coordinated, but also need to be aware of each other's position in order to make the most out of the shared data~\cite{stroupe2001distributed, queralta2019collaborative}. Situated communication refers to wireless communication technologies that enable simultaneous data transfer while locating the data source~\cite{stoy2001using}. Over the past decade, there have been significant advances towards enabling localization based on traditional and ubiquitous wireless technologies such as WiFi and Bluetooth~\cite{biswas2010wifi, kotaru2015spotfi, sun2018augmentation, altini2010bluetooth, kriz2016improving, wisanmongkol2019multipath, kanaris2017fusing}. These approaches have been traditionally based on the received signal strength indicator (RSSI) and the utilization of either Bluetooth beacons in known locations~\cite{altini2010bluetooth, kriz2016improving, wisanmongkol2019multipath}, or radio maps that define the strength of the signal of different access points over a predefined and surveyed area~\cite{biswas2010wifi, sun2018augmentation}. More recently, other approaches rely on angle-of-arrival~\cite{kotaru2015spotfi}, now built-in in Bluetooth 5.1 devices~\cite{suryavanshi2019direction}. A recent trend has also been to apply deep learning in positioning estimation~\cite{khan2018angle}.

While situated communication often refers to relative localization in two or three-dimensional spaces, if enough information is available (e.g., through sensor fusion or external transmitters in known positions), global localization might be also possible. For example, in~\cite{ito2014w} the authors fuse depth information from an RGB-D camera with the WiFi signal strength to estimate the position of a robot given a floor plan of the environment with minimal information. Alternatively, in~\cite{raghavan2010accurate} Bluetooth beacons in known, predefined locations allow the robots to locate themselves within a global reference frame. In recent years, different wireless technologies have emerged enabling more accurate localization while simultaneously providing a communication channel. Among these, ultra wide-band (UWB) wireless technology has emerged as a robust localization system for mobile robots and, in particular, multi-robot systems~\cite{shule2020uwb}. With most existing research relying on fixed UWB transceivers in known locations~\cite{queralta2020uwbbased}, recent works also show promising results in mobile positioning systems or collaborative localization~\cite{almansa2020autocalibration}.

From the point of view of multi-robot coordination, maintaining connectivity between the different agents participating in a SAR mission is critical. Agents can be robots, human operators, or any other systems that are connected to one or more of the previous and are either producing or processing data. Connectivity maintenance in wireless sensor networks has been a topic of study for the past two decades~\cite{tian2005connectivity}. In recent years, it has gained more attention in the fields of multi-robot systems with decentralized approaches~\cite{sabattini2013decentralized}. Connectivity maintenance algorithms can be designed coupled with distributed control in multi-robot systems~\cite{sabattini2013distributed}, or collision avoidance~\cite{wang2016multi}. Within the literature in this area, global connectivity maintenance algorithms ensure that any two agents are able to communicate either directly or through a multi-hop path~\cite{gasparri2017bounded}. More in line with SAR robotics, Xiao et al. have recently presented a cooperative multi-agent search algorithm with connectivity maintenance~\cite{xiao2019cooperative}. Similar works aiming at cooperative search, surveillance or tracking with multi-robot systems focus on optimizing the data paths~\cite{zhu2017connectivity} or fallible robots~\cite{panerati2019robust, ghedini2018decentralized}. Another recent work in area coverage with connectivity maintenance is available in~\cite{siligardi2019robust}. A comparison of local and global methods for connectivity maintenance of multi-robot networks from Khateri et al. is available in~\cite{khateri2019comparison}. We discuss further the search and area coverage algorithms in Section~\ref{sec:coverage}.

%% file: fig/heterogeneous.tex
\tikzstyle{rw} = [rectangle, minimum width=0.12\textwidth, minimum height=1cm, draw=black, fill=white, align=center]

\begin{tikzpicture}[node distance=2.5cm]

    %
    %
    \node (start) [rw, minimum width=0.18\textwidth, minimum height=6em] {\small{Heterogeneous} \\[+3pt] \small{Multi-Robot Systems} \\[+3pt] \small{in Search and Rescue}};
    
    \node (urban) [rw, right of=start, yshift=+0em, xshift=+3em] {\small Urban \\ \small SAR};
    \node (rural) [rw, right of=start, yshift=-7em, xshift=+3em] {\small Wilderness \\ \small SAR};
    \node (marit) [rw, right of=start, yshift=+7em, xshift=+3em] {\small Maritime \\ \small SAR};
    
    \node (u1) [right of=urban, yshift=+2em, minimum width=0.60\textwidth, align=left, anchor=west, text width=0.60\textwidth, xshift=-2em] {\footnotesize UAVs: aid in initial assessment, emergency networks, and surveillance.};
    \node (u2) [right of=urban, yshift=+0em, minimum width=0.60\textwidth, align=left, anchor=west, text width=0.60\textwidth, xshift=-2em] {\footnotesize UGVs: able of dexterous manipulation, full-body telepresence, and reaching to victims.};
    \node (u3) [right of=urban, yshift=-2em, minimum width=0.60\textwidth, align=left, anchor=west, text width=0.60\textwidth, xshift=-2em] {\footnotesize USVs: support units in flooded coastal areas and rivers.};
    
    \node (r1) [right of=rural, yshift=+1em, minimum width=0.60\textwidth, align=left, anchor=west, text width=0.60\textwidth, xshift=-2em] {\footnotesize UAVs: mapping, search of victims, monitoring, and transportation in remote areas.};
    \node (r2) [right of=rural, yshift=-1em, minimum width=0.60\textwidth, align=left, anchor=west, text width=0.60\textwidth, xshift=-2em] {\footnotesize UGVs: aid in underground caves and mines, searching victims, identifying hazards.};
    
    \node (m1) [right of=marit, yshift=+2em, minimum width=0.60\textwidth, align=left, anchor=west, text width=0.60\textwidth, xshift=-2em] {\footnotesize UAVs: aid in enhancing the situational awareness of surface units from the air.};
    \node (m2) [right of=marit, yshift=+0em, minimum width=0.60\textwidth, align=left, anchor=west, text width=0.60\textwidth, xshift=-2em] {\footnotesize USVs: main actors in transportation and reaching to victims.};
    \node (m3) [right of=marit, yshift=-2em, minimum width=0.60\textwidth, align=left, anchor=west, text width=0.60\textwidth, xshift=-2em] {\footnotesize UUVs: operate in harsh environments, search victims and assess underwater damages.};

    %
    %
    \draw [arrow] (start) |- ($(start.east) + (1em,0)$) |- (urban);
    \draw [arrow] (start) |- ($(start.east) + (1em,0)$) |- (rural);
    \draw [arrow] (start) |- ($(start.east) + (1em,0)$) |- (marit);
    
    \draw [arrow] (urban) |- ($(urban.east) + (1em,0)$) |- (u1);
    \draw [arrow] (urban) |- ($(urban.east) + (1em,0)$) |- (u2);
    \draw [arrow] (urban) |- ($(urban.east) + (1em,0)$) |- (u3);
    
    \draw [arrow] (rural) |- ($(rural.east) + (1em,0)$) |- (r1);
    \draw [arrow] (rural) |- ($(rural.east) + (1em,0)$) |- (r2);

    \draw [arrow] (marit) |- ($(marit.east) + (1em,0)$) |- (m1);
    \draw [arrow] (marit) |- ($(marit.east) + (1em,0)$) |- (m2);
    \draw [arrow] (marit) |- ($(marit.east) + (1em,0)$) |- (m3);
    
\end{tikzpicture}

%% file: sections/03_Coordination_Planning.tex
\section{Multi-Robot Coordination}\label{sec:coverage}

In this section, we describe the main algorithms required for multi-robot coordination in collaborative applications. We discuss this mainly from the point of view of cooperative multi-robot systems, while focusing on their applicability towards SAR missions. The literature in multi-robot cooperative exploration or collaborative sensing contains mostly generic approaches that consider multiple applications. Whenever the literature has enough examples, we discuss how SAR approaches differ or are characterized.

Through this section, we describe and differentiate between centralized and distributed multi-agent control and coordination approaches, while also describing the main algorithms utilized in path planning and area coverage for single and multiple robots. In addition, we put these algorithms into the context of deployment across the different SAR scenarios, describing the main constraints to consider as well as the predominant approaches in each field.

\subsection{Multi-Robot Task Allocation}
\label{subsect:task}

Search and rescue operations with multi-robot systems involve aspects including collaborative mapping and situational assessment~\cite{tian2019reliable}, distributed and cooperative area coverage~\cite{maza2007multiple}, or cooperative search~\cite{suarez2011survey}. These or other cooperative tasks involve the distribution of tasks and objectives among the robots (e.g., areas to be searched, or positions to be occupied to ensure connectivity among the robots and with the base station). In a significant part of the existing multi-robot SAR literature, this is predefined or done in a centralized manner~\cite{merino2005cooperative, kruijff2014designing, surmann2019integration, cubber2017search}. Here, we discuss instead distributed multi-robot task allocation algorithms that can be applied to SAR operations. Distributed algorithms have the general advantage of being more robust in adverse environments against the loss of individual agents or when the communication with the base station is unstable.

A comparative study on task allocation algorithms for multi-robot exploration was carried out by Faigl et al. in~\cite{faigl2014comparison}, considering five distinct strategies: greedy assignment, iterative assignment, Hungarian assignment, multiple traveling salesman assignment, and MinPos. However, most of these approaches are often centralized from the decision-making point of view, even if they are implemented in a distributed manner. Successive works have been presenting more decentralized methods. Decentralized task allocation algorithms for autonomous robots are very often based on market-based approaches and auction mechanisms to achieve consensus among the agents~\cite{hussein2014multi, zhao2015heuristic, tang2018using, tadewos2019fly}. Both of this approaches have been extensively studied for the past two decades within the multi-robot and multi-agent systems communities~\cite{dias2006market, mosteo2010survey}. Bio-inspired algorithms have also been widely studied within the multi-robot and swarm robotics domains. For instance, in~\cite{kurdi2016bio}, Kurdi et al. present a task allocation algorithm for multi-UAV SAR systems inspired by locust insects. Active perception techniques have also been incorporated in multi-robot planning algorithms in existing works~\cite{sage, Best2018}

An early work in multi-robot task allocation for SAR missions was presented by Hussein et al.~\cite{hussein2014multi}, with a market-based approach formulated as a multiple traveling salesman problem. The authors applied their algorithm to real robots with simulated victim locations that the robots had to divide among themselves and visit. The solution was optimal (from the point of view of distance travelled by the robots) and path planning for each of the robots was also taken into account. The authors, however, did not study the potential for scalability with the number of robots or victim locations, or consider the computational complexity of the algorithm. In that sense, and with the aim of optimizing the computational cost owing to the non-polynomial complexity nature of optimal task allocation mechanisms, Zhao et al. presented a heuristic approach~\cite{zhao2015heuristic}. The authors introduced a significance measure for each of the tasks, and utilized both victim locations and terrain information as optimization parameters within their proposed methodology. The algorithm was tested under a simulation environment with a variable number of rescue robots and number of survivor locations to test the scalability and optimality under different conditions.

An auction-based approach aimed at optimizing a cooperative rescue plan within multi-robot SAR systems was proposed by Tang et al.~\cite{tang2018using}. In this work, the emphasis was also put on the design of a lightweight algorithm more appropriate for ad-hoc deployment in SAR scenarios.

A different approach where a human supervisor was considered appears in~\cite{liu2015supervisory}. Liu et al. presented in this work a methodology for task allocation in heterogeneous multi-robot systems supporting USAR missions. By relying on a supervised system, the authors show better adaptability to situations with robot failures. The algorithm was tested under a simulation environment where multiple semi-autonomous robots were controlled by a single human operator.


\subsection{Formation Control and Pattern Formation}
\label{subsect:formation}

Formation control or pattern formation algorithms are those that define spatial configurations in multi-robot systems~\cite{oh2015survey, mccord2019progressive}.  Most formation control algorithms for multi-agent systems can be roughly classified in three categories from the point of view of the variables that are measured and actively controlled by each of the agents: position-based control, displacement-based control, and distance or bearing-based control~\cite{oh2015survey}. 
%
%
Formation control algorithms requiring global positioning are often implemented in a centralized manner, or through collaborative decision making. Displacement and distance or bearing-based control, however, enable more distributed implementations with only local interactions among the different agents~\cite{shamma2008cooperative, queralta2019progressive}, as well as those algorithms where no communication is required among the agents~\cite{queralta2019indexfree}. 

In SAR operations, formation control algorithms are an integral part of multi-robot ad-hoc networks or MANETs~\cite{abbasi2014link, lin2014adaptive}, multi-robot emergency surveillance and situational awareness networks~\cite{ray2006dynamic}, or even a source of communication in human-swarm interaction~\cite{saez2010multi}. In MANETs, formation control algorithms are utilized to maintain configurations meeting certain requirements in terms of coverage or bandwidth provided by the network. Simple distance-based formation control includes flocking algorithms, where agents try to maintain a constant and common distance between them and each of their nearest neighbors, thus achieving a mostly homogeneous tessellation covering a certain area. This can be applied in MANETs for covering a certain area homogeneously~\cite{aftab2017self}. In surveillance and monitoring missions, formation control algorithms can aid in fixing the viewpoint of different agents for collaborative perception of the same area or subject. Finally, in human-swarm interaction, these algorithms can be employed to communicate different messages from a swarm by assigning different meanings or messages to a predefined set of spatial configurations. This have been studied, to some extent, in swarms of robots aiding firefighting~\cite{saez2010multi}.

\begin{figure}
    \centering
    \begin{subfigure}[t]{0.23\textwidth}
        \centering
        \includegraphics[width=.9\textwidth]{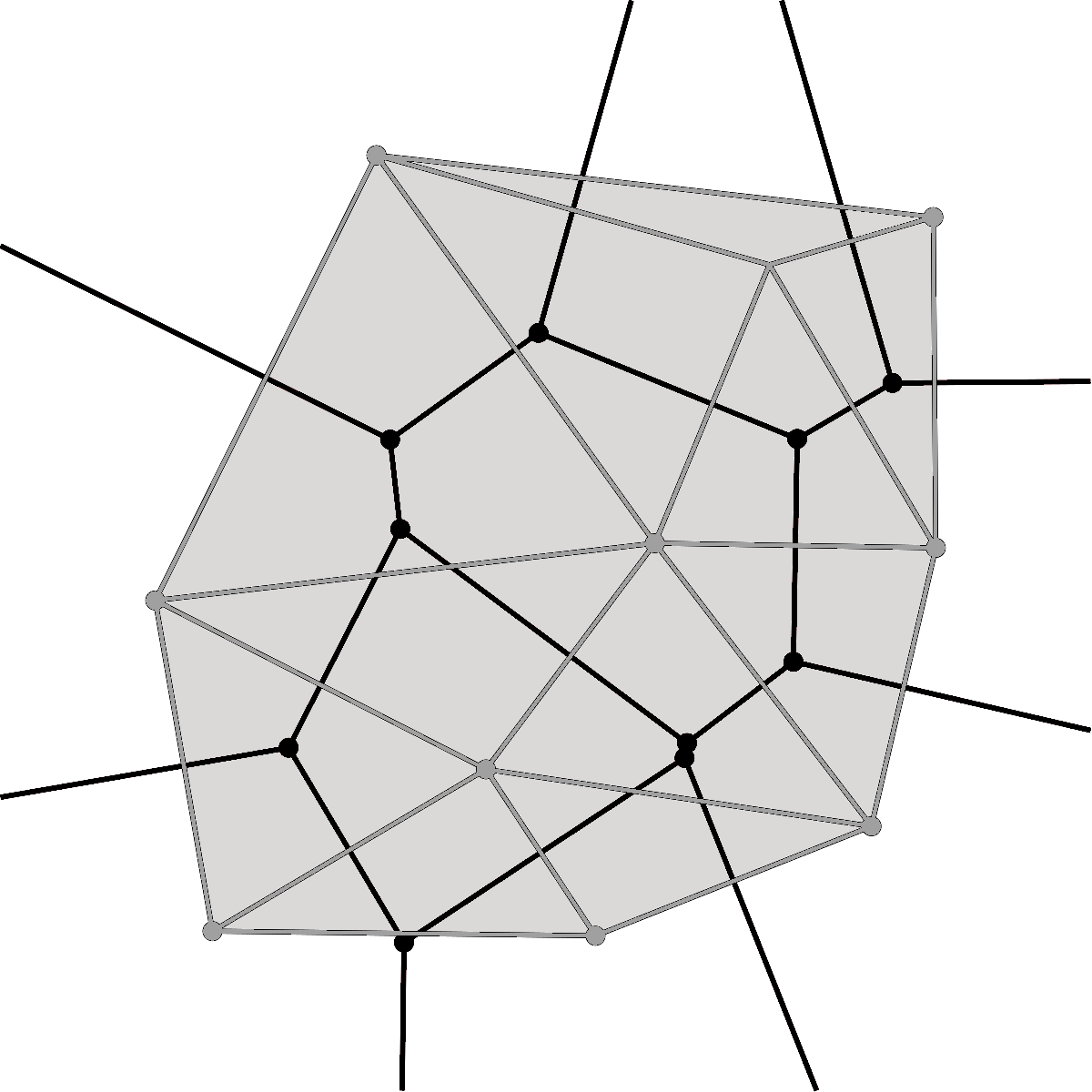}
        \caption{Voronoi regions}
        \label{subfig:voronoi}
    \end{subfigure}
    \begin{subfigure}[t]{0.23\textwidth}
        \includegraphics[width=.9\textwidth]{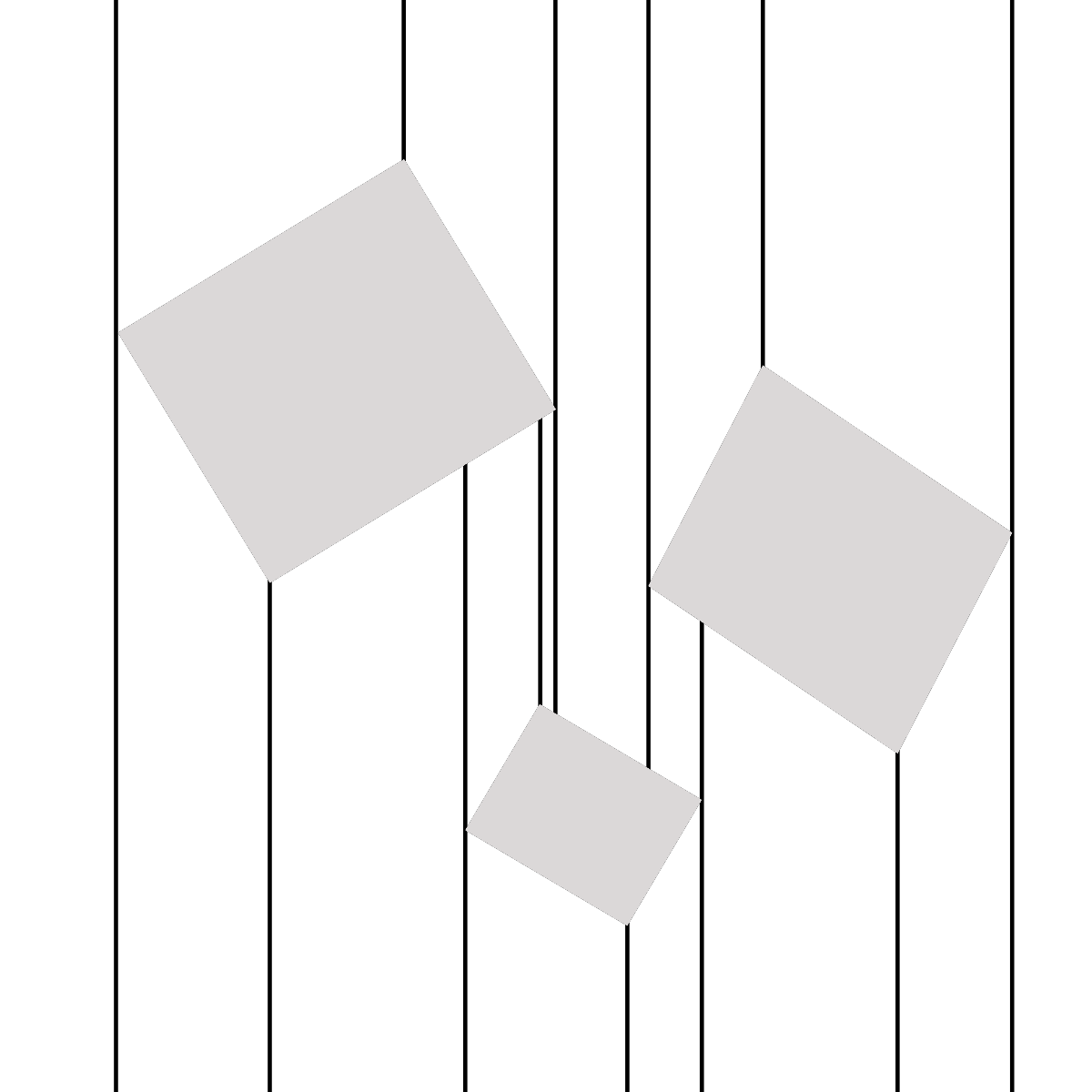}
        \caption{Exact cells}
    \end{subfigure}
    
    \vspace{1ex}
    
    \begin{subfigure}[t]{0.23\textwidth}
        \includegraphics[width=0.8\textwidth]{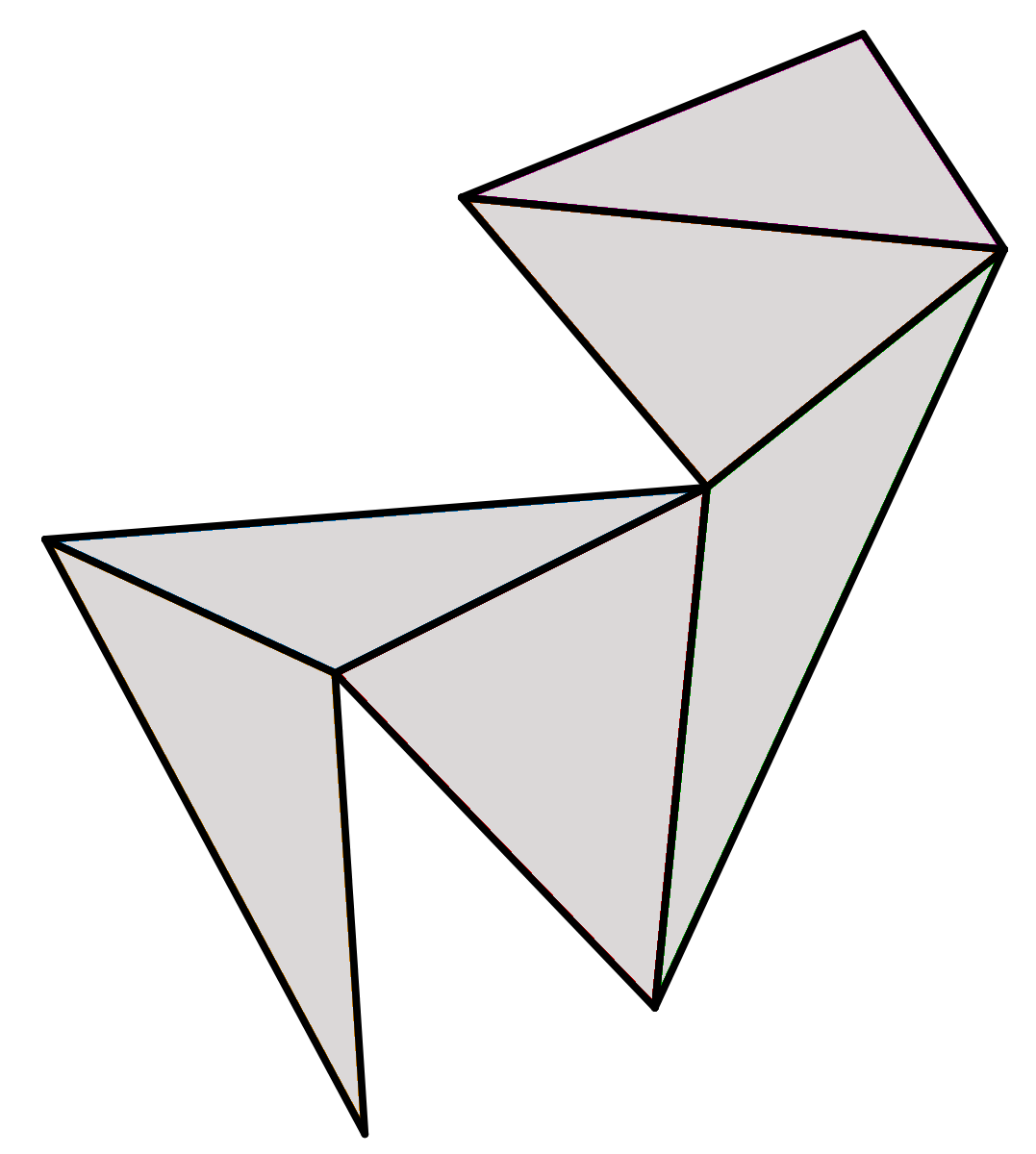}
        \caption{Area triangulation}
    \end{subfigure}
    \begin{subfigure}[t]{0.23\textwidth}
        \includegraphics[width=\textwidth]{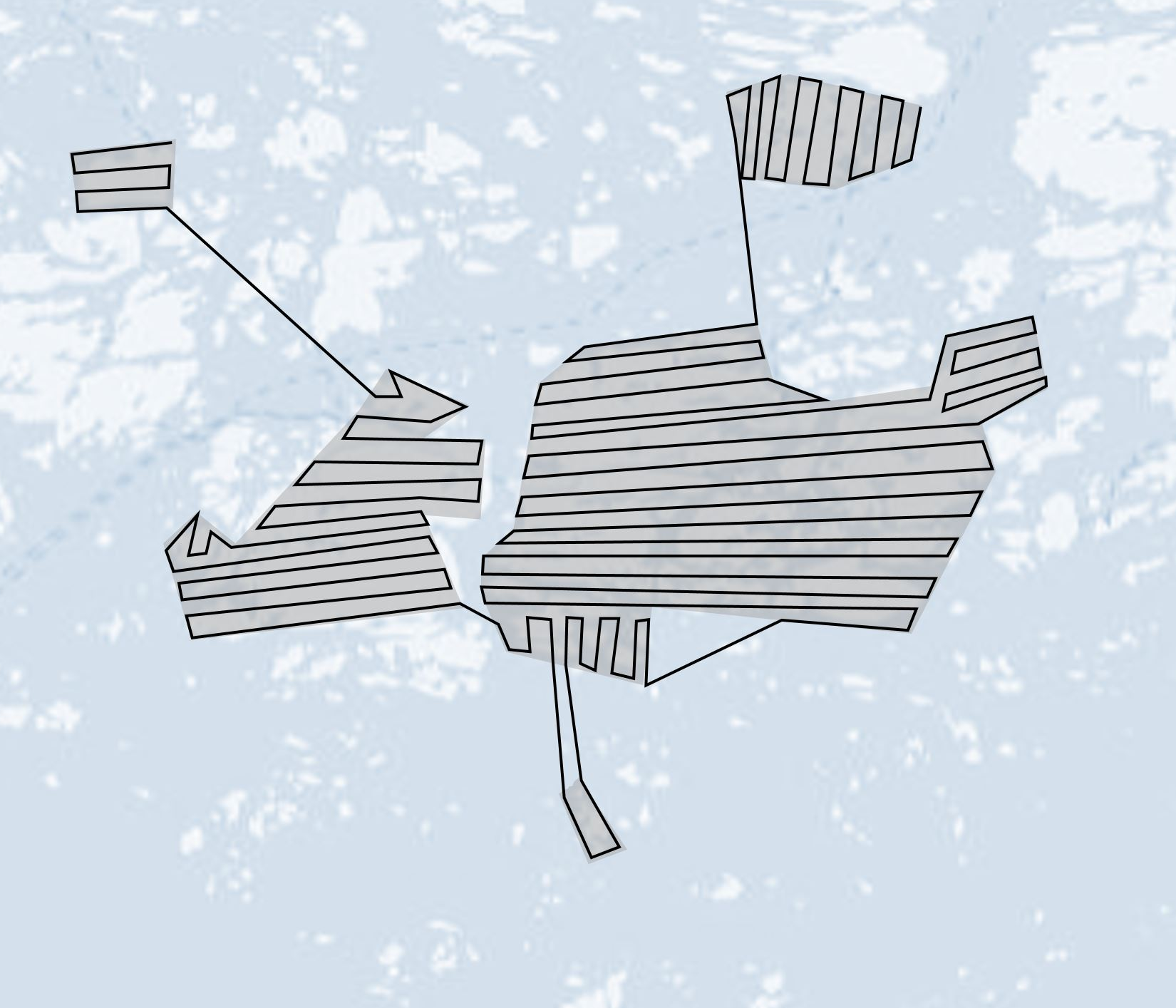}
        \caption{Disjoint area coverage}
    \end{subfigure}
    \caption{Illustration of different basic area decomposition and coverage algorithms: (i) decomposition through voronoi regions, (ii) exact cell decomposition, (iii) polygonal decomposition (triangular in this case), and (iv) disjoint area coverage. The resulting decompositions or coverage paths are marked with black lines, while the original areas are shown in gray colors.}
    \vspace{-1em}
    \label{fig:area_decomposition}
\end{figure}

\subsection{Area Coverage}
\label{subsect:coverage}

The first step in SAR missions is the search and localization of the persons to be rescued. Therefore, an essential part of autonomous SAR operations is path planning and area coverage. To this end, multiple algorithms have been presented for different types of robots or scenarios. Coverage path planning algorithms have been recently focused towards UAVs~\cite{cabreira2019survey}, owing to their higher degrees of freedom of movement when compared to UGVs. Path planning and area coverage algorithms take into account mainly the shape of the objective area to be surveyed. Nonetheless, a number of other variables are also considered in more complex algorithms, such as energy consumption, range of communication and bandwidth, environmental conditions, or the probability of failure. The specific dynamics and capabilities of the robots being used can also be utilized to optimize the performance of the area coverage, for example when comparing the maneuverability of quadrotors and fixed-wing UAVs. 

Area coverage algorithms can be broadly classified in terms of the assumptions they make on the geometry of the area to be covered. The most basic approaches consider only convex and joint areas~\cite{maza2007multiple}, for which paths can be efficiently generated based on area decomposition algorithms~\cite{hert1998polygon, araujo2013multiple}. 
Some of the most common area decomposition and coverage algorithms are shown in Fig.~\ref{fig:area_decomposition}.
In the presence of known obstacles inside the objective area, the search area can be considered non-convex~\cite{kantaros2014visibility}. However, non-convex approaches can often be applied to more general environments. More realistic scenarios, in particular in the field of SAR operations, often require the exploration of disjoint areas~\cite{vasquez2018coverage}. The problem of disjoint area search can be formulated as a multi-objective optimization problem, where each of the joint subareas can be considered a single objective in the path planning object. This leads to the differentiation between multi-agent single-objective planning and multi-agent multi-objective optimization and planning. The former case is not necessarily a subset of the latter, as it also includes use cases such as collaborative transportation, applicable in emergency scenarios, or can provide higher degrees of fault-tolerance and robustness against the loss of agents. The latter case, nonetheless, is more significant within the scope of this survey as multi-agent multi-objective optimization algorithms enable more efficient search in complex environments with distributed systems~\cite{hayat2017multi, georgiadou2017multi}. Finally, the most comprehensive approaches also account for the existence of unknown environments in the areas to be searched, with the existence of potential obstacles that are a priori unknown. In order to reach real-world deployment of autonomous robots in post-disaster and unknown environments for SAR tasks, algorithms that consider uncertainty in the environment must be further developed~\cite{papatheodorou2016distributed, zhao2018decentralized}.

When multi-robot systems are utilized, the increased number of variables already involved in single-agent planning increase the complexity of the optimization problems while at the same time bring new possibilities to more efficient area coverage. For instance, energy awareness among the agents could enable robots with less operational time to survey areas near the deployment point, while other robots can be put in charge of farther zones. The communication system being utilized and strategies for connectivity maintenance play a more important role in multi-robot systems. If the algorithms are implemented in a distributed manner or the robots rely on online path planning, then the paths themselves must ensure robust connectivity enabling proper operation of the system. Security within the communication, while also important in the single-agent case, plays again a more critical role when multiple robots communicate among themselves in order to take cooperative decisions in real-time.

Furthermore, the optimization problems upon which multi-robot area coverage algorithms build are known to belong to the NP-hard class of non-deterministic polynomial time algorithms~\cite{garey1982complexity}. Therefore, part of the existing research has focused towards probabilistic approaches. This naturally fits to SAR operations since, after an initial assessment of the environment, SAR personnel can get an a priori idea of the most probable locations for victims~\cite{le2012adaptive}. The idea of using probability distributions in the multi-objective search optimization problem has also been extended towards actively updating these distributions as new sensor data becomes available~\cite{schwager2009decentralized}.


\subsection{Single-Agent Planning}
\label{subsect:single_planning}

The most basic algorithms consider only convex area coverage, except for potential obstacles or no-flight areas that might appear within the objective area. An outline of these algorithms is presented by Cabreira et al. in a recent survey on coverage path planning for UAVs~\cite{cabreira2019survey}. This survey covers mainly single-agent planning for either convex or concave areas (with the presence of obstacles or no-flight zones) and puts the focus on algorithms for area decomposition.

Planning in SAR scenarios can pose additional challenges to well-established planning strategies for autonomous robots. In particular, the locations of victims trapped under debris or inside cave-like structures might be relatively easy to determine but significantly complex to access, thus requiring specific planning strategies. In~\cite{suarez2011survey}, Suarez et al. present a survey of animal foraging strategies applied to rescue robotics. The main methods that are discussed are directed search (search space division with memory- and sensory-based search) and persistent search (with either predefined time limits or constraint-optimization for deciding how long to persist on the search).

Path planning algorithms can be part of area coverage algorithms or implemented separately for robots to cover their assigned areas individually. In any case, when area coverage algorithms consider path planning, it is often from a global point of view, leaving the local planning to the individual agents. A detailed description of path planning algorithms including approaches of linear programming, control theory, multi-objective optimization models, probabilistic models, and meta-heuristic models for different types of UAVs is available in~\cite{aggarwal2020path}. While some of these algorithms are generic and only take into account the origin and objective position, together with obstacle positions, others also consider the dynamics of the vehicles and constraints that these naturally impose in local curvatures, such as Dubin curves~\cite{aggarwal2020path}.

Recent works have considered more complex environments. For instance, in~\cite{xie2020path}, Xie et al. presented a path planning algorithm for UAVs covering disjoint convex regions. The authors' method considered an integration of both coverage path planning and the traveling salesman problem. In order to account for scalability and real-time execution, two approaches were presented: a near-optimal solution based on dynamic programming, and a heuristic approach able to efficiently generate high-quality paths, both tested under simulation environments. Also aiming at disjoint but convex areas, Vazquez et al. proposed a similar method that separates the optimization of the order in which the different areas were visited and the path generation for each of them~\cite{vasquez2018coverage}. Both of this cases, however, provide solutions for individual UAVs.

\subsection{Planning for different robots: UAVs, UGVs, UUVs and USVs}
\label{subsect:different_planning}

Mobile robots operating on different mediums necessarily have different constraints and a variable number of degrees of freedom. For local path planning, a key aspect to consider when designing control systems is the holonomic nature of the robot. In a holonomic robot, the number of controllable degrees of freedom is equal to the number of degrees of freedom defining the robot's state. In practice, most robots are non-holonomic, with some having significant limitations to their local motion such as fixed-wing UAVs~\cite{lugo2014dubins}, or USVs~\cite{liao2010full}. However, quadrotor UAVs, which have gained considerable momentum owing to their flexibility and relatively simple control, can be considered holonomic~\cite{cetinsoy2013design}. Ground robots equipped with omniwheel mechanisms and able of omnidirectional motion can be also considered holonomic if they operate on favorable surfaces~\cite{damoto2001holonomic}.

\begin{figure*}
    \centering
    \footnotesize
    \input{fig/constraints}
    \vspace{1ex}
    \caption{Main path planning constraints that autonomous robots in different domains need to account for. Some of these aspects are common across the different types of robots, such as energy efficiency and inherent constraints from the robots' dynamics, but become more predominant in UAVs and USVs, for instance.}
    \label{fig:constraints}
\end{figure*}
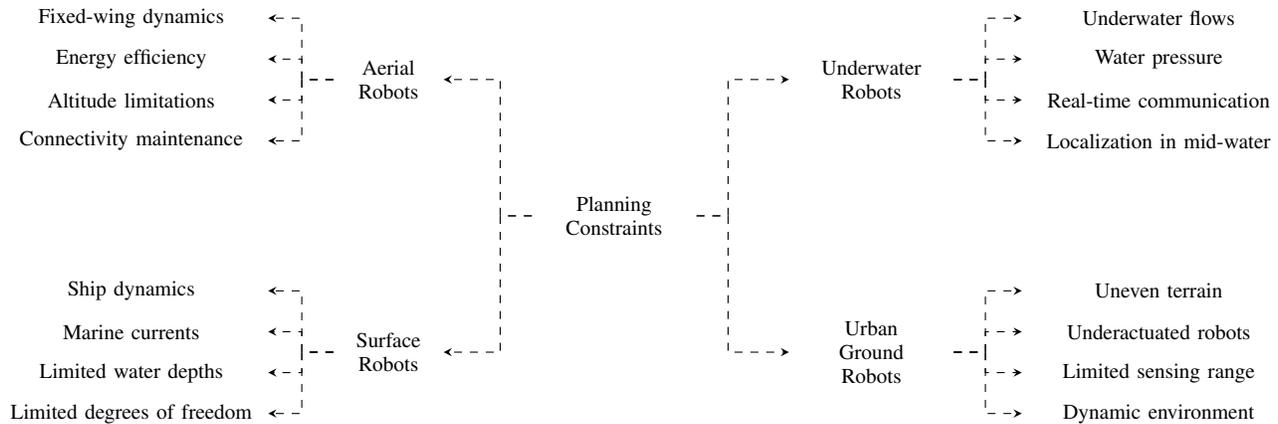

Multiple works have been devoted to reviewing the different path planning strategies for unmanned vehicles in different mediums: aerial robots~\cite{aggarwal2020path}, surface robots~\cite{campbell2012review}, underwater robots~\cite{zeng2015survey, li2018path}, and ground robots for urban~\cite{liu2013robotic}, or wilderness~\cite{santos2020path} environments. From these works, we have summarized the main constraints to be considered in path planning algorithms in Fig.~\ref{fig:constraints}.

The main limitations in robot navigation, and therefore path planning, in different mediums can be roughly characterized by: (i) dynamic environments and movement limitations in ground robots; (ii) energy efficiency, situational awareness, and weather conditions in aerial robots; (iii) underactuation and environmental effects in surface robots, with currents, winds and water depth constraints; and (iv) localization and communication in underwater robots. Furthermore, these constraints increase significantly in SAR operations, with earthquakes aggravating the movement limitations of UGVs, or fires and smoke preventing normal operation of UAVs. Some emergency scenarios, such as flooded coastal areas, combine multiple of the above mediums making the deployment of autonomous robots even more challenging. For instance, in~\cite{ozkan2019rescue}, the authors describe path planning techniques for rescue vessels in flooded urban environments, where many of the limitations of urban navigation are added to the already limited navigation of surface robots in shallow waters.

A key parameter to take into account in autonomous robots, and particularly in UAVs, is energy consumption. This becomes critical in SAR operations owing to the time constraints and need for optimizing search tasks. UAVs are known to have relatively limited operational time, and therefore energy consumption is a variable to consider in the different optimization problems to be solved. In this direction, Di Franco et al. presented an algorithm for energy-aware path planning with UAVs~\cite{di2015energy}. A more recent work considering energy-aware path planning for area coverage introduces a novel algorithm for path planning that minimizes turns~\cite{cabreira2018energy}. The authors report energy savings of 10\% to 15\% with their novel spiral-inspired path planning algorithm, while meeting minimum requirements from the point of view of visual sensing and altitude maintenance for achieving a resolution enabling mission success. Energy efficiency is a topic that has also been considered in USVs. In~\cite{lee2015energy}, the authors introduced an energy-efficient 3D (two-dimensional positioning and one-dimension for orientation) path planning algorithm that would take into account both environmental effects (marine currents, limited water depth) and the heading or orientation of the vehicle (in the start and end positions). 

Owing to the flexibility of quadrotor UAVs, they have been utilized with different roles in more complex robotic systems. For instance, in~\cite{nazarova2020application} the authors describe a heterogeneous multi-UAV system for earthquake SAR where some of the UAVs are in charge of providing reliable network connection, as a sort of air communication station, while smaller UAVs flying close to the ground are in charge of the actual search tasks.

\subsection{Multi-Robot Planning}
\label{subsect:multi_planning}

Research in the field of multi-robot path planning has been ongoing for over two decades. An early approach to multi-robot cooperation was presented in~\cite{alami1995multi} in 1995, where the authors introduced an incremental plan-merging approach that defined a global plan shared among the robots. A relatively simple yet effective mechanism was utilized to maintain a consistent global plan: the robots would ask others for the right to plan for themselves and update the global plan accordingly one by one. This approach, while distributed, would not match the real-time needs and standards of today, nor does it exploit parallel operations at the robots during the distributed planning. In~\cite{hert1998polygon}, an early generalization of previous algorithms towards nonconvex and nonsimply connected areas was presented, enabling deployment in more realistic scenarios. The advances since then have been significant in multiple directions. With the idea or providing fault-tolerant systems, in~\cite{maza2007multiple} the authors introduced a reconfiguration process that would account in real-time for malfunctioning or missing agents, and adjust the paths of remaining agents accordingly. Considering the need of inter-robot communication for aggregating and merging data, a cooperative approach to multi-robot exploration that considers the range limitations of the communication system between robots was introduced in~\cite{yuan2010cooperative}. Non-polygonal area partitioning methods have also been proposed. In~\cite{jain2012multi}, a circle partitioning method that the authors claim to be applicable to real-world SAR operations was presented.

Existing approaches often differentiate between area coverage and area exploration. In area coverage algorithms, algorithms focus on optimally planning paths for traversing a known area, or dividing a known area among multiple agents to optimize the time it takes to analyze it. Area exploration algorithms focus instead on the coverage and mapping of potentially unknown environments. The two terms, however, are often used interchangeably in the literature. An overview and comparison of multi-robot area exploration algorithms is available in~\cite{singh2014comparative}.

In~\cite{choi2019multi}, Choi et al. present a solution for multi-UAV systems, which is in turn focused at disaster relief scenarios. In particular, the authors developed this solution in order to improve the utilization of UAVs when fighting multiple wildfires simultaneously. Also considering multi-UAV path planning, but including non-convex disjoint areas, Wolf et al. proposed a method were the operator could input a desired overlap in the search areas~\cite{wolf2019path}. This can be of particular interest in heterogeneous multi-robot systems where different robots have different sensors, and the search personnel wants multiple robots to travel over some of the areas. Finally, another recent work in cooperative path planning that focuses on mountain environments and can be of specific interest in WiSAR operations was presented by Li et al.~\cite{li2018multi}.


\subsection{Multi-Objective Multi-Agent Optimization}
\label{subsect:multi_objective}

From a theoretical point of view, a multi-agent collaborative search problem can be formulated and solved as a multi-agent and multi-objective optimization problem in a certain space~\cite{ricciardi2019improved, vasile2011multiagentcollaborativesearch}.

In post-disaster scenarios and emergency situations in general, an initial assessment of the environment often provides rescue personnel an idea of the potential spatial distribution of victims~\cite{nazarova2020application}. In those cases, different a priori probabilities can be assigned to different areas, providing a ranking of locations for the multi-objective optimization problem. The literature involving multi-agent multi-objective optimization for SAR operations is, however, sparse. In~\cite{hayat2017multi}, Hayat et al. proposed a genetic algorithm for multi-UAV search in a bounded area. One of the key novelties of this work is that the authors consider simultaneously connectivity maintenance among the UAV network as well as optimization of area coverage. Moreover, the algorithm could be adjusted to give more priority to either coverage or connectivity, depending on the mission requirements.

A different approach to multi-objective optimization within the SAR domain was taken by Georgiadou et al.~\cite{georgiadou2017multi}. In this work, the authors presented a method for improving disaster response after accidents in chemical plants. Rather than considering locations as objectives for the optimization problem, the authors introduced multiple criteria within their algorithm, from the assessment of hazards in the environment to the evacuation or protection of buildings. In a similar direction, a multi-objective evolutionary algorithm aimed at general emergency response planning was proposed by Narzisi et al. in~\cite{narzisi2006multi}.

\subsection{Planning in Heterogeneous Multi-Robot Systems}
\label{subsect:planning_heterogeneous}

Most existing approaches for multi-robot exploration or area coverage either assume that all agents share similar operational capabilities, or that the characteristics of the different agents are known a priori. Emergency deployments in post-disaster scenarios for SAR of victims, however, requires flexible and adaptive systems. Therefore, algorithms able to adapt to heterogeneous robots that potentially operate on different mediums and with different constraints (e.g., UAVs and UGV collaborating in USAR scenarios) need to be utilized. In this direction, Mueke et al. presented a system-level approach for distributed control of heterogeneous systems with applications to SAR scenarios~\cite{muecke2011distributed}. In general, we see a lack of further research in this area, as most existing projects and systems involving heterogeneous robots predefine the way in which they are meant to cooperate. From a more general perspective, an extensive review on control strategies for collaborative area coverage in heterogeneous multi-robot systems was recently presented by Abbasi~\cite{abbasi2016coverage}. Also from a general perspective, a survey on cooperative heterogeneous multi-robot systems by Rizk et al. is available in~\cite{rizk2019cooperative}.


%% file: fig/constraints.tex
\begin{tikzpicture}[node distance=3cm]

    %
    %
    \node (start) [text centered, minimum width=0.12\textwidth, align=center] {Planning \\ Constraints};
    
    \node (aerial) [left of=start, yshift=+0.1\textwidth, align=center, minimum width=0.08\textwidth] {Aerial \\ Robots};
    \node (surface) [left of=start, yshift=-0.1\textwidth, align=center, minimum width=0.08\textwidth] {Surface \\ Robots};
    
    \node (underwater) [right of=start, yshift=+0.1\textwidth, xshift=0.42cm, align=center, minimum width=0.12\textwidth] {Underwater \\ Robots};
    \node (ground) [right of=start, yshift=-0.1\textwidth, xshift=0.42cm, align=center, minimum width=0.12\textwidth] {Urban \\ Ground \\ Robots};
    
    \node (dyn) [left of=aerial, yshift=+0.045\textwidth, xshift=-0.42cm, minimum width=0.2\textwidth] {Fixed-wing dynamics};
    \node (ene) [left of=aerial, yshift=+0.015\textwidth, xshift=-0.42cm, minimum width=0.2\textwidth] {Energy efficiency};
    \node (alt) [left of=aerial, yshift=-0.015\textwidth, xshift=-0.42cm, align=left, minimum width=0.2\textwidth] {Altitude limitations};
    \node (con) [left of=aerial, yshift=-0.045\textwidth, xshift=-0.42cm, align=left, minimum width=0.2\textwidth] {Connectivity maintenance};
    
    \node (sdy) [left of=surface, yshift=+0.045\textwidth, xshift=-0.42cm, minimum width=0.2\textwidth] {Ship dynamics};
    \node (cur) [left of=surface, yshift=+0.015\textwidth, xshift=-0.42cm, minimum width=0.2\textwidth] {Marine currents};
    \node (dep) [left of=surface, yshift=-0.015\textwidth, xshift=-0.42cm, align=left, minimum width=0.2\textwidth] {Limited water depths};
    \node (sth) [left of=surface, yshift=-0.045\textwidth, xshift=-0.42cm, align=left, minimum width=0.2\textwidth] {Limited degrees of freedom};
    
    \node (uw1) [right of=underwater, yshift=+0.045\textwidth, xshift=0.82cm, minimum width=0.2\textwidth] {Underwater flows};
    \node (uw2) [right of=underwater, yshift=+0.015\textwidth, xshift=0.82cm, minimum width=0.2\textwidth] {Water pressure};
    \node (uw3) [right of=underwater, yshift=-0.015\textwidth, xshift=0.82cm, align=left, minimum width=0.2\textwidth] {Real-time communication};
    \node (uw4) [right of=underwater, yshift=-0.045\textwidth, xshift=0.82cm, align=left, minimum width=0.2\textwidth] {Localization in mid-water};
    
    \node (su1) [right of=ground, yshift=+0.045\textwidth, xshift=0.82cm, minimum width=0.2\textwidth] {Uneven terrain};
    \node (su2) [right of=ground, yshift=+0.015\textwidth, xshift=0.82cm, minimum width=0.2\textwidth] {Underactuated robots};
    \node (su3) [right of=ground, yshift=-0.015\textwidth, xshift=0.82cm, align=left, minimum width=0.2\textwidth] {Limited sensing range};
    \node (su4) [right of=ground, yshift=-0.045\textwidth, xshift=0.82cm, align=left, minimum width=0.2\textwidth] {Dynamic environment};
    
    \draw [arrow] (start) -| ($(start.west) - (0.42cm, 0)$) |- (aerial);
    \draw [arrow] (start) -| ($(start.west) - (0.42cm, 0)$) |- (surface);
    
    \draw [arrow] (start) -| ($(start.east) + (0.42cm, 0)$) |- (underwater);
    \draw [arrow] (start) -| ($(start.east) + (0.42cm, 0)$) |- (ground);
    
    \draw [arrow] (aerial) -| ($(aerial.west) - (0.42cm, 0)$) |- (dyn);
    \draw [arrow] (aerial) -| ($(aerial.west) - (0.42cm, 0)$) |- (ene);
    \draw [arrow] (aerial) -| ($(aerial.west) - (0.42cm, 0)$) |- (alt);
    \draw [arrow] (aerial) -| ($(aerial.west) - (0.42cm, 0)$) |- (con);
    
    \draw [arrow] (surface) -| ($(surface.west) - (0.42cm, 0)$) |- (sdy);
    \draw [arrow] (surface) -| ($(surface.west) - (0.42cm, 0)$) |- (cur);
    \draw [arrow] (surface) -| ($(surface.west) - (0.42cm, 0)$) |- (dep);
    \draw [arrow] (surface) -| ($(surface.west) - (0.42cm, 0)$) |- (sth);
    
    \draw [arrow] (underwater) -| ($(underwater.east) + (0.42cm, 0)$) |- (uw1);
    \draw [arrow] (underwater) -| ($(underwater.east) + (0.42cm, 0)$) |- (uw2);
    \draw [arrow] (underwater) -| ($(underwater.east) + (0.42cm, 0)$) |- (uw3);
    \draw [arrow] (underwater) -| ($(underwater.east) + (0.42cm, 0)$) |- (uw4);
    
    \draw [arrow] (ground) -| ($(ground.east) + (0.42cm, 0)$) |- (su1);
    \draw [arrow] (ground) -| ($(ground.east) + (0.42cm, 0)$) |- (su2);
    \draw [arrow] (ground) -| ($(ground.east) + (0.42cm, 0)$) |- (su3);
    \draw [arrow] (ground) -| ($(ground.east) + (0.42cm, 0)$) |- (su4);
    
    

    %
    %
    
    
    

\end{tikzpicture}

%% file: sections/04_Perception.tex
\begin{figure*}
    \centering
    \begin{subfigure}[t]{0.48\textwidth}
        \centering
        \includegraphics[width=\columnwidth]{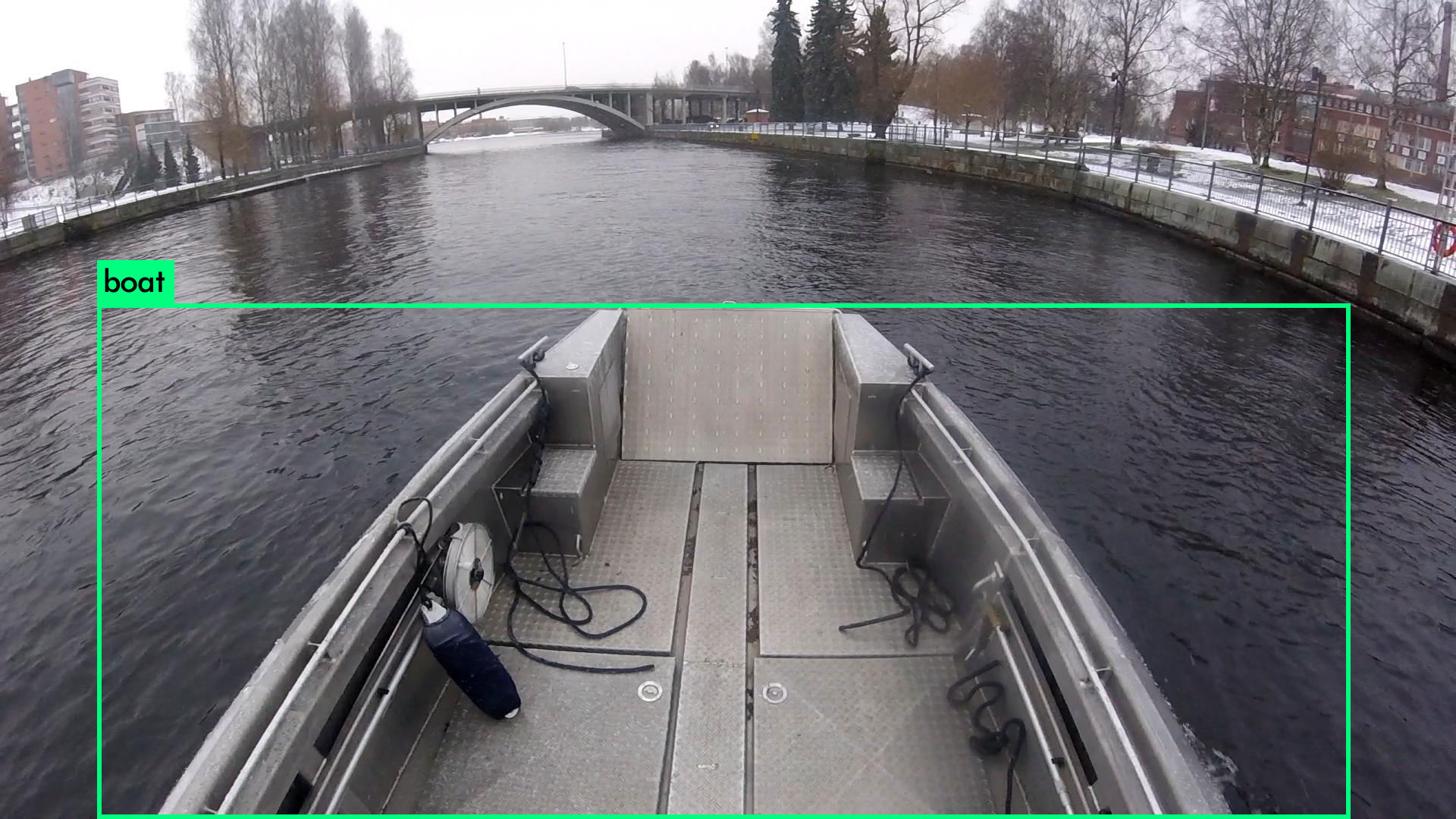}
        \caption{Object detection}
        \label{fig:yolo}
    \end{subfigure}
    \begin{subfigure}[t]{0.48\textwidth}
        \centering
        \includegraphics[width=\columnwidth]{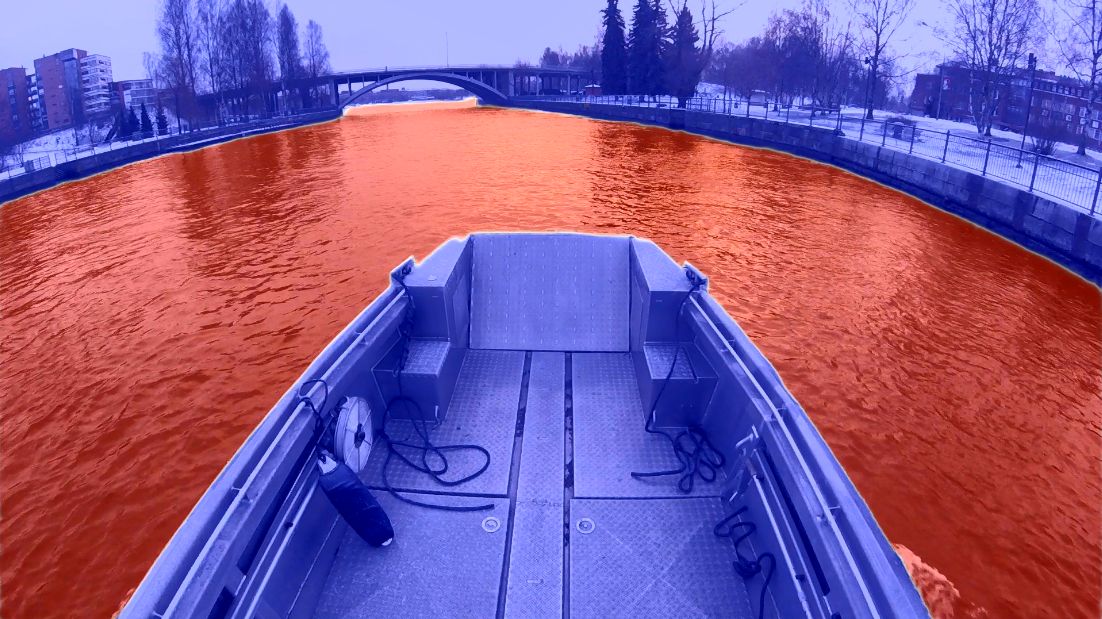}
        \caption{Image segmentation}
        \label{fig:watersegmentation}
    \end{subfigure}
    \caption{Examples of (a) an image detection algorithm, Yolov3~\cite{yolov3} with Darknet, which detects a boat with 91\% confidence, and (b) a water segmentation output~\cite{Taipalmaa-MLSP}.}
    \label{fig:detection_vs_segmentation}
\end{figure*}

\section{Single and Multi-agent Perception}
\label{sec:vision}


For autonomous robots meant to support SAR missions, it is essential to be able to quickly detect humans. For example, persons drowning at sea or lakes are in quick need of a rescue, but they are not easy to detect and the weather conditions can make the task even more difficult. In SAR operations, perception methods that take minutes or even multiple seconds to process an input frame may not be considered as viable solutions. This requirement for real-time processing speed sets restrictions for possible solutions. Furthermore, on self-operating agents, like UAVs, where it is possible to carry only a limited amount of equipment, it is necessary to either process data on the edge devices or use cloud offloading for more computational power, which both slow down the process. Currently, deep learning is one of the most studied fields in machine perception based on vision or other sensors. State-of-the-art deep learning models often lead to heavy and slow methods, but recent research has also focused towards the development of lighter and faster models able to operate in real-time with limited hardware resources. In~\cite{CV-preview}, the authors provide a broad overview of the progress of computer vision covering all sorts of emergencies. 

In this section, we discuss machine perception methods, focusing in SAR-like missions and environment. As mentioned in Section III, cameras are the most common sensors in SAR robotics and, therefore, we first concentrate on image-based perception, i.e., semantic segmentation and object detection. In semantic segmentation, everything that the agent perceives is labeled, and in object detection, only the objects of interest are labeled. The difference is illustrated in Fig.~\ref{fig:detection_vs_segmentation}. Semantic segmentation can be used to reduce the region of interest, i.e., if there is a person in water, semantic segmentation can reduce the area of interest to the area that contains water and object detection can be performed more efficiently for the smaller region. On the other hand, in semantic segmentation every pixel of an image is processed, so it is more time consuming than plain object detection. We also discuss computationally efficient models, which are critical for UAV applications but also for others robots. As there may be also other sensors, such as thermal cameras, GPS, and LiDAR, we also discuss multi-modal sensor fusion. Finally, we focus on specific the challenges in multi-agent perception. 

\subsection{Semantic Segmentation}
\label{subsec:segmentation}

Semantic segmentation is a process, where each pixel in an image is linked to a class label, such as sky, road, or forest. These pixels then form larger areas of adjacent pixels that are labeled with the same class label and recognized as objects. A survey on semantic segmentation using deep learning techniques available in~\cite{semantic-Survey} provides an extensive view of the methods provided to tackle this problem. In autonomous agents in general, the use of semantic segmentation has been studied fairly well in autonomous road vehicles. Siam et al.~\cite{Driving-Survey} have done an in-depth comparison of such semantic segmentation methods for autonomous driving and proposed a real-time segmentation benchmarking framework.

In marine environment, the study of semantic segmentation has been less common. In~\cite{Semantic-Benchmark}, three commonly used state-of-the-art deep learning semantic segmentation methods (U-Net~\cite{U-Net}, PSP-Net~\cite{MarineSS-13} and DeepLabv2~\cite{MarineSS-6}) are benchmarked on a maritime environment. The leaderboard for one of the largest publicly available datasets, Modd2 \cite{Modd2}, also contains a listing of semantic segmentation method capable to perform in marine environment~\cite{MarineSS-1, MarineSS-2, MarineSS-3, MarineSS-4, MarineSS-6, MarineSS-7, MarineSS-12, MarineSS-13}. 

In our former studies~\cite{Taipalmaa-MLSP, Taipalmaa-ICIP}, we have focused on semantic segmentation to separate water surface from everything else that appears in the image, which is similar to the process that is performed in self-driving cars for road detection. While excellent results can be obtained when the algorithm is applied in conditions that resemble the training images (see Fig~\ref{fig:watersegmentation}), it was observed the performance decreases notably in different conditions. This highlights the need of diverse training images as well as domain adaption techniques that help to adjust to unseen conditions \cite{zhang2017adaptation}. 

\subsection{Object Detection}
\label{subsec:object}

Object detection is a technique related to computer vision and image processing which deals with detecting instances of semantic objects of a certain class in digital images and videos. Object detectors can usually be divided into two categories: two-stage detectors and one-stage detectors. Two-stage detectors first propose candidate object bounding boxes,  and then features are extracted from each candidate box for the following classification and bounding-box regression tasks. The one-stage detectors propose predicted boxes from input images directly without region proposal step. Two-stage detectors have high localization and object recognition accuracy, while the one-stage detectors achieve high inference speed. A survey of deep learning based object detection~\cite{DetectionSurvey} has been published recently.   

Object detection tasks require high computing power and memory for real-time applications. Therefore, cloud computing~\cite{real-time} or small-sized object detection methods have been used for UAV applications~\cite{slimyolov3,skynet,dronet,Vaddi2019}. Cloud computing assists the system with high computing power and memory. However, communicating with a cloud server brings unpredictable delay from the network. In~\cite{real-time}, authors used cloud computing for object detection while keeping low-level object detection and navigation on the UAV.

Another option is to rely on specific object detection models~\cite{slimyolov3, skynet, dronet, Vaddi2019}, designed for limited computational power and memory. The papers proposed new object detection models, by using old detection models as their base structure and scaling the original network by reducing the number of filters or changing the layers and they achieved comparable detection accuracy besides the speed on real-time applications on drones. In~\cite{dronet}, authors observed a slight decrease on the accuracy while the new network was faster comparing to the old structure. In~\cite{tsaiSpatialSearch2019}, an adaptive submodularity and deep learning-based spatial search method for detecting humans with UAV in a 3D environment was proposed.

\subsection{Fast and computationally light methods}
\label{subsec:light}

As mentioned before, some solutions can be rather slow and computationally heavy, but in SAR operations it is vital that the used algorithms are as real-time as possible while still working with high level of confidence. The faster the algorithm can work, the faster the agent can search the area and that probably could lead to faster rescue of the persons in distress. Also the high confidence assures that no important information is missed.

Currently, you only look once (YOLO) is the state-of-the-art, real-time object detection system, and YOLOv3~\cite{yolov3} is stated to be extremely fast and accurate compared to methods like R-CNN~\cite{girshick14CVPR} and Fast R-CNN~\cite{girshickICCV15fastrcnn}. An example of the YOLOv3 output is shown in Fig.~\ref{fig:yolo}.

There is active research on methods that can produce more compact networks with improved prediction capability. Common approaches include knowledge distillation \cite{tung2019distillation}, where a compact student network is trained to mimic a larger network, e.g., by guiding the network to produce similar activations for similar inputs, and advanced network models, such as Operational Neural Networks \cite{kiranyaz2019ONN}, where the linear operators of CNNs are replaced by various (non-)linear operations, which allows to produce complex outputs which much fewer parameters.   

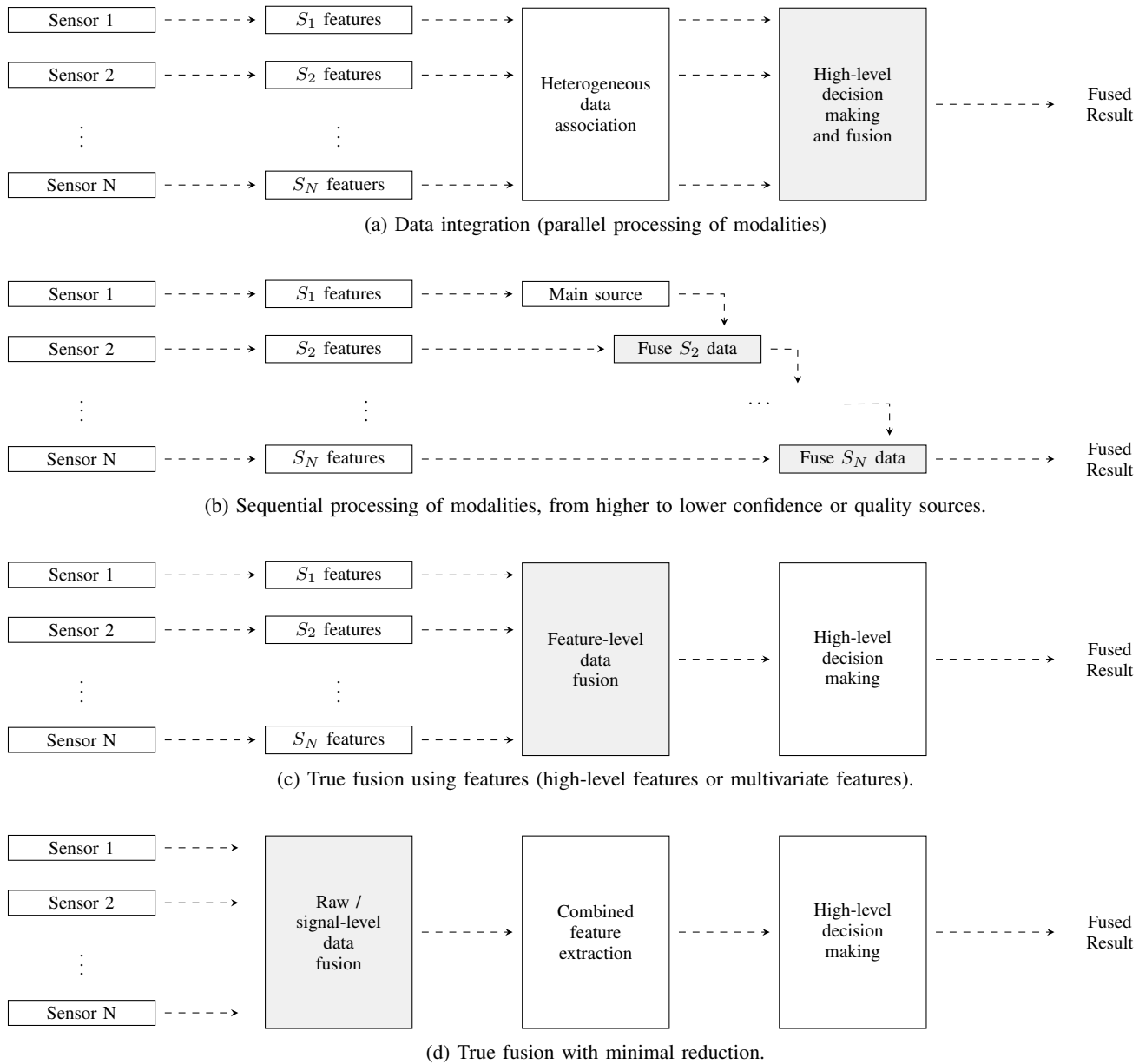
\begin{figure*}
    \centering
    \footnotesize
    \input{fig/data_fusion}
    \caption{Different multi-modal data fusion approaches: (a) parallel data integration with high-level decision making, (b) sequential processing of modalities when different modalities have difference conficende or quality levels, (c) true fusion with high-level features or with multivariate features, and (d) true fusion with minimal reduction~\cite{khan2019paradox, lahat2015multimodal}. In gray, we highlight the stage in which the fusion happens.}
    \label{fig:fusion_approaches}
\end{figure*}

\subsection{Multi-Modal Information Fusion}
\label{subsec:multimodal}

Multi-modal information fusion aims at combining data from a multiple sources, e.g., images and LiDAR. Information fusion techniques have been actively researched for decades and there is a myriad of different ways to approach the problem. The approaches can be roughly divided into techniques fusing information on raw data/input level, on feature/intermediate level, or on decision/output level~\cite{meng2020fusion}. An overview of the main data fusion approaches in multi-modal scenarios is illustrated in Fig.~\ref{fig:fusion_approaches}.

Some of the main challenges include \textit{representation}, i.e., how to represent multi-modal data taking into account complementarity and redundancy of multiple modalities, \textit{translation}, i.e., how to map the data from different modalities to a joint space, \textit{alignment}, i.e., how to understand the  relations of the elements of data from different modalities, for example, which parts of the data describe the same object in an image and in a point-cloud produced by LiDAR, \textit{fusion}, i.e., how to combine the information to form a prediction, and \textit{co-learning}, i.e., how to transfer knowledge between the modalities, which may be needed, for example, when one the modalities is not properly annotated~\cite{baltrusaitis2019multimodal}.  The main challenges related to multi-modal data are listed in Table~\ref{tab:challenges}.

In research years, also the information fusion techniques have focused more and more on big data and deep learning. Typical deep learning data fusion techniques have some layers specific to each data source and the features can be then combined before the final layers or processed separately all the way to the network output, while the representations are coordinated through a constraint such as a similarity distance~\cite{liu2020urban, baltrusaitis2019multimodal}. 

In SAR operations, the most relevant data fusion applications concern images and depth information~\cite{chen2020rgbd, wang2020environmental}. A recent deep learning based approach uses the initial image-based object detection results are to extract the corresponding depth information~\cite{wang2020environmental} and, thus, fuses the modalities on the output level. Another recent work proposed a multi-scale multi-path fusion network with that follows a two-stream fusion architecture with cross-modal interactions in multiple layers for coordinated representations~\cite{chen2020rgbd}. Simultaneous localization and mapping (SLAM) aims at constructing or updating a map of the environment of an agent, while simultaneously keeping track of the agent's position. In SLAM, RGB-D data is used to build a dense 3D map and the data fusion technique applied in a single-agent SLAM is typically extended Kalman Filter (EKF)~\cite{wang2018slam}. Fusing RGB and thermal image data can be needed, for example, in man overboard situations~\cite{katsamenis2020manoverboard}. Typically, there is much less training data available for thermal images and, therefore, domain adaption between RGB and thermal images may help~\cite{herrmann2018thermal}.

\begin{table*}
    \centering
    \renewcommand{\arraystretch}{1.42}
    \caption{Main challenges in multi-modal and multi-source data fusion}
    \label{tab:challenges}
    \begin{tabular}{@{}p{0.15\textwidth}p{0.8\textwidth}@{}}
        \toprule
        \textbf{Challenge} \hspace{2em} & \textbf{Description} \\
        \midrule
        Noisy data & Different data sources will suffer from different types and magnitudes of noise. A heterogeneous set of data sources naturally comes with heterogeneous sources of noise, from calibration errors to thermal noise. \\
        Unbalanced data & Having different data sources often involves data with different characteristics in terms of quality and confidence, but also in terms of spatial and temporal resolution. \\
        Conflicting data & Data from different sources might yield conflicting features. For example, in the case of autonomous robots, different types of sensors (visual sensors, laser rangefinders or radars) might detect obstacles at different distances. Missing data over a certain time interval from one of the sources might also affect the data fusion. \\
        \bottomrule
    \end{tabular}
\end{table*}

\subsection{Multi-Agent Perception}
\label{subsec:multi_perception}

To get the full benefit of the multi-robot approach in SAR operations, there should be also information fusion between the agents. For example, an object seen from two different angles can be recognized with a higher accuracy. The sensors carried by different robots may be the same, typically cameras, or different as the presence of multiple agents makes it possible to distribute some of the sensors' weight between the agents, which is important especially in UAV applications. The goal is that the perception the agents have of their environment is based on aggregating information from multiple sources and the agents share information steadily between themselves or to a control station.

The challenges and approaches are similar to those discussed in Section \ref{subsec:multimodal} for multi-modal information fusion, but the situation is further complicated by the fact that the data to be fused is located in different physical locations and the sensors are now moving with respect to each other. Some of the challenges that need to be solved are where to perform data fusion, how to evaluate whether different agents are observing the same objects or not, or how to rank observations from different agents. For many of the challenges, there are no efficient solutions yet.

There are several works concentrating on target tracking by multiple agents. These can be divided into four main categories: 1) Cooperative Tracking (CT), which aims tracking moving objects, 2) cooperative multi-robot observation of multiple moving targets (CMOMMTs), where the goal is to increase the total time of observation for all targets, 3) cooperative search,
acquisition, and tracking (CSAT), which alternates between the searching and tracking of moving targets, and 4) multi-robot pursuit evasion (MPE) \cite{khan2018cooperative, silva2019cooperative}. In SAR operations, especially CSAT approaches can be important after the victims have been initially located, for example, in marine SAR operations, where the victims are floating in the water. For initial search of the victims, a simulated cooperative approach using scanning laser range finders was proposed in \cite{zadorozhny2013collective}, but multi-view image fusion techniques for SAR operations are not yet operational.




%% file: fig/data_fusion.tex
\tikzstyle{rw} = [rectangle, minimum width=8em, minimum height=1.23em, draw=black, fill=white, align=center]
\tikzstyle{lrw} = [rectangle, minimum width=8em, minimum height=1.23em, draw=black, fill=white, align=center]
\tikzstyle{rrw} = [rectangle, minimum width=8em, minimum height=1.23em, draw=black, fill=white, align=center]
\tikzstyle{rempty} = [minimum width=8em, minimum height=1.23em, align=center]
\tikzstyle{empty} = [minimum width=8em, minimum height=1.23em, align=center]

\begin{subfigure}{\textwidth}
    \centering
    \begin{tikzpicture}[node distance=3em]
        
        %
        %
        \node (s1) [rw]                                     {Sensor 1};
        \node (s2) [rw, below of=s1]                        {Sensor 2};
        \node (sX) [empty, below of=s2]                     {$\vdots$};
        \node (sN) [rw, below of=sX]                        {Sensor N};
        
        \node (ft1) [rw, right of=s1, xshift=+11em]          {$S_1$ features};
        \node (ft2) [rw, right of=s2, xshift=+11em]          {$S_2$ features};
        \node (ftX) [empty, right of=sX, xshift=+11em]       {$\vdots$};
        \node (ftN) [rw, right of=sN, xshift=+11em]          {$S_N$ featuers};
        
    
        \node (integration) [   
                                rw, 
                                right of=ft1, 
                                xshift=+11em, 
                                minimum height=10.5em, 
                                yshift=-4.6em, 
                                minimum width=8em
                            ]  
                            {Heterogeneous \\ data \\ association};
                            
        \node (decision)    [
                                rw, 
                                right of=integration, 
                                xshift=+11em, 
                                minimum height=10.5em, 
                                minimum width=8em,
                                fill=white!94!black,
                            ]      
                            {High-level \\ decision \\ making \\ and fusion};
                            
        \node (fused)    [
                                empty, 
                                right of=decision, 
                                xshift=+11em, 
                                minimum height=10.5em, 
                                minimum width=8em
                            ]      
                            {Fused \\ Result};
        %
        %
        \draw [arrow]   ($(s1.east) + (0.5em, 0)$) -- ($(ft1.west) - (0.5em, 0)$);
        \draw [arrow]   ($(s2.east) + (0.5em, 0)$) -- ($(ft2.west) - (0.5em, 0)$);
        \draw [arrow]   ($(sN.east) + (0.5em, 0)$) -- ($(ftN.west) - (0.5em, 0)$);
        
        \draw [arrow]   ($(ft1.east) + (0.5em, 0)$) -- ($(ft1.east) + (5.5em, 0)$);
        \draw [arrow]   ($(ft2.east) + (0.5em, 0)$) -- ($(ft2.east) + (5.5em, 0)$);
        \draw [arrow]   ($(ftN.east) + (0.5em, 0)$) -- ($(ftN.east) + (5.5em, 0)$);
        
        \draw [arrow]   ($(ft1.east) + (14.5em, 0)$) -- ($(ft1.east) + (19.5em, 0)$);
        \draw [arrow]   ($(ft2.east) + (14.5em, 0)$) -- ($(ft2.east) + (19.5em, 0)$);
        \draw [arrow]   ($(ftN.east) + (14.5em, 0)$) -- ($(ftN.east) + (19.5em, 0)$);
        
        \draw [arrow]   ($(decision.east) + (0.5em, 0)$) -- ($(fused.west) + (1em, 0)$);
        
    \end{tikzpicture}
    \caption{Data integration (parallel processing of modalities)}
    \label{subfig:data_integration}
    \vspace{2.3em}
\end{subfigure}

\begin{subfigure}{\textwidth}
    \centering
    \begin{tikzpicture}[node distance=3em]
    
        %
        %
        \node (s1) [rw]                                     {Sensor 1};
        \node (s2) [rw, below of=s1]                        {Sensor 2};
        \node (sX) [empty, below of=s2]                     {$\vdots$};
        \node (sN) [rw, below of=sX]                        {Sensor N};
        
        \node (ft1) [rw, right of=s1, xshift=+11em]         {$S_1$ features};
        \node (ft2) [rw, right of=s2, xshift=+11em]         {$S_2$ features};
        \node (ftX) [empty, right of=sX, xshift=+11em]      {\hspace{3em}$\vdots$};
        \node (ftN) [rw, right of=sN, xshift=+11em]         {$S_N$ features};
        
        \node (res1) [rw, right of=ft1, xshift=+11em]                           {Main source};
        \node (res12) [rw, right of=ft2, xshift=+16em, fill=white!94!black]     {Fuse $S_2$ data};
        \node (resX) [empty, right of=ftX, xshift=+20em]                        {$\cdots$};
        \node (res12N) [rw, right of=ftN, xshift=+25em, fill=white!94!black]    {Fuse $S_N$ data};
    
        \node (fused)   [
                            empty, 
                            right of=res12N,
                            xshift=+11em,
                        ]      
                        {Fused \\ Result};

        %
        %
        \draw [arrow]   ($(s1.east) + (0.5em, 0)$) -- ($(ft1.west) - (0.5em, 0)$);
        \draw [arrow]   ($(s2.east) + (0.5em, 0)$) -- ($(ft2.west) - (0.5em, 0)$);
        \draw [arrow]   ($(sN.east) + (0.5em, 0)$) -- ($(ftN.west) - (0.5em, 0)$);
        
        \draw [arrow]   ($(ft1.east) + (0.5em, 0)$) -- ($(res1.west) - (0.5em, 0)$);
        \draw [arrow]   ($(ft2.east) + (0.5em, 0)$) -- ($(res12.west) - (0.5em, 0)$);
        \draw [arrow]   ($(ftN.east) + (0.5em, 0)$) -- ($(res12N.west) - (0.5em, 0)$);
        
        \draw [arrow]   ($(res1.east) + (0.5em, 0)$) -- ($(res12.north) + (2em, 2.23em)$) -- ($(res12.north) + (2em, 0.5em)$) ;
        \draw [arrow]   ($(res12.east) + (0.5em, 0)$) -- ($(resX.north) + (2em, 2.35em)$) -- ($(resX.north) + (2em, 0.5em)$) ;
        \draw [arrow]   ($(resX.east) + (0.5em, 0)$) -- ($(res12N.north) + (2em, 2.23em)$) -- ($(res12N.north) + (2em, 0.5em)$) ;
        
        \draw [arrow]   ($(res12N.east) + (0.5em, 0)$) -- ($(fused.west) + (1em, 0)$);
        
    \end{tikzpicture}
    \caption{Sequential processing of modalities, from higher to lower confidence or quality sources.}
    \label{subfig:sequential}
    \vspace{2.3em}
\end{subfigure}

\begin{subfigure}{\textwidth}
    \centering
    \begin{tikzpicture}[node distance=3em]
    
        %
        %
        \node (s1) [rw]                                     {Sensor 1};
        \node (s2) [rw, below of=s1]                        {Sensor 2};
        \node (sX) [empty, below of=s2]                     {$\vdots$};
        \node (sN) [rw, below of=sX]                        {Sensor N};
        
        \node (ft1) [lrw, right of=s1, xshift=+11em]       {$S_1$ features};
        \node (ft2) [lrw, right of=s2, xshift=+11em]       {$S_2$ features};
        \node (ftX) [empty, right of=sX, xshift=+11em]     {$\vdots$};
        \node (ftN) [lrw, right of=sN, xshift=+11em]       {$S_N$ features};
        
        \node (fusion) [   
                                rw, 
                                right of=ft1, 
                                xshift=+11em, 
                                minimum height=10.5em, 
                                yshift=-4.6em, 
                                minimum width=8em,
                                fill=white!94!black,
                            ]  
                            {Feature-level \\ data \\ fusion};
        
        \node (decision)    [
                                rw, 
                                right of=fusion, 
                                xshift=+11em, 
                                minimum height=10.5em, 
                                minimum width=8em,
                            ]      
                            {High-level \\ decision \\ making};
                            
        \node (final)    [
                                empty, 
                                right of=decision, 
                                xshift=+11em, 
                                minimum height=10.5em, 
                                minimum width=8em
                            ]      
                            {Fused \\ Result};
        
        %
        %
        \draw [arrow]   ($(s1.east) + (0.5em, 0)$) -- ($(ft1.west) - (0.5em, 0)$);
        \draw [arrow]   ($(s2.east) + (0.5em, 0)$) -- ($(ft2.west) - (0.5em, 0)$);
        \draw [arrow]   ($(sN.east) + (0.5em, 0)$) -- ($(ftN.west) - (0.5em, 0)$);
        
        \draw [arrow]   ($(ft1.east) + (0.5em, 0)$) -- ($(ft1.east) + (5.5em, 0)$);
        \draw [arrow]   ($(ft2.east) + (0.5em, 0)$) -- ($(ft2.east) + (5.5em, 0)$);
        \draw [arrow]   ($(ftN.east) + (0.5em, 0)$) -- ($(ftN.east) + (5.5em, 0)$);
        
        \draw [arrow]   ($(fusion.east) + (0.5em, 0)$) -- ($(decision.west) - (0.5em, 0)$);
        \draw [arrow]   ($(decision.east) + (0.5em, 0)$) -- ($(final.west) + (1em, 0)$);

    \end{tikzpicture}
    \caption{True fusion using features (high-level features or multivariate features).}
    \label{subfig:features}
    \vspace{2.3em}
\end{subfigure}

\begin{subfigure}{\textwidth}
    \centering
    \begin{tikzpicture}[node distance=3em]
    
        %
        %
        \node (s1) [rw]                                     {Sensor 1};
        \node (s2) [rw, below of=s1]                        {Sensor 2};
        \node (sX) [empty, below of=s2]                     {$\vdots$};
        \node (sN) [rw, below of=sX]                        {Sensor N};
        
        \node (fusion) [   
                                rw, 
                                right of=s1, 
                                xshift=+11em, 
                                minimum height=10.5em, 
                                yshift=-4.6em, 
                                minimum width=8em,
                                fill=white!94!black,
                            ]  
                            {Raw / \\signal-level \\ data \\ fusion};
                            
        \node (extraction)    [
                                rw, 
                                right of=fusion, 
                                xshift=+11em, 
                                minimum height=10.5em, 
                                minimum width=8em,
                            ]      
                            {Combined \\ feature \\ extraction};
        
        \node (decision)    [
                                rw, 
                                right of=extraction, 
                                xshift=+11em, 
                                minimum height=10.5em, 
                                minimum width=8em,
                            ]      
                            {High-level \\ decision \\ making};
                            
        \node (final)    [
                                empty, 
                                right of=decision, 
                                xshift=+11em, 
                                minimum height=10.5em, 
                                minimum width=8em
                            ]      
                            {Fused \\ Result};
        
        %
        %
        \draw [arrow]   ($(s1.east) + (0.5em, 0)$) -- ($(s1.east) + (4.5em, 0)$);
        \draw [arrow]   ($(s2.east) + (0.5em, 0)$) -- ($(s2.east) + (4.5em, 0)$);
        \draw [arrow]   ($(sN.east) + (0.5em, 0)$) -- ($(sN.east) + (4.5em, 0)$);
        
        \draw [arrow]   ($(fusion.east) + (0.5em, 0)$) -- ($(extraction.west) - (0.5em, 0)$);
        \draw [arrow]   ($(extraction.east) + (0.5em, 0)$) -- ($(decision.west) - (0.5em, 0)$);
        \draw [arrow]   ($(decision.east) + (0.5em, 0)$) -- ($(final.west) + (1em, 0)$);

    \end{tikzpicture}
    \caption{True fusion with minimal reduction.}
    \label{subfig:minimal}
    \vspace{1em}
\end{subfigure}

%% file: sections/05_ActivePerception.tex
\section{Closing the loop: \\ Active Perception in Multi-Robot Systems}\label{sec:active_perception}

While we above discussed coverage planning, formation control, and perception aspects of SAR as separate operations, it is obvious that all the components need to function seamlessly together in order achieve optimal performance. This means that coverage planning and formation control need to be adjusted based on the observations and the perception algorithms need to be optimized to support and take full advantage of overall adaptive multi-agent systems. This can be achieved via active perception techniques~\cite{bajcsy1988active, gallos2019active}. While the passive perception techniques simply utilize whatever inputs they are given, active perception methods adapt the behavior of the agent(s) in order to obtain better inputs. 

Active perception has been defined as: 
\begin{quotation}{An agent is an active perceiver if it knows \textit{why} it wishes to sense, and then chooses \textit{what} to perceive, and determines \textit{how}, \textit{when}, and \textit{where} to achieve that perception. \cite{bajcsy2018active}}\end{quotation} In the case of searching a victim, this can mean that the robots are aware that the main purpose is to save humans (why), and are able adapt their actions to achieve better sightings of people in need of help (what) by, for example, zooming the camera to a potential observation (how) or by moving to a position that allows a better view (where and when). 

In a SAR operation, active perception can help in multiple subtasks in the search for victims, such as path finding in complex environments~\cite{falanga2017gaps}, obstacle avoidance~\cite{chessa2014obstacle}, or target detection~\cite{sandino2020target}. Once a victim has been detected, it is also important to keep following him/her. For instance, in maritime SAR operations, there is a high probability that the survivors are floating in the sea and drifting due to the wind or marine currents. In such scenarios, it is essential that the robots are able to continuously update the position of survivors so that path planning for the rescue vessel can be re-optimized and recalculated in real-time in an autonomous manner. This requires active tracking of the target~\cite{zhong2018tracking}. 

While our main interest lies in active perception for multi-robot SAR operations, the literature directly focusing on this specific field is still scarce. Nevertheless, active perception is a rapidly developing research topic and we believe that it will be one of the key elements also in the future research on multi-robot SAR operations and will utilize the state-of-the-art techniques developed on related applications. Therefore, we will start by introducing the main ideas presented in single-agent active perception and then turn our attention on works that consider active perception in formation control and multi-robot planning.  The essence of active perception is understanding, adapting to changes in the environment and taking action for the next mission. This adaptation can happen in the sensors as well as in deciding a possible future mission.

\subsection{Single-Agent Active Perception}

Besides performing their main task (e.g., object detection), active perception algorithms use the same input data to predict the the next action that can help them to improve their performance. This is a challenge for training data collection, because typically there is high number of possible actions in any given situation and it is not always straightforward to decide which actions would be good or bad. A benchmark dataset~\cite{ammirato2017data} provides 9000 real indoor input images along with the information showing what would be seen next if a specific action is carried out when a specific image is seen. Another possibility is to create simulated training environments~\cite{tzimas2020simulator}, where actions can be taken in a more natural manner. With such simulators, it is critical that the simulator is realistic enough so that employment in the real world is possible. To facilitate the transition, \textit{Sim2Real} learning methods can be used~\cite{sadeghi2018sim2real}. Finally, it is also possible to use real equipment and environments~\cite{falanga2017gaps, calli2018realtraining}, but such training is slow and requires having access to suitable equipment. Therefore, training setups are typically simplistic. Furthermore, real-world training makes it more complicated to compare different approaches.

Currently, the most active research direction in active perception is reinforcement learning~\cite{gallos2019active}. Instead of learning from labeled input-output pairs, reinforcement learning is based on rewards and punishment given to the agents based on their actions. Robots can start by some random actions and gradually, via rewards and/or punishments, they learn to follow desired behavior. A critical question for the training of reinforcement learning methods is the selection of the reward function. In object detection, the agents are typically rewarded when they manage to reduce the uncertainty of the detection~\cite{calli2018realtraining, mathe2016detection}. In active tracking, the reward can use a desired distance and viewing angle~\cite{tzimas2020simulator}. In SAR operations, it must be taken into account that due to changes in the environment, e.g., occlusion, tracking can be temporarily lost and the person must be redetected. 

While reinforcement learning is expected to be the future direction is active perception, its applicability in SAR operations is reduced by the problems of collecting or creating sufficient training data and experiences. Therefore, simpler approaches that use deep neural networks only for visual data analysis but use traditional approaches, such as proportional-integral-derivative (PID) controllers~\cite{andersson2017pid}, for control may be currently easier to implement. A way to use active perception in a simulated setting of searching a lost child indoors using a single UAV is described in~\cite{sandino2020target}. The paper presents an autonomous Sequential Decision Process (SDP) to control UAV navigation under uncertainty as a multi-objective problem comprising path planning, obstacle avoidance, motion control, and target detection tasks. Also in this work, one of the goals is to reduce target detection uncertainty in deep learning object detectors by encouraging the UAV to fly closer to the detected object and to adjust its orientations to get better pictures.


\subsection{Active Perception and Formation Control}

In previous sections, we have described the importance of formation control algorithms as the backbone of multi-robot collaborative sensing. The literature describing the integration of formation control algorithms with active perception for collaborative search or tracking is sparse, and no direct applications to SAR robotics have been published, to the best of our knowledge. Therefore, we describe here the most relevant works and discuss their potential in SAR operations.

Active perception for formation control algorithms has been mostly studied under the problem of multi-robot collaborative active tracking. Early works in this area include~\cite{zhou2011multirobot}, where the authors explored the generation of optimal trajectories for a heterogeneous multi-robot system to actively and collaboratively track a mobile target. This particular work has been a reference of a significant amount of research in the field for the past decade. Some of the assumptions, nonetheless, limited the applicability of these early results, such as the need for accurate global localization of all sensors in the multi-robot system. 

A work on the combination of cooperative tracking together with formation control algorithms for multi-robot systems was introduced in~\cite{ahmad2013perception}. The authors proposed a perception-driven formation control algorithms that aimed at maximizing the performance of multi-robot collaborative perception of a tracked subject through a non-linear model predictive control (MPC) strategy. One of the key contributions of this work compared to previous literature was that the same strategy could be easily adapted to different objectives: optimization of target perception, collision avoidance, or formation maintenance, by adapting the weights of the different parts in the MPC formulation. Moreover, the authors show that their approach can be utilized within multi-robot systems with variable dynamics. However, all robots are assumed to be holonomic, and the integration of heterogeneous systems with non-holonomic robots was left for future work.

In a similar research direction, Tallamraju et al. described in a recent work a formation control algorithm for active multi-UAV tracking based on MPC~\cite{tallamraju2019active}. One of the main novelties of this work is that the MPC is built from decoupling the minimization of the tracking error (distance from the UAVs to the person) and the minimization of the formation error (constraints on the relative bearing of the UAVs with respect to the tracked person). Another key novelty is that the authors incorporated collision avoidance within the main control loop, avoiding non-convexity in the optimization problem by calculating first the collision avoidance constraints and adding them as control inputs to the MPC formulation.

In more practical terms, the results of~\cite{tallamraju2019active} enable online calculation of collision-free path planning while tracking a movable subject and maintaining a certain formation configuration around the tracked subject, optimizing the estimation of the object's position during tracking and maintaining it close to the center of the field of view of each of the robots deployed for collaborative tracking. Compared to other recent works, the authors are able to obtain the best accuracy in the estimation of the tracked person's position, while only trading off a negligible increase in error of the self-localization estimation of each of the tracking robots.

A more general result, with no direct linkage to SAR operations, is ActiveMoCap~\cite{kiciroglu2020activemocap}. The authors presented a system for tracking a person while optimizing the relative and global three-dimensional human pose by choosing the position with the least pose uncertainty. This can be applied in a variety of scenarios, where the viewpoint of an autonomous or robot needs to be optimized to improve 3D human pose estimation.

In terms of leader-follower formation control, Chen et al. presented an active vision approach for non-holonomic robotic agents~\cite{chen2015adaptive}. While these types of method have no direct applicability to SAR operations, they can be employed to improve the coordination of robots by improving the perception that each robot has of its collaborating peers in the different multi-robot applications that have been described throughout this survey.

Time-varying sensing topologies in multi-robot active tracking were considered by Zhang et al. in~\cite{zhang2019optimized}. The authors consider multi-robot systems with a single leader and multiple followers able of only range measurements. The authors acknowledge that introducing time-varying perception topology significantly increases the difficulty of the optimization problem, and future works need to present novel ideas to solve these issues for more realistic applications.

\subsection{Perception Feedback in Multi-Robot Planning and Multi-Robot Search}

Other works in cooperative active tracking and cooperative active localization, have been presented without necessarily considering spatial coordination of fixed formations among the collaborative robots. In~\cite{morbidi2012active}, active perception was incorporated in a collaborative multi-robot tracking application by planning the paths to minimize the uncertainty in the location of both each individual robot and the target. The robots were UAVs equipped with lidar sensors. In~\cite{gurcuoglu2013hierarchical}, the authors extend the previous work towards incorporating the dynamics of the UAVs in the position estimators, as well as perform real-world experiments. In this second work, a hierarchical control approach was utilized to generate the paths for the different robots.

An extensive description of methods for (i) localization of a stationary target with one and many robots, (ii) active localization of clusters of targets, (iii) guaranteed localization of multiple targets, and (iv) tracking adversarial targets, is presented in~\cite{vander2015active}. The different methods incorporate both active perception and active localization approaches, and they are mainly focused at ranging measurements based on wireless signals. In terms of SAR robotics and the different systems described in this survey, these type of methods have the most potential in avalanche events for locating ATs, or in other scenarios if the victims have known devices emitting some sort of wireless signal.

In the area of multi-robot search, Acevedo et al. recently presented a cooperative multi-robot search algorithm based on a particle filter and active perception~\cite{acevedo2020dynamic}. The approach presented in that paper can be exported to SAR scenarios, as the authors focus on optimizing the collaborative search by actively maximizing the information that robots acquire of the search area. One of the most significant contributions within the scope of this survey is that the authors work on the assumption of uncertainty in the data, and therefore propose the particle filter for active collaborative perception. This results in a dynamic reallocation of the robots to different search areas. The system, while mostly distributed, requires the robots to communicate with each other to maintain a common copy of the particle filter. The authors claim that future works will be directed towards further decentralizing the algorithms by enabling asynchronous communication and local particle filters at each of the robots.

In between the areas of multi-robot active coverage and active tracking and localization, Tokekat and Vander et al. have presented methods for localizing and monitoring radio-tagged invasive fish with an autonomous USV~\cite{tokekar2012coverage, vander2013local}. Other authors have presented methods for actively acquiring information about the environment. For instance, a significant work in this area that has direct application to the initial assessment and posterior monitoring of the area in SAR scenarios is~\cite{atanasov2015decentralized}, where the authors present a decentralized multi-robot simultaneous localization and mapping (SLAM) algorithm. The authors identified that optimal path planning algorithms maximizing active perception had a computational complexity that would grow exponentially with both the number of sensors and the planning horizon. To address this issue, they proposed an approximation algorithm and a decentralized implementation with only linear complexity demonstrating good performance in multi-robot SLAM.

A more general approach to collaborative active sensing was presented in~\cite{schlotfeldt2019maximum}, where the authors proposed a method for planning multi-robot trajectories. This approach could be applied to different tasks including active mapping with both static and dynamic targets, as well as mapping environments with obstacles.

%% file: sections/06_Discussion.tex
\section{Discussion and Open Research Questions} \label{sec:discussion}

Research efforts have mainly focused on the design of individual robots autonomously operating in emergency scenarios, such as those presented in the European Robotics League Emergency Tournament. Most of the existing literature in multi-robot systems for SAR either relies on an external control center for route planning and monitoring, on a static base station and predefined patterns for finding objectives, or have predefined interactions between different robotic units. Therefore, there is a big potential to be unlocked throughout a wider adoption of distributed multi-robot systems. Key advances will require embedding more intelligence in the robots with lightweight deep learning perception models, the design and development of novel distributed control techniques, as well as the closer integration of perception and control algorithms. Moreover, heterogeneous multi-robot systems have shown significant benefits when compared to homogeneous systems. In that area, nonetheless, further research needs to focus on interoperability and ad-hoc deployments of multi-robot systems.

Based on the different aspects of multi-robot SAR that have been described in this survey, both at the system level and from the coordination and perception perspectives, we have summarized the main research directions where we see the greatest potential. Further development in these areas is required to advance towards a wider adoption of multi-robot SAR systems.

\subsection{Shared Autonomy}

With the increasing adoption of multi-robot systems for SAR operations over individual and complex robots, the number of degrees of freedom that can be controlled has risen dramatically. To enable efficient SAR support from these systems without the need for a large number of SAR personnel controlling or supervising the robots, the concept of shared autonomy needs to be further explored.

The applications of more efficient shared autonomy and control interfaces are multiple. For instance, groups of UAVs flying in different formation configurations could provide real-time imagery and other sensor information from a large area after merging the data from all the units. In that scenario, the SAR personnel controlling the multi-UAV system would only need to specify the formation configuration and control the whole system as a single UAV would be controlled in a more traditional setting.

While some of the directions towards designing control interfaces for scalable homogeneous multi-robot systems are relatively clear, further research needs to be carried out in terms of conceptualization and design of interfaces for controlling heterogeneous robots. These include land-air systems (UGV+UAV), sea-land systems (USV+UAV), and also surface-underwater systems (USV+UUV), among other possibilities. In these cases, owing to the variability of their operational capabilities and significant differences in the robots dynamics and degrees of freedom, a shared autonomy strategy is not straightforward.

\subsection{Operational Environments}

Some of the main open research questions and opportunities that we see for each of the scenarios described in this paper in terms of deployment of multi-robot SAR systems are the following:
\begin{itemize}
    \item {Urban SAR:} we have described the various types of ground robots being utilized in USAR scenarios, as well as collaborative UGV+UAV systems. In this area, we see the main opportunities and open challenges to be in (i) collaborative localization in GNSS denied environments; (ii) collaborative perception of victims from different perspectives; (iii) ability to perform remote triage and establish a communication link between SAR personnel and victims, or to transport medicines and food; and (iv) more scalable heterogeneous systems with various sizes of robots (both UGVs and UAVs) capable to collaboratively mapping and monitoring harsh environments or post-disaster scenarios.
    \item {Marine SAR:} throughout this survey, we have seen that marine SAR operations are one of the scenarios where heterogeneous multi-robot systems have been most widely adopted. Nonetheless, there are multiple challenges remaining in terms of interoperability and deployability. In particular, few works have explored the potential in closely designing perception and control strategies for collaborative multi-robot systems including underwater, surface and aerial robots~\cite{escusol2017autonomous}. Moreover, while the degree of autonomy of UAVs and UUVs has advanced considerably in recent years, USVs can benefit from the data gathered by these to increase their autonomy. In terms of deployability, more robust solutions are needed for autonomous take-off and docking of UAVs or UUVs from surface robots. Finally, owing to the large areas in which search for victims takes place in maritime SAR operations, active perception approaches increasing the efficiency of search tasks have the most potential in these environments.
    \item {Wilderness SAR:} some of the most important challenges in WiSAR operations are the potentially remote and unexplored environments posing challenges to both communication and perception. Therefore, an essential step towards more efficient multi-robot operations in WiSAR scenarios is to increase the level of autonomy as well as the operational time of the robots. Long-term autonomy and embedded intelligence on the robots for decision-making without human supervision are some of the key research directions in this area in terms of multi-robot systems.
\end{itemize}

\subsection{Sim-to-real Methods for Deep Learning}
\label{subsec:sim2real}

Deep-learning-based methods are flexible and can be adapted to a wide variety of applications and scenarios. Good performance, however, comes at the cost of enough training data and an efficient training process that is carried out offline. Other deep learning methods, and particularly deep reinforcement learning (DRL), rely heavily on simulation environments for converging towards working control policies or stable inference, with training happening on a trial-and-error basis. Search and rescue robots are meant to be deployed in real scenarios where the conditions can be more challenging than those of more traditional robots. Therefore, an important aspect to take into account is the transferability of the models trained in simulation to the reality.

Recent years have seen an increasing research interest in closing the gap between simulation and reality in DRL~\cite{arndt2019meta}. In the field of SAR robotics, a relevant example of the utilization of both DL and DRL techniques was presented by Sampedro et al.~\cite{sampedro2019fully}. The authors developed a fully autonomous aerial robot for USAR operations in which a CNN was trained to for target-background segmentation, while reinforcement learning was utilized for vision-based control methods. Most of the training happened with a Gazebo simulation and ROS, and the method was tested also in real indoor cluttered environments. In general, and compared with other DL methods, DRL has the advantage in that it can be used to provide an end-to-end model from sensing to actuation, therefore integrating the perception and control aspects within a single model. Other recent applications of DRL for SAR robotics include the work of Niroui et al.~\cite{niroui2019deep}, with an approach to navigation in complex and unknown USAR cluttered environments that used DRL for frontier exploration. In this case, the authors put an emphasis on the efficiency of the simulation-to-reality transfer. Another recent work by Li et al.~\cite{li2020deep} showed the versatility of DRL for autonomous exploration and the ability of transferring the model from simulation to reality in unknown environments. We discuss the role of DRL in active perception in Section~\ref{sec:active_perception}. Bridging the gap between simulation and reality is thus another challenge in some of the current SAR robotic systems.

\subsection{Human Condition Awareness and Triage}

As we have discussed in multiple occasions throughout this survey, the current applicability of SAR robotics is mainly in the search of victims or the assessment and monitoring of the area by autonomously mapping and analyzing the accident or disaster scenario. However, only a relatively small amount of works in multi-robot SAR robotics have been paying attention to the development of methods for increasing the awareness of the status of the victims in the area or performing remote triage. 

The potential for lifesaving applications in this area is significant. The design and development of methods for robots to be able to better understand the conditions of survivors after an accident is therefore a research topic with multiple open questions and challenges. Nonetheless, it is important to take into account that this most likely requires the robots to reach to the victims or navigate near them. The control of the robot and its awareness of its localization and environment thus need to be very accurate, as otherwise operating in such safety-critical scenario might be counterproductive. Therefore, before being able to deploy in a real scenario novel techniques for human condition awareness and remote triage, the robustness of navigation and localization methods in such environments needs to be significantly streamlined.

\subsection{Heterogeneous Multi-Robot Systems}

Across the different types of SAR missions that have been discussed in this survey, the literature regarding the utilization of heterogeneous robots has shown the clear benefits of combining either different types of sensors, different perspectives, or different computational or operational capabilities. Nonetheless, most of the existing literature assumes that the identity and nature of the robots is known a priori, as well as the way in which they communicate and share data. A wider adoption and deployment of heterogeneous multi-robot systems therefore needs research to advance in the following practical areas:

\begin{itemize}
    \item {Interoperability:} flexible deployment of a variable type and number of robots for SAR missions requires the collaborative methods to be designed with wider interoperability in mind. Interoperability has been the focus of both the ICARUS and DARIUS projects~\cite{serrano2015icarus, cubber2017introduction}. Moreover, extensive research has been carried out in interoperable communication systems, and current robotic middlewares, such as ROS2~\cite{maruyama2016exploring}, enable distributed robotic systems to share data and instructions with standard data types. Nonetheless, there is still a lack of interoperability in terms of high-level planning and coordination for specific missions. In SAR robotics, these include collaborative search and collaborative mapping and perception.
    \item {Ad-hoc systems:} closely related to the concept of interoperability in terms of high-level planning, wider adoption of multi-robot SAR systems requires these systems to be deployed in an ad-hoc manner, where the type or number of robots does not need to be predefined. This has been explored, to some extent, in works utilizing online planning strategies that account for the possibility of malfunctioning or missing robots~\cite{maza2007multiple}.
    \item {Situational awareness and awareness of other robots:} the wide variety of robots being utilized in SAR missions, and the different scenarios in which they can be applied, calls for the abstraction and definition of models defining these scenarios but also the way in which robots can operate with them. In heterogeneous multi-robot systems, distributed high-level collaborative planning requires robots to understand not only how can they operate in their current environment and what are the main limitations or constraints, but also those conditions of different robots operating in the same environment. For instance, a USV collaborating with other USVs and UAVs in a maritime SAR mission needs to be aware of the different perspectives that UAVs can bring into the scene, but also of their limitations in terms of operational time or weather conditions.
\end{itemize}

\subsection{Active Perception}

We have closed this survey exploring the literature in active perception for multi-robot systems, where we have seen a clear lack of research within the SAR robotics domain. Current approaches for area coverage in SAR missions, for instance, mostly consider an a priori partition of the area among the available robots. Dynamic or online area partitioning algorithms are only considered either in the presence of obstacles, or when the number of robots changes~\cite{maza2007multiple}. Other works also consider an a priori estimation of the probability of locating victims across different areas to optimize the path planning~\cite{le2012adaptive, schwager2009decentralized}. These and other works are all based in either a priori-knowledge of the area, or otherwise partition the search space in a mostly homogeneous manner. Therefore, there is an evident need for more efficient multi-robot search strategies 

Active perception can be merged into current multi-robot SAR systems in multiple directions: actively updating and estimating the probabilities of victims’ locations, but also with active SLAM techniques by identifying the most severely affected areas in post-disaster scenarios. In wilderness and maritime search and rescue where tracking of the victims might be necessary even after they have been found, active perception has the potential to significantly decrease the probability of missing a target.

In general, we also see the potential of active perception within the concepts of human-robot and human-swarm cooperation, as well as in terms of increasing the awareness that robots have of victims’ conditions. Regarding human-robot and human-swarm cooperation, active perception can bring important advantages in the understanding the actions of SAR personnel and being able to provide more relevant support during the missions.

%% file: sections/07_Conclusion.tex
\section{Conclusion}\label{sec:conclusion}

Among the different civil applications where multi-robot systems can be deployed, search and rescue (SAR) operations are one of the fields where the impact can be most significant. In this survey, we have reviewed the status of SAR robotics with a special focus on multi-robot SAR systems. While SAR robots have been a topic of increasing research attention for over two decades, the design and deployment of multi-robot systems for real-world SAR missions has only been effective more recently. Multiple challenges remain at the system-level (interoperability, design of more robust robots, and deployment of heterogeneous multi-robot systems, among others), as well as from the algorithmic point of view of multi-agent control and multi-agent perception. This is the first survey, to the best of our knowledge, to analyze these two different points of view complementing the system-level view that other surveys have given. Moreover, this work differentiates from others in its discussion of both heterogeneous systems and active perception techniques that can be applied to multi-robot SAR systems. Finally, we have listed the main open research questions in these directions.

%% file: main.bbl
\begin{thebibliography}{100}
\providecommand{\url}[1]{#1}
\csname url@samestyle\endcsname
\providecommand{\newblock}{\relax}
\providecommand{\bibinfo}[2]{#2}
\providecommand{\BIBentrySTDinterwordspacing}{\spaceskip=0pt\relax}
\providecommand{\BIBentryALTinterwordstretchfactor}{4}
\providecommand{\BIBentryALTinterwordspacing}{\spaceskip=\fontdimen2\font plus
\BIBentryALTinterwordstretchfactor\fontdimen3\font minus
  \fontdimen4\font\relax}
\providecommand{\BIBforeignlanguage}[2]{{%
\expandafter\ifx\csname l@#1\endcsname\relax
\typeout{** WARNING: IEEEtran.bst: No hyphenation pattern has been}%
\typeout{** loaded for the language `#1'. Using the pattern for}%
\typeout{** the default language instead.}%
\else
\language=\csname l@#1\endcsname
\fi
#2}}
\providecommand{\BIBdecl}{\relax}
\BIBdecl

\bibitem{INGRAND201710}
F.~Ingrand and M.~Ghallab, ``Deliberation for autonomous robots: A survey,''
  \emph{Artificial Intelligence}, vol. 247, pp. 10 -- 44, 2017, special Issue
  on AI and Robotics.

\bibitem{deng2018survey}
C.~Deng, G.~Liu, and F.~Qu, ``Survey of important issues in multi unmanned
  aerial vehicles imaging system,'' 2018.

\bibitem{shakhatreh2019unmanned}
H.~Shakhatreh, A.~H. Sawalmeh, A.~Al-Fuqaha, Z.~Dou, E.~Almaita, I.~Khalil,
  N.~S. Othman, A.~Khreishah, and M.~Guizani, ``Unmanned aerial vehicles
  (uavs): A survey on civil applications and key research challenges,''
  \emph{IEEE Access}, vol.~7, pp. 48\,572--48\,634, 2019.

\bibitem{mehmood2018multi}
S.~Mehmood, S.~Ahmed, A.~S. Kristensen, and D.~Ahsan, ``Multi criteria decision
  analysis (mcda) of unmanned aerial vehicles (uavs) as a part of standard
  response to emergencies,'' in \emph{4th International Conference on Green
  Computing and Engineering Technologies; Niels Bohrs Vej 8, Esbjerg, Denmark},
  2018.

\bibitem{roberts2016unmanned}
W.~Roberts, K.~Griendling, A.~Gray, and D.~Mavris, ``Unmanned vehicle
  collaboration research environment for maritime search and rescue,'' in
  \emph{30th Congress of the International Council of the Aeronautical
  Sciences}.\hskip 1em plus 0.5em minus 0.4em\relax International Council of
  the Aeronautical Sciences (ICAS) Bonn, Germany, 2016.

\bibitem{luk2005intelligent}
B.~L. Luk, D.~S. Cooke, S.~Galt, A.~A. Collie, and S.~Chen, ``Intelligent
  legged climbing service robot for remote maintenance applications in
  hazardous environments,'' \emph{Robotics and Autonomous Systems}, vol.~53,
  no.~2, pp. 142--152, 2005.

\bibitem{lunghi2019multimodal}
G.~Lunghi, R.~Marin, M.~Di~Castro, A.~Masi, and P.~J. Sanz, ``Multimodal
  human-robot interface for accessible remote robotic interventions in
  hazardous environments,'' \emph{IEEE Access}, vol.~7, pp. 127\,290--127\,319,
  2019.

\bibitem{sung2019multi}
Y.~Sung, ``Multi-robot coordination for hazardous environmental monitoring,''
  Ph.D. dissertation, Virginia Tech, 2019.

\bibitem{merino2005cooperative}
L.~Merino, F.~Caballero, J.~Martinez-de Dios, and A.~Ollero, ``Cooperative fire
  detection using unmanned aerial vehicles,'' in \emph{Proceedings of the 2005
  IEEE international conference on robotics and automation}.\hskip 1em plus
  0.5em minus 0.4em\relax IEEE, 2005, pp. 1884--1889.

\bibitem{brenner2017new}
S.~Brenner, S.~Gelfert, and H.~Rust, ``New approach in 3d mapping and
  localization for search and rescue missions,'' \emph{CERC2017}, p. 105, 2017.

\bibitem{hayat2016survey}
S.~Hayat, E.~Yanmaz, and R.~Muzaffar, ``Survey on unmanned aerial vehicle
  networks for civil applications: A communications viewpoint,'' \emph{IEEE
  Communications Surveys \& Tutorials}, vol.~18, no.~4, pp. 2624--2661, 2016.

\bibitem{grogan2018use}
S.~Grogan, R.~Pellerin, and M.~Gamache, ``The use of unmanned aerial vehicles
  and drones in search and rescue operations--a survey,'' \emph{Proceedings of
  the PROLOG}, 2018.

\bibitem{grayson2014search}
S.~Grayson, ``Search \& rescue using multi-robot systems,'' \emph{School of
  Computer Science and Informatics, University College Dublin}, 2014.

\bibitem{queralta2020autosos}
J.~{Pe\~{n}a Queralta}, J.~Raitoharju, T.~N. Gia, N.~Passalis, and T.~Westerlund,
  ``Autosos: Towards multi-uav systems supporting maritime search and rescue
  with lightweight ai and edge computing,'' \emph{arXiv preprint
  arXiv:2005.03409}, 2020.

\bibitem{nazarova2020application}
A.~V. Nazarova and M.~Zhai, ``The application of multi-agent robotic systems
  for earthquake rescue,'' in \emph{Robotics: Industry 4.0 Issues \& New
  Intelligent Control Paradigms}.\hskip 1em plus 0.5em minus 0.4em\relax
  Springer, 2020, pp. 133--146.

\bibitem{klamt2019flexible}
T.~Klamt, D.~Rodriguez, L.~Baccelliere, X.~Chen, D.~Chiaradia, T.~Cichon,
  M.~Gabardi, P.~Guria, K.~Holmquist, M.~Kamedula \emph{et~al.}, ``Flexible
  disaster response of tomorrow: Final presentation and evaluation of the
  centauro system,'' \emph{IEEE Robotics \& Automation Magazine}, vol.~26,
  no.~4, pp. 59--72, 2019.

\bibitem{ollero2005multiple}
A.~Ollero, S.~Lacroix, L.~Merino, J.~Gancet, J.~Wiklund, V.~Remu{\ss}, I.~V.
  Perez, L.~G. Guti{\'e}rrez, D.~X. Viegas, M.~A.~G. Benitez \emph{et~al.},
  ``Multiple eyes in the skies: architecture and perception issues in the
  comets unmanned air vehicles project,'' \emph{IEEE robotics \& automation
  magazine}, vol.~12, no.~2, pp. 46--57, 2005.

\bibitem{kruijff2014designing}
G.-J.~M. Kruijff, I.~Kruijff-Korbayov{\'a}, S.~Keshavdas, B.~Larochelle,
  M.~Jan{\'\i}{\v{c}}ek, F.~Colas, M.~Liu, F.~Pomerleau, R.~Siegwart, M.~A.
  Neerincx \emph{et~al.}, ``Designing, developing, and deploying systems to
  support human--robot teams in disaster response,'' \emph{Advanced Robotics},
  vol.~28, no.~23, pp. 1547--1570, 2014.

\bibitem{de2018persistent}
J.~De~Greeff, T.~Mioch, W.~Van~Vught, K.~Hindriks, M.~A. Neerincx, and
  I.~Kruijff-Korbayov{\'a}, ``Persistent robot-assisted disaster response,'' in
  \emph{Companion of the 2018 ACM/IEEE International Conference on Human-Robot
  Interaction}, 2018, pp. 99--100.

\bibitem{cubber2017search}
G.~D. Cubber, D.~Doroftei, K.~Rudin, K.~Berns, D.~Serrano, J.~Sanchez,
  S.~Govindaraj, J.~Bedkowski, and R.~Roda, ``Search and rescue robotics-from
  theory to practice,'' 2017.

\bibitem{matos2016multiple}
A.~Matos, A.~Martins, A.~Dias, B.~Ferreira, J.~M. Almeida, H.~Ferreira,
  G.~Amaral, A.~Figueiredo, R.~Almeida, and F.~Silva, ``Multiple robot
  operations for maritime search and rescue in eurathlon 2015 competition,'' in
  \emph{OCEANS 2016-Shanghai}.\hskip 1em plus 0.5em minus 0.4em\relax IEEE,
  2016, pp. 1--7.

\bibitem{gawel2018x}
A.~Gawel, C.~Del~Don, R.~Siegwart, J.~Nieto, and C.~Cadena, ``X-view:
  Graph-based semantic multi-view localization,'' \emph{IEEE Robotics and
  Automation Letters}, vol.~3, no.~3, pp. 1687--1694, 2018.

\bibitem{freda20193d}
L.~Freda, M.~Gianni, F.~Pirri, A.~Gawel, R.~Dub{\'e}, R.~Siegwart, and
  C.~Cadena, ``3d multi-robot patrolling with a two-level coordination
  strategy,'' \emph{Autonomous Robots}, vol.~43, no.~7, pp. 1747--1779, 2019.

\bibitem{fritsche2016radar}
P.~Fritsche, S.~Kueppers, G.~Briese, and B.~Wagner, ``Radar and lidar
  sensorfusion in low visibility environments.'' in \emph{ICINCO (2)}, 2016,
  pp. 30--36.

\bibitem{wei2016multi}
G.~Wei, J.~W. Gardner, M.~Cole, and Y.~Xing, ``Multi-sensor module for a mobile
  robot operating in harsh environments,'' in \emph{2016 IEEE SENSORS}.\hskip
  1em plus 0.5em minus 0.4em\relax IEEE, 2016, pp. 1--3.

\bibitem{gancet2005task}
J.~Gancet, G.~Hattenberger, R.~Alami, and S.~Lacroix, ``Task planning and
  control for a multi-uav system: architecture and algorithms,'' in \emph{2005
  IEEE/RSJ International Conference on Intelligent Robots and Systems}.\hskip
  1em plus 0.5em minus 0.4em\relax IEEE, 2005, pp. 1017--1022.

\bibitem{surmann2019integration}
H.~Surmann, R.~Worst, T.~Buschmann, A.~Leinweber, A.~Schmitz, G.~Senkowski, and
  N.~Goddemeier, ``Integration of uavs in urban search and rescue missions,''
  in \emph{2019 IEEE International Symposium on Safety, Security, and Rescue
  Robotics (SSRR)}.\hskip 1em plus 0.5em minus 0.4em\relax IEEE, 2019, pp.
  203--209.

\bibitem{guldenring2019heterogeneous}
J.~G{\"u}ldenring, L.~Koring, P.~Gorczak, and C.~Wietfeld, ``Heterogeneous
  multilink aggregation for reliable uav communication in maritime search and
  rescue missions,'' in \emph{2019 International Conference on Wireless and
  Mobile Computing, Networking and Communications (WiMob)}.\hskip 1em plus
  0.5em minus 0.4em\relax IEEE, 2019, pp. 215--220.

\bibitem{cubber2017introduction}
G.~Cubber, D.~Doroftei, K.~Rudin, K.~Berns, A.~Matos, D.~Serrano, J.~Sanchez,
  S.~Govindaraj, J.~Bedkowski, R.~Roda \emph{et~al.}, ``Introduction to the use
  of robotic tools for search and rescue,'' 2017.

\bibitem{winfield2016eurathlon}
A.~F. Winfield, M.~P. Franco, B.~Brueggemann, A.~Castro, M.~C. Limon, G.~Ferri,
  F.~Ferreira, X.~Liu, Y.~Petillot, J.~Roning \emph{et~al.}, ``eurathlon 2015:
  A multi-domain multi-robot grand challenge for search and rescue robots,'' in
  \emph{Annual Conference Towards Autonomous Robotic Systems}.\hskip 1em plus
  0.5em minus 0.4em\relax Springer, 2016, pp. 351--363.

\bibitem{berns2017unmanned}
K.~Berns, A.~Nezhadfard, M.~Tosa, H.~Balta, and G.~De~Cubber, ``Unmanned ground
  robots for rescue tasks,'' in \emph{Search and Rescue Robotics-From Theory to
  Practice}.\hskip 1em plus 0.5em minus 0.4em\relax IntechOpen, 2017.

\bibitem{konyo2019impact}
M.~Konyo, Y.~Ambe, H.~Nagano, Y.~Yamauchi, S.~Tadokoro, Y.~Bando, K.~Itoyama,
  H.~G. Okuno, T.~Okatani, K.~Shimizu \emph{et~al.}, ``Impact-trc thin
  serpentine robot platform for urban search and rescue,'' in \emph{Disaster
  Robotics}.\hskip 1em plus 0.5em minus 0.4em\relax Springer, 2019, pp. 25--76.

\bibitem{goodrich2008supporting}
M.~A. Goodrich, B.~S. Morse, D.~Gerhardt, J.~L. Cooper, M.~Quigley, J.~A.
  Adams, and C.~Humphrey, ``Supporting wilderness search and rescue using a
  camera-equipped mini uav,'' \emph{Journal of Field Robotics}, vol.~25, no.
  1-2, pp. 89--110, 2008.

\bibitem{qi2016search}
J.~Qi, D.~Song, H.~Shang, N.~Wang, C.~Hua, C.~Wu, X.~Qi, and J.~Han, ``Search
  and rescue rotary-wing uav and its application to the lushan ms 7.0
  earthquake,'' \emph{Journal of Field Robotics}, vol.~33, no.~3, pp. 290--321,
  2016.

\bibitem{sun2016camera}
J.~Sun, B.~Li, Y.~Jiang, and C.-y. Wen, ``A camera-based target detection and
  positioning uav system for search and rescue (sar) purposes,''
  \emph{Sensors}, vol.~16, no.~11, p. 1778, 2016.

\bibitem{konrad2017unmanned}
R.~Konrad, D.~Serrano, and P.~Strupler, ``Unmanned aerial systems,''
  \emph{Search and Rescue Robotics—From Theory to Practice}, pp. 37--52,
  2017.

\bibitem{jorge2019survey}
V.~A. Jorge, R.~Granada, R.~G. Maidana, D.~A. Jurak, G.~Heck, A.~P. Negreiros,
  D.~H. Dos~Santos, L.~M. Gon{\c{c}}alves, and A.~M. Amory, ``A survey on
  unmanned surface vehicles for disaster robotics: Main challenges and
  directions,'' \emph{Sensors}, vol.~19, no.~3, p. 702, 2019.

\bibitem{matos2017unmanned}
A.~Matos, E.~Silva, J.~Almeida, A.~Martins, H.~Ferreira, B.~Ferreira, J.~Alves,
  A.~Dias, S.~Fioravanti, D.~Bertin \emph{et~al.}, ``Unmanned maritime systems
  for search and rescue,'' \emph{Search and Rescue Robotics; IntechOpen:
  London, UK}, pp. 77--92, 2017.

\bibitem{serrano2015icarus}
D.~Serrano, G.~De~Cubber, G.~Leventakis, P.~Chrobocinski, D.~Moore, and
  S.~Govindaraj, ``Icarus and darius approaches towards interoperability,'' in
  \emph{IARP RISE Workshop, At Lisbon, Portugal. Proceedings of the NATO STO
  Lecture Series SCI-271}.\hskip 1em plus 0.5em minus 0.4em\relax Citeseer,
  2015.

\bibitem{chrobocinski2012darius}
P.~Chrobocinski, E.~Makri, N.~Zotos, C.~Stergiopoulos, and G.~Bogdos, ``Darius
  project: Deployable sar integrated chain with unmanned systems,'' in
  \emph{2012 International Conference on Telecommunications and Multimedia
  (TEMU)}.\hskip 1em plus 0.5em minus 0.4em\relax IEEE, 2012, pp. 220--226.

\bibitem{lopez2017interoperability}
D.~S. L{\'o}pez, G.~Moreno, J.~Cordero, J.~Sanchez, S.~Govindaraj, M.~M.
  Marques, V.~Lobo, S.~Fioravanti, A.~Grati, K.~Rudin \emph{et~al.},
  ``Interoperability in a heterogeneous team of search and rescue robots,'' in
  \emph{Search and Rescue Robotics-From Theory to Practice}.\hskip 1em plus
  0.5em minus 0.4em\relax Rijeka: InTech, 2017.

\bibitem{quigley2009ros}
M.~Quigley, K.~Conley, B.~Gerkey, J.~Faust, T.~Foote, J.~Leibs, R.~Wheeler, and
  A.~Y. Ng, ``Ros: an open-source robot operating system,'' in \emph{ICRA
  workshop on open source software}, vol.~3, no. 3.2.\hskip 1em plus 0.5em
  minus 0.4em\relax Kobe, Japan, 2009, p.~5.

\bibitem{davids2002urban}
A.~Davids, ``Urban search and rescue robots: from tragedy to technology,''
  \emph{IEEE Intelligent systems}, vol.~17, no.~2, pp. 81--83, 2002.

\bibitem{shah2004survey}
B.~Shah and H.~Choset, ``Survey on urban search and rescue robots,''
  \emph{Journal of the Robotics Society of Japan}, vol.~22, no.~5, pp.
  582--586, 2004.

\bibitem{liu2013robotic}
Y.~Liu and G.~Nejat, ``Robotic urban search and rescue: A survey from the
  control perspective,'' \emph{Journal of Intelligent \& Robotic Systems},
  vol.~72, no.~2, pp. 147--165, 2013.

\bibitem{ganz2015urban}
A.~Ganz, J.~M. Schafer, J.~Tang, Z.~Yang, J.~Yi, and G.~Ciottone, ``Urban
  search and rescue situational awareness using diorama disaster management
  system,'' \emph{Procedia Engineering}, vol. 107, pp. 349--356, 2015.

\bibitem{chen2017robust}
X.~Chen, H.~Zhang, H.~Lu, J.~Xiao, Q.~Qiu, and Y.~Li, ``Robust slam system
  based on monocular vision and lidar for robotic urban search and rescue,'' in
  \emph{2017 IEEE International Symposium on Safety, Security and Rescue
  Robotics (SSRR)}.\hskip 1em plus 0.5em minus 0.4em\relax IEEE, 2017, pp.
  41--47.

\bibitem{lewis2019developing}
M.~Lewis, K.~Sycara, and I.~Nourbakhsh, ``Developing a testbed for studying
  human-robot interaction in urban search and rescue,'' in \emph{Proceedings of
  the 10th International Conference on Human Computer Interaction (HCII'03)},
  2019, pp. 270--274.

\bibitem{murphy2012marine}
R.~R. Murphy, K.~L. Dreger, S.~Newsome, J.~Rodocker, B.~Slaughter, R.~Smith,
  E.~Steimle, T.~Kimura, K.~Makabe, K.~Kon \emph{et~al.}, ``Marine
  heterogeneous multirobot systems at the great eastern japan tsunami
  recovery,'' \emph{Journal of Field Robotics}, vol.~29, no.~5, pp. 819--831,
  2012.

\bibitem{mendoncca2016cooperative}
R.~Mendon{\c{c}}a, M.~M. Marques, F.~Marques, A.~Lourenco, E.~Pinto,
  P.~Santana, F.~Coito, V.~Lobo, and J.~Barata, ``A cooperative multi-robot
  team for the surveillance of shipwreck survivors at sea,'' in \emph{OCEANS
  2016 MTS/IEEE Monterey}.\hskip 1em plus 0.5em minus 0.4em\relax IEEE, 2016,
  pp. 1--6.

\bibitem{zhao2019search}
Q.~{Zhao}, J.~{Ding}, B.~{Xia}, Y.~{Guo}, B.~{Ge}, and K.~{Yang}, ``Search and
  rescue at sea: Situational factors analysis and similarity measure,'' in
  \emph{2019 IEEE International Conference on Systems, Man and Cybernetics
  (SMC)}, 2019, pp. 2678--2683.

\bibitem{silvagni2017multipurpose}
M.~Silvagni, A.~Tonoli, E.~Zenerino, and M.~Chiaberge, ``Multipurpose uav for
  search and rescue operations in mountain avalanche events,'' \emph{Geomatics,
  Natural Hazards and Risk}, vol.~8, no.~1, pp. 18--33, 2017.

\bibitem{bryant2019autonomous}
G.~Bryant, ``An autonomous multi-uav system for avalanche search,'' Master's
  thesis, NTNU, 2019.

\bibitem{chikwanha2012survey}
A.~Chikwanha, S.~Motepe, and R.~Stopforth, ``Survey and requirements for search
  and rescue ground and air vehicles for mining applications,'' in \emph{2012
  19th International Conference on Mechatronics and Machine Vision in Practice
  (M2VIP)}.\hskip 1em plus 0.5em minus 0.4em\relax IEEE, 2012, pp. 105--109.

\bibitem{zhao2017search}
J.~Zhao, J.~Gao, F.~Zhao, and Y.~Liu, ``A search-and-rescue robot system for
  remotely sensing the underground coal mine environment,'' \emph{Sensors},
  vol.~17, no.~10, p. 2426, 2017.

\bibitem{xiangRipRescue2016}
G.~{Xiang}, A.~{Hardy}, M.~{Rajeh}, and L.~{Venuthurupalli}, ``Design of the
  life-ring drone delivery system for rip current rescue,'' in \emph{2016 IEEE
  Systems and Information Engineering Design Symposium (SIEDS)}, 2016, pp.
  181--186.

\bibitem{yeong2015review}
S.~Yeong, L.~King, and S.~Dol, ``A review on marine search and rescue
  operations using unmanned aerial vehicles,'' \emph{Int. J. Mech. Aerosp. Ind.
  Mech. Manuf. Eng}, vol.~9, no.~2, pp. 396--399, 2015.

\bibitem{mittal2019vision}
M.~Mittal, R.~Mohan, W.~Burgard, and A.~Valada, ``Vision-based autonomous uav
  navigation and landing for urban search and rescue,'' \emph{arXiv preprint
  arXiv:1906.01304}, 2019.

\bibitem{tomic2012toward}
T.~Tomic, K.~Schmid, P.~Lutz, A.~Domel, M.~Kassecker, E.~Mair, I.~L. Grixa,
  F.~Ruess, M.~Suppa, and D.~Burschka, ``Toward a fully autonomous uav:
  Research platform for indoor and outdoor urban search and rescue,''
  \emph{IEEE robotics \& automation magazine}, vol.~19, no.~3, pp. 46--56,
  2012.

\bibitem{hatazaki2007active}
K.~Hatazaki, M.~Konyo, K.~Isaki, S.~Tadokoro, and F.~Takemura, ``Active scope
  camera for urban search and rescue,'' in \emph{2007 IEEE/RSJ International
  Conference on Intelligent Robots and Systems}.\hskip 1em plus 0.5em minus
  0.4em\relax IEEE, 2007, pp. 2596--2602.

\bibitem{namari2012tube}
H.~Namari, K.~Wakana, M.~Ishikura, M.~Konyo, and S.~Tadokoro, ``Tube-type
  active scope camera with high mobility and practical functionality,'' in
  \emph{2012 IEEE/RSJ International Conference on Intelligent Robots and
  Systems}.\hskip 1em plus 0.5em minus 0.4em\relax IEEE, 2012, pp. 3679--3686.

\bibitem{myeongFireproofDrone2017}
W.~C. {Myeong}, K.~Y. {Jung}, and H.~{Myung}, ``Development of faros
  (fire-proof drone) using an aramid fiber armor and air buffer layer,'' in
  \emph{2017 14th International Conference on Ubiquitous Robots and Ambient
  Intelligence (URAI)}, 2017, pp. 204--207.

\bibitem{ranjan2019wireless}
A.~Ranjan, H.~Sahu, and P.~Misra, ``Wireless robotics networks for search and
  rescue in underground mines: Taxonomy and open issues,'' in \emph{Exploring
  critical approaches of evolutionary computation}.\hskip 1em plus 0.5em minus
  0.4em\relax IGI Global, 2019, pp. 286--309.

\bibitem{macwanMultirobot2015}
A.~{Macwan}, J.~{Vilela}, G.~{Nejat}, and B.~{Benhabib}, ``A multirobot
  path-planning strategy for autonomous wilderness search and rescue,''
  \emph{IEEE Transactions on Cybernetics}, vol.~45, no.~9, pp. 1784--1797,
  2015.

\bibitem{wangVortexSearch2018}
C.~{Wang}, P.~{Liu}, T.~{Zhang}, and J.~{Sun}, ``The adaptive vortex search
  algorithm of optimal path planning for forest fire rescue uav,'' in
  \emph{2018 IEEE 3rd Advanced Information Technology, Electronic and
  Automation Control Conference (IAEAC)}, 2018, pp. 400--403.

\bibitem{kashino2019aerial}
Z.~Kashino, G.~Nejat, and B.~Benhabib, ``Aerial wilderness search and rescue
  with ground support,'' \emph{Journal of Intelligent \& Robotic Systems}, pp.
  1--17, 2019.

\bibitem{kinugasa2016validation}
T.~Kinugasa, K.~Matsuoka, N.~Miyamoto, R.~Iwado, K.~Yoshida, M.~Okugawa,
  T.~Nara, and M.~Kurisu, ``Validation of avalanche beacons implemented in a
  robot for urban search and rescue,'' in \emph{2016 11th France-Japan \& 9th
  Europe-Asia Congress on Mechatronics (MECATRONICS)/17th International
  Conference on Research and Education in Mechatronics (REM)}.\hskip 1em plus
  0.5em minus 0.4em\relax IEEE, 2016, pp. 111--116.

\bibitem{zhilenkovForest2018}
A.~A. {Zhilenkov} and I.~R. {Epifantsev}, ``System of autonomous navigation of
  the drone in difficult conditions of the forest trails,'' in \emph{2018 IEEE
  Conference of Russian Young Researchers in Electrical and Electronic
  Engineering (EIConRus)}, 2018, pp. 1036--1039.

\bibitem{fan2019towards}
H.~Fan, V.~Hernandez~Bennetts, E.~Schaffernicht, and A.~J. Lilienthal,
  ``Towards gas discrimination and mapping in emergency response scenarios
  using a mobile robot with an electronic nose,'' \emph{Sensors}, vol.~19,
  no.~3, p. 685, 2019.

\bibitem{murphy2013interacting}
R.~R. Murphy, V.~Srinivasan, Z.~Henkel, J.~Suarez, M.~Minson, J.~Straus,
  S.~Hempstead, T.~Valdez, and S.~Egawa, ``Interacting with trapped victims
  using robots,'' in \emph{2013 IEEE International Conference on Technologies
  for Homeland Security (HST)}.\hskip 1em plus 0.5em minus 0.4em\relax IEEE,
  2013, pp. 32--37.

\bibitem{chang2007towards}
C.~Chang and R.~R. Murphy, ``Towards robot-assisted mass-casualty triage,'' in
  \emph{2007 IEEE International Conference on Networking, Sensing and
  Control}.\hskip 1em plus 0.5em minus 0.4em\relax IEEE, 2007, pp. 267--272.

\bibitem{nordstrom2016vessel}
J.~Nordstr{\"o}m, F.~Goerlandt, J.~Sarsama, P.~Lepp{\"a}nen, M.~Nissil{\"a},
  P.~Ruponen, T.~L{\"u}bcke, and S.~Sonninen, ``Vessel triage: A method for
  assessing and communicating the safety status of vessels in maritime distress
  situations,'' \emph{Safety science}, vol.~85, pp. 117--129, 2016.

\bibitem{sa2014inspection}
I.~Sa, S.~Hrabar, and P.~Corke, ``Inspection of pole-like structures using a
  vision-controlled vtol uav and shared autonomy,'' in \emph{2014 IEEE/RSJ
  International Conference on Intelligent Robots and Systems}.\hskip 1em plus
  0.5em minus 0.4em\relax IEEE, 2014, pp. 4819--4826.

\bibitem{marion2018director}
P.~Marion, M.~Fallon, R.~Deits, A.~Valenzuela, C.~P. D’Arpino, G.~Izatt,
  L.~Manuelli, M.~Antone, H.~Dai, T.~Koolen \emph{et~al.}, ``Director: A user
  interface designed for robot operation with shared autonomy,'' in \emph{The
  DARPA Robotics Challenge Finals: Humanoid Robots To The Rescue}.\hskip 1em
  plus 0.5em minus 0.4em\relax Springer, 2018, pp. 237--270.

\bibitem{masone2014semi}
C.~Masone, P.~R. Giordano, H.~H. B{\"u}lthoff, and A.~Franchi,
  ``Semi-autonomous trajectory generation for mobile robots with integral
  haptic shared control,'' in \emph{2014 IEEE International Conference on
  Robotics and Automation (ICRA)}.\hskip 1em plus 0.5em minus 0.4em\relax IEEE,
  2014, pp. 6468--6475.

\bibitem{franchi2012shared}
A.~Franchi, C.~Secchi, M.~Ryll, H.~H. Bulthoff, and P.~R. Giordano, ``Shared
  control: Balancing autonomy and human assistance with a group of quadrotor
  uavs,'' \emph{IEEE Robotics \& Automation Magazine}, vol.~19, no.~3, pp.
  57--68, 2012.

\bibitem{lee2013semiautonomous}
D.~Lee, A.~Franchi, H.~I. Son, C.~Ha, H.~H. B{\"u}lthoff, and P.~R. Giordano,
  ``Semiautonomous haptic teleoperation control architecture of multiple
  unmanned aerial vehicles,'' \emph{IEEE/ASME Transactions on Mechatronics},
  vol.~18, no.~4, pp. 1334--1345, 2013.

\bibitem{kolling2013human}
A.~Kolling, K.~Sycara, S.~Nunnally, and M.~Lewis, ``Human-swarm interaction: An
  experimental study of two types of interaction with foraging swarms,''
  \emph{Journal of Human-Robot Interaction}, vol.~2, no.~2, pp. 103--129, 2013.

\bibitem{naghsh2008analysis}
A.~M. Naghsh, J.~Gancet, A.~Tanoto, and C.~Roast, ``Analysis and design of
  human-robot swarm interaction in firefighting,'' in \emph{RO-MAN 2008-The
  17th IEEE International Symposium on Robot and Human Interactive
  Communication}.\hskip 1em plus 0.5em minus 0.4em\relax IEEE, 2008, pp.
  255--260.

\bibitem{marcotte2019adaptive}
{R. Marcotte}, ``Adaptive communication for mobile multi-robot systems,'' Ph.D.
  dissertation, 2019.

\bibitem{jawhar2018networking}
I.~Jawhar, N.~Mohamed, J.~Wu, and J.~Al-Jaroodi, ``Networking of multi-robot
  systems: architectures and requirements,'' \emph{Journal of Sensor and
  Actuator Networks}, vol.~7, no.~4, p.~52, 2018.

\bibitem{de2010autonomous}
J.~De~Hoog, S.~Cameron, and A.~Visser, ``Autonomous multi-robot exploration in
  communication-limited environments,'' in \emph{Proceedings of the Conference
  on Towards Autonomous Robotic Systems}.\hskip 1em plus 0.5em minus
  0.4em\relax Citeseer, 2010, pp. 68--75.

\bibitem{anjum2017review}
S.~S. Anjum, R.~M. Noor, and M.~H. Anisi, ``Review on manet based communication
  for search and rescue operations,'' \emph{Wireless personal communications},
  vol.~94, no.~1, pp. 31--52, 2017.

\bibitem{atoev2017data}
S.~Atoev, K.-R. Kwon, S.-H. Lee, and K.-S. Moon, ``Data analysis of the mavlink
  communication protocol,'' in \emph{2017 International Conference on
  Information Science and Communications Technologies (ICISCT)}.\hskip 1em plus
  0.5em minus 0.4em\relax IEEE, 2017, pp. 1--3.

\bibitem{curtis2017uav}
N.~Curtis-Brown, I.~Guzman, T.~Sherman, J.~Tellez, E.~Gomez, and S.~Bhandari,
  ``Uav collision detection and avoidance using ads-b sensor and custom ads-b
  like solution,'' in \emph{Proceedings of Infotech@ Aerospace Conference},
  2017, pp. 9--13.

\bibitem{sherman2017autonomous}
T.~Sherman, M.~Caudle, H.~Haideri, J.~Lopez, and S.~Bhandari, ``Autonomous
  collision avoidance of uavs using ads-b transponders,'' 2017.

\bibitem{stroupe2001distributed}
A.~W. Stroupe, M.~C. Martin, and T.~Balch, ``Distributed sensor fusion for
  object position estimation by multi-robot systems,'' in \emph{Proceedings
  2001 ICRA. IEEE international conference on robotics and automation (Cat. No.
  01CH37164)}, vol.~2.\hskip 1em plus 0.5em minus 0.4em\relax IEEE, 2001, pp.
  1092--1098.

\bibitem{queralta2019collaborative}
J.~{Pe\~{n}a Queralta}, T.~N. Gia, H.~Tenhunen, and T.~Westerlund, ``Collaborative
  mapping with {IoE}-based heterogeneous vehicles for enhanced situational
  awareness,'' in \emph{IEEE Sensors Applications Symposium (SAS)}.\hskip 1em
  plus 0.5em minus 0.4em\relax IEEE, 2019.

\bibitem{stoy2001using}
K.~St{\o}y, ``Using situated communication in distributed autonomous mobile
  robotics.'' in \emph{SCAI}, vol.~1.\hskip 1em plus 0.5em minus 0.4em\relax
  Citeseer, 2001, pp. 44--52.

\bibitem{biswas2010wifi}
J.~Biswas and M.~Veloso, ``Wifi localization and navigation for autonomous
  indoor mobile robots,'' in \emph{2010 IEEE international conference on
  robotics and automation}.\hskip 1em plus 0.5em minus 0.4em\relax IEEE, 2010,
  pp. 4379--4384.

\bibitem{kotaru2015spotfi}
M.~Kotaru, K.~Joshi, D.~Bharadia, and S.~Katti, ``Spotfi: Decimeter level
  localization using wifi,'' in \emph{Proceedings of the 2015 ACM Conference on
  Special Interest Group on Data Communication}, 2015, pp. 269--282.

\bibitem{sun2018augmentation}
W.~Sun, M.~Xue, H.~Yu, H.~Tang, and A.~Lin, ``Augmentation of fingerprints for
  indoor wifi localization based on gaussian process regression,'' \emph{IEEE
  Transactions on Vehicular Technology}, vol.~67, no.~11, pp. 10\,896--10\,905,
  2018.

\bibitem{altini2010bluetooth}
M.~Altini, D.~Brunelli, E.~Farella, and L.~Benini, ``Bluetooth indoor
  localization with multiple neural networks,'' in \emph{IEEE 5th International
  Symposium on Wireless Pervasive Computing 2010}.\hskip 1em plus 0.5em minus
  0.4em\relax IEEE, 2010, pp. 295--300.

\bibitem{kriz2016improving}
P.~Kriz, F.~Maly, and T.~Kozel, ``Improving indoor localization using bluetooth
  low energy beacons,'' \emph{Mobile Information Systems}, vol. 2016, 2016.

\bibitem{wisanmongkol2019multipath}
J.~Wisanmongkol, L.~Klinkusoom, T.~Sanpechuda, L.-o. Kovavisaruch, and
  K.~Kaemarungsi, ``Multipath mitigation for rssi-based bluetooth low energy
  localization,'' in \emph{2019 19th International Symposium on Communications
  and Information Technologies (ISCIT)}.\hskip 1em plus 0.5em minus 0.4em\relax
  IEEE, 2019, pp. 47--51.

\bibitem{kanaris2017fusing}
L.~Kanaris, A.~Kokkinis, A.~Liotta, and S.~Stavrou, ``Fusing bluetooth beacon
  data with wi-fi radiomaps for improved indoor localization,'' \emph{Sensors},
  vol.~17, no.~4, p. 812, 2017.

\bibitem{suryavanshi2019direction}
N.~B. Suryavanshi, K.~V. Reddy, and V.~R. Chandrika, ``Direction finding
  capability in bluetooth 5.1 standard,'' in \emph{International Conference on
  Ubiquitous Communications and Network Computing}.\hskip 1em plus 0.5em minus
  0.4em\relax Springer, 2019, pp. 53--65.

\bibitem{khan2018angle}
A.~Khan, S.~Wang, and Z.~Zhu, ``Angle-of-arrival estimation using an adaptive
  machine learning framework,'' \emph{IEEE Communications Letters}, vol.~23,
  no.~2, pp. 294--297, 2018.

\bibitem{ito2014w}
S.~Ito, F.~Endres, M.~Kuderer, G.~D. Tipaldi, C.~Stachniss, and W.~Burgard,
  ``W-rgb-d: floor-plan-based indoor global localization using a depth camera
  and wifi,'' in \emph{2014 IEEE international conference on robotics and
  automation (ICRA)}.\hskip 1em plus 0.5em minus 0.4em\relax IEEE, 2014, pp.
  417--422.

\bibitem{raghavan2010accurate}
A.~N. Raghavan, H.~Ananthapadmanaban, M.~S. Sivamurugan, and B.~Ravindran,
  ``Accurate mobile robot localization in indoor environments using
  bluetooth,'' in \emph{2010 IEEE international conference on robotics and
  automation}.\hskip 1em plus 0.5em minus 0.4em\relax IEEE, 2010, pp.
  4391--4396.

\bibitem{shule2020uwb}
W.~Shule, C.~M. Almansa, J.~{Pe\~{n}a Queralta}, Z.~Zou, and T.~Westerlund,
  ``Uwb-based localization for multi-uav systems and collaborative
  heterogeneous multi-robot systems: a survey,'' \emph{arXiv preprint
  arXiv:2004.08174}, 2020.

\bibitem{queralta2020uwbbased}
J.~{Pe\~{n}a Queralta}, C.~M. Almansa, F.~Schiano, D.~Floreano, and T.~Westerlund,
  ``Uwb-based system for uav localization in gnss-denied environments:
  Characterization and dataset,'' 2020.

\bibitem{almansa2020autocalibration}
C.~M. Almansa, W.~Shule, J.~{Pe\~{n}a Queralta}, and T.~Westerlund,
  ``Autocalibration of a mobile uwb localization system for ad-hoc multi-robot
  deployments in gnss-denied environments,'' \emph{arXiv preprint
  arXiv:2004.06762}, 2020.

\bibitem{tian2005connectivity}
D.~Tian and N.~D. Georganas, ``Connectivity maintenance and coverage
  preservation in wireless sensor networks,'' \emph{Ad Hoc Networks}, vol.~3,
  no.~6, pp. 744--761, 2005.

\bibitem{sabattini2013decentralized}
L.~Sabattini, N.~Chopra, and C.~Secchi, ``Decentralized connectivity
  maintenance for cooperative control of mobile robotic systems,'' \emph{The
  International Journal of Robotics Research}, vol.~32, no.~12, pp. 1411--1423,
  2013.

\bibitem{sabattini2013distributed}
L.~Sabattini, C.~Secchi, N.~Chopra, and A.~Gasparri, ``Distributed control of
  multirobot systems with global connectivity maintenance,'' \emph{IEEE
  Transactions on Robotics}, vol.~29, no.~5, pp. 1326--1332, 2013.

\bibitem{wang2016multi}
L.~Wang, A.~D. Ames, and M.~Egerstedt, ``Multi-objective compositions for
  collision-free connectivity maintenance in teams of mobile robots,'' in
  \emph{2016 IEEE 55th Conference on Decision and Control (CDC)}.\hskip 1em
  plus 0.5em minus 0.4em\relax IEEE, 2016, pp. 2659--2664.

\bibitem{gasparri2017bounded}
A.~Gasparri, L.~Sabattini, and G.~Ulivi, ``Bounded control law for global
  connectivity maintenance in cooperative multirobot systems,'' \emph{IEEE
  Transactions on Robotics}, vol.~33, no.~3, pp. 700--717, 2017.

\bibitem{xiao2019cooperative}
H.~Xiao, R.~Cui, and D.~Xu, ``Cooperative multi-agent search using bayesian
  approach with connectivity maintenance,'' \emph{Assembly Automation}, 2019.

\bibitem{zhu2017connectivity}
Q.~Zhu, R.~Zhou, and J.~Zhang, ``Connectivity maintenance based on multiple
  relay uavs selection scheme in cooperative surveillance,'' \emph{Applied
  Sciences}, vol.~7, no.~1, p.~8, 2017.

\bibitem{panerati2019robust}
J.~Panerati, M.~Minelli, C.~Ghedini, L.~Meyer, M.~Kaufmann, L.~Sabattini, and
  G.~Beltrame, ``Robust connectivity maintenance for fallible robots,''
  \emph{Autonomous Robots}, vol.~43, no.~3, pp. 769--787, 2019.

\bibitem{ghedini2018decentralized}
C.~Ghedini, C.~H. Ribeiro, and L.~Sabattini, ``A decentralized control strategy
  for resilient connectivity maintenance in multi-robot systems subject to
  failures,'' in \emph{Distributed Autonomous Robotic Systems}.\hskip 1em plus
  0.5em minus 0.4em\relax Springer, 2018, pp. 89--102.

\bibitem{siligardi2019robust}
L.~Siligardi, J.~Panerati, M.~Kaufmann, M.~Minelli, C.~Ghedini, G.~Beltrame,
  and L.~Sabattini, ``Robust area coverage with connectivity maintenance,'' in
  \emph{2019 International Conference on Robotics and Automation (ICRA)}.\hskip
  1em plus 0.5em minus 0.4em\relax IEEE, 2019, pp. 2202--2208.

\bibitem{khateri2019comparison}
K.~Khateri, M.~Pourgholi, M.~Montazeri, and L.~Sabattini, ``A comparison
  between decentralized local and global methods for connectivity maintenance
  of multi-robot networks,'' \emph{IEEE Robotics and Automation Letters},
  vol.~4, no.~2, pp. 633--640, 2019.

\bibitem{tian2019reliable}
Y.~Tian, ``Reliable and resource-aware collaborative slam for multi-robot
  search and rescue,'' Ph.D. dissertation, Massachusetts Institute of
  Technology, 2019.

\bibitem{maza2007multiple}
I.~Maza and A.~Ollero, ``Multiple uav cooperative searching operation using
  polygon area decomposition and efficient coverage algorithms,'' in
  \emph{Distributed Autonomous Robotic Systems 6}.\hskip 1em plus 0.5em minus
  0.4em\relax Springer, 2007, pp. 221--230.

\bibitem{suarez2011survey}
J.~Suarez and R.~Murphy, ``A survey of animal foraging for directed, persistent
  search by rescue robotics,'' in \emph{2011 IEEE International Symposium on
  Safety, Security, and Rescue Robotics}.\hskip 1em plus 0.5em minus
  0.4em\relax IEEE, 2011, pp. 314--320.

\bibitem{faigl2014comparison}
J.~Faigl, O.~Simonin, and F.~Charpillet, ``Comparison of task-allocation
  algorithms in frontier-based multi-robot exploration,'' in \emph{European
  Conference on Multi-Agent Systems}.\hskip 1em plus 0.5em minus 0.4em\relax
  Springer, 2014, pp. 101--110.

\bibitem{hussein2014multi}
A.~Hussein, M.~Adel, M.~Bakr, O.~M. Shehata, and A.~Khamis, ``Multi-robot task
  allocation for search and rescue missions,'' in \emph{Journal of Physics:
  Conference Series}, vol. 570, no.~5, 2014, p. 052006.

\bibitem{zhao2015heuristic}
W.~Zhao, Q.~Meng, and P.~W. Chung, ``A heuristic distributed task allocation
  method for multivehicle multitask problems and its application to search and
  rescue scenario,'' \emph{IEEE transactions on cybernetics}, vol.~46, no.~4,
  pp. 902--915, 2015.

\bibitem{tang2018using}
J.~Tang, K.~Zhu, H.~Guo, C.~Gong, C.~Liao, and S.~Zhang, ``Using auction-based
  task allocation scheme for simulation optimization of search and rescue in
  disaster relief,'' \emph{Simulation Modelling Practice and Theory}, vol.~82,
  pp. 132--146, 2018.

\bibitem{tadewos2019fly}
T.~G. Tadewos, L.~Shamgah, and A.~Karimoddini, ``On-the-fly decentralized
  tasking of autonomous vehicles,'' in \emph{2019 IEEE 58th Conference on
  Decision and Control (CDC)}.\hskip 1em plus 0.5em minus 0.4em\relax IEEE,
  2019, pp. 2770--2775.

\bibitem{dias2006market}
M.~B. Dias, R.~Zlot, N.~Kalra, and A.~Stentz, ``Market-based multirobot
  coordination: A survey and analysis,'' \emph{Proceedings of the IEEE},
  vol.~94, no.~7, pp. 1257--1270, 2006.

\bibitem{mosteo2010survey}
A.~R. Mosteo and L.~Montano, ``A survey of multi-robot task allocation,''
  \emph{Instituto de Investigacin en Ingenier{\l}a de Aragn (I3A), Tech. Rep},
  2010.

\bibitem{kurdi2016bio}
H.~Kurdi, J.~How, and G.~Bautista, ``Bio-inspired algorithm for task allocation
  in multi-uav search and rescue missions,'' in \emph{Aiaa guidance,
  navigation, and control conference}, 2016, p. 1377.

\bibitem{sage}
``{Dec-MCTS: Decentralized planning for multi-robot active perception},''
  \emph{The International Journal of Robotics Research}, vol.~38, no. 2-3, pp.
  316--337, 2019.

\bibitem{Best2018}
G.~Best, J.~Faigl, and R.~Fitch, ``{Online planning for multi-robot active
  perception with self-organising maps},'' \emph{Autonomous Robots}, vol.~42,
  no.~4, pp. 715--738, 2018.

\bibitem{liu2015supervisory}
Y.~Liu, M.~Ficocelli, and G.~Nejat, ``A supervisory control method for
  multi-robot task allocation in urban search and rescue,'' in \emph{2015 IEEE
  International Symposium on Safety, Security, and Rescue Robotics
  (SSRR)}.\hskip 1em plus 0.5em minus 0.4em\relax IEEE, 2015, pp. 1--6.

\bibitem{oh2015survey}
K.-K. Oh, M.-C. Park, and H.-S. Ahn, ``A survey of multi-agent formation
  control,'' \emph{Automatica}, vol.~53, pp. 424--440, 2015.

\bibitem{mccord2019progressive}
{C. McCord, J. Pe\~{n}a Queralta, T. N. Gia, T. Westerlund}, ``{Distributed
  Progressive Formation Control for Multi-Agent Systems: {2D} and {3D}
  deployment of {UAVs} in {ROS}/Gazebo with RotorS},'' in \emph{2019 European
  Conference on Mobile Robots (ECMR)}.\hskip 1em plus 0.5em minus 0.4em\relax
  IEEE, 2019.

\bibitem{shamma2008cooperative}
J.~Shamma, \emph{Cooperative control of distributed multi-agent systems}.\hskip
  1em plus 0.5em minus 0.4em\relax John Wiley \& Sons, 2008.

\bibitem{queralta2019progressive}
J.~{Pe\~{n}a Queralta}, L.~Qingqing, T.~N. Gia, H.~Tenhunen, and T.~Westerlund,
  ``Distributed progressive formation control with one-way communication for
  multi-agent systems,'' in \emph{2019 IEEE Symposium Series on Computational
  Intelligence}, 2019.

\bibitem{queralta2019indexfree}
J.~{Pe\~{n}a Queralta}, C.~McCord, T.~N. Gia, H.~Tenhunen, and T.~Westerlund,
  ``Communication-free and index-free distributed formation control algorithm
  for multi-robot systems,'' \emph{Procedia Computer Science}, 2019, the 10th
  ANT Conference.

\bibitem{abbasi2014link}
A.~Abbasi, ``Link formation pattern during emergency response network
  dynamics,'' \emph{Natural Hazards}, vol.~71, no.~3, pp. 1957--1969, 2014.

\bibitem{lin2014adaptive}
J.~Lin, Y.~Wu, G.~Wu, and J.~Xu, ``An adaptive approach for multi-agent
  formation control in manet based on cps perspective,'' \emph{Journal of
  Networks}, vol.~9, no.~5, p. 1169, 2014.

\bibitem{ray2006dynamic}
L.~Ray, J.~Joslin, J.~Murphy, J.~Barlow, D.~Brande, and D.~Balkcom, ``Dynamic
  mobile robots for emergency surveillance and situational awareness,'' in
  \emph{IEEE International Workshop on Safety, Security, and Rescue Robotics},
  2006.

\bibitem{saez2010multi}
J.~Saez-Pons, L.~Alboul, J.~Penders, and L.~Nomdedeu, ``Multi-robot team
  formation control in the guardians project,'' \emph{Industrial Robot: An
  International Journal}, 2010.

\bibitem{aftab2017self}
F.~Aftab, Z.~Zhang, and A.~Ahmad, ``Self-organization based clustering in
  manets using zone based group mobility,'' \emph{IEEE Access}, vol.~5, pp.
  27\,464--27\,476, 2017.

\bibitem{cabreira2019survey}
T.~M. Cabreira, L.~B. Brisolara, and P.~R. Ferreira~Jr, ``Survey on coverage
  path planning with unmanned aerial vehicles,'' \emph{Drones}, vol.~3, no.~1,
  p.~4, 2019.

\bibitem{hert1998polygon}
S.~Hert and V.~Lumelsky, ``Polygon area decomposition for multiple-robot
  workspace division,'' \emph{International Journal of Computational Geometry
  \& Applications}, vol.~8, no.~04, pp. 437--466, 1998.

\bibitem{araujo2013multiple}
J.~Araujo, P.~Sujit, and J.~B. Sousa, ``Multiple uav area decomposition and
  coverage,'' in \emph{2013 IEEE symposium on computational intelligence for
  security and defense applications (CISDA)}.\hskip 1em plus 0.5em minus
  0.4em\relax IEEE, 2013, pp. 30--37.

\bibitem{kantaros2014visibility}
Y.~Kantaros, M.~Thanou, and A.~Tzes, ``Visibility-oriented coverage control of
  mobile robotic networks on non-convex regions,'' in \emph{2014 IEEE
  International Conference on Robotics and Automation (ICRA)}.\hskip 1em plus
  0.5em minus 0.4em\relax IEEE, 2014, pp. 1126--1131.

\bibitem{vasquez2018coverage}
J.~I. Vasquez-Gomez, J.-C. Herrera-Lozada, and M.~Olguin-Carbajal, ``Coverage
  path planning for surveying disjoint areas,'' in \emph{2018 International
  Conference on Unmanned Aircraft Systems (ICUAS)}.\hskip 1em plus 0.5em minus
  0.4em\relax IEEE, 2018, pp. 899--904.

\bibitem{hayat2017multi}
S.~Hayat, E.~Yanmaz, T.~X. Brown, and C.~Bettstetter, ``Multi-objective uav
  path planning for search and rescue,'' in \emph{2017 IEEE International
  Conference on Robotics and Automation (ICRA)}.\hskip 1em plus 0.5em minus
  0.4em\relax IEEE, 2017, pp. 5569--5574.

\bibitem{georgiadou2017multi}
P.~S. Georgiadou, I.~A. Papazoglou, C.~T. Kiranoudis, and N.~C. Markatos,
  ``Multi-objective emergency response optimization around chemical plants,''
  in \emph{MULTI-OBJECTIVE OPTIMIZATION: Techniques and Application in Chemical
  Engineering}.\hskip 1em plus 0.5em minus 0.4em\relax World Scientific, 2017,
  pp. 355--378.

\bibitem{papatheodorou2016distributed}
S.~Papatheodorou, Y.~Stergiopoulos, and A.~Tzes, ``Distributed area coverage
  control with imprecise robot localization,'' in \emph{2016 24th Mediterranean
  Conference on Control and Automation (MED)}.\hskip 1em plus 0.5em minus
  0.4em\relax IEEE, 2016, pp. 214--219.

\bibitem{zhao2018decentralized}
W.~Zhao and B.~Bonakdarpour, ``Decentralized multi-uav routing in the presence
  of disturbances,'' \emph{arXiv preprint arXiv:1807.04823}, 2018.

\bibitem{garey1982complexity}
M.~{Garey}, D.~{Johnson}, and H.~{Witsenhausen}, ``The complexity of the
  generalized lloyd - max problem (corresp.),'' \emph{IEEE Transactions on
  Information Theory}, vol.~28, no.~2, pp. 255--256, 1982.

\bibitem{le2012adaptive}
J.~Le~Ny and G.~J. Pappas, ``Adaptive deployment of mobile robotic networks,''
  \emph{IEEE Transactions on automatic control}, vol.~58, no.~3, pp. 654--666,
  2012.

\bibitem{schwager2009decentralized}
M.~Schwager, D.~Rus, and J.-J. Slotine, ``Decentralized, adaptive coverage
  control for networked robots,'' \emph{The International Journal of Robotics
  Research}, vol.~28, no.~3, pp. 357--375, 2009.

\bibitem{aggarwal2020path}
S.~Aggarwal and N.~Kumar, ``Path planning techniques for unmanned aerial
  vehicles: A review, solutions, and challenges,'' \emph{Computer
  Communications}, vol. 149, pp. 270--299, 2020.

\bibitem{xie2020path}
J.~Xie, L.~R.~G. Carrillo, and L.~Jin, ``Path planning for uav to cover
  multiple separated convex polygonal regions,'' \emph{IEEE Access}, vol.~8,
  pp. 51\,770--51\,785, 2020.

\bibitem{lugo2014dubins}
I.~Lugo-C{\'a}rdenas, G.~Flores, S.~Salazar, and R.~Lozano, ``Dubins path
  generation for a fixed wing uav,'' in \emph{2014 International conference on
  unmanned aircraft systems (ICUAS)}.\hskip 1em plus 0.5em minus 0.4em\relax
  IEEE, 2014, pp. 339--346.

\bibitem{liao2010full}
Y.~Liao, L.~Wan, and J.~Zhuang, ``Full state-feedback stabilization of an
  underactuated unmanned surface vehicle,'' in \emph{2010 2nd International
  Conference on Advanced Computer Control}, vol.~4.\hskip 1em plus 0.5em minus
  0.4em\relax IEEE, 2010, pp. 70--74.

\bibitem{cetinsoy2013design}
E.~Cetinsoy, ``Design and flight tests of a holonomic quadrotor uav with
  sub-rotor control surfaces,'' in \emph{2013 IEEE International Conference on
  Mechatronics and Automation}.\hskip 1em plus 0.5em minus 0.4em\relax IEEE,
  2013, pp. 1197--1202.

\bibitem{damoto2001holonomic}
R.~Damoto, W.~Cheng, and S.~Hirose, ``Holonomic omnidirectional vehicle with
  new omni-wheel mechanism,'' in \emph{Proceedings 2001 ICRA. IEEE
  International Conference on Robotics and Automation (Cat. No. 01CH37164)},
  vol.~1.\hskip 1em plus 0.5em minus 0.4em\relax IEEE, 2001, pp. 773--778.

\bibitem{campbell2012review}
S.~Campbell, W.~Naeem, and G.~W. Irwin, ``A review on improving the autonomy of
  unmanned surface vehicles through intelligent collision avoidance
  manoeuvres,'' \emph{Annual Reviews in Control}, vol.~36, no.~2, pp. 267--283,
  2012.

\bibitem{zeng2015survey}
Z.~Zeng, L.~Lian, K.~Sammut, F.~He, Y.~Tang, and A.~Lammas, ``A survey on path
  planning for persistent autonomy of autonomous underwater vehicles,''
  \emph{Ocean Engineering}, vol. 110, pp. 303--313, 2015.

\bibitem{li2018path}
D.~Li, P.~Wang, and L.~Du, ``Path planning technologies for autonomous
  underwater vehicles-a review,'' \emph{IEEE Access}, vol.~7, pp. 9745--9768,
  2018.

\bibitem{santos2020path}
L.~C. Santos, F.~N. Santos, E.~S. Pires, A.~Valente, P.~Costa, and
  S.~Magalh{\~a}es, ``Path planning for ground robots in agriculture: a short
  review,'' in \emph{2020 IEEE International Conference on Autonomous Robot
  Systems and Competitions (ICARSC)}.\hskip 1em plus 0.5em minus 0.4em\relax
  IEEE, 2020, pp. 61--66.

\bibitem{ozkan2019rescue}
M.~F. Ozkan, L.~R.~G. Carrillo, and S.~A. King, ``Rescue boat path planning in
  flooded urban environments,'' in \emph{2019 IEEE International Symposium on
  Measurement and Control in Robotics (ISMCR)}.\hskip 1em plus 0.5em minus
  0.4em\relax IEEE, 2019, pp. B2--2.

\bibitem{di2015energy}
C.~Di~Franco and G.~Buttazzo, ``Energy-aware coverage path planning of uavs,''
  in \emph{2015 IEEE international conference on autonomous robot systems and
  competitions}.\hskip 1em plus 0.5em minus 0.4em\relax IEEE, 2015, pp.
  111--117.

\bibitem{cabreira2018energy}
T.~M. Cabreira, C.~Di~Franco, P.~R. Ferreira, and G.~C. Buttazzo,
  ``Energy-aware spiral coverage path planning for uav photogrammetric
  applications,'' \emph{IEEE Robotics and Automation Letters}, vol.~3, no.~4,
  pp. 3662--3668, 2018.

\bibitem{lee2015energy}
T.~Lee, H.~Kim, H.~Chung, Y.~Bang, and H.~Myung, ``Energy efficient path
  planning for a marine surface vehicle considering heading angle,''
  \emph{Ocean Engineering}, vol. 107, pp. 118--131, 2015.

\bibitem{alami1995multi}
R.~Alami, F.~Robert, F.~Ingrand, and S.~Suzuki, ``Multi-robot cooperation
  through incremental plan-merging,'' in \emph{Proceedings of 1995 IEEE
  International Conference on Robotics and Automation}, vol.~3.\hskip 1em plus
  0.5em minus 0.4em\relax IEEE, 1995, pp. 2573--2579.

\bibitem{yuan2010cooperative}
J.~Yuan, Y.~Huang, T.~Tao, and F.~Sun, ``A cooperative approach for multi-robot
  area exploration,'' in \emph{2010 IEEE/RSJ International Conference on
  Intelligent Robots and Systems}.\hskip 1em plus 0.5em minus 0.4em\relax IEEE,
  2010, pp. 1390--1395.

\bibitem{jain2012multi}
U.~Jain, R.~Tiwari, S.~Majumdar, and S.~Sharma, ``Multi robot area exploration
  using circle partitioning method,'' \emph{Procedia Engineering}, vol.~41, pp.
  383--387, 2012.

\bibitem{singh2014comparative}
R.~K. Singh and N.~Jain, ``Comparative study of multi-robot area exploration
  algorithms,'' \emph{International Journal}, vol.~4, no.~8, 2014.

\bibitem{choi2019multi}
Y.~Choi, Y.~Choi, S.~Briceno, and D.~N. Mavris, ``Multi-uas path-planning for a
  large-scale disjoint disaster management,'' in \emph{2019 International
  Conference on Unmanned Aircraft Systems (ICUAS)}.\hskip 1em plus 0.5em minus
  0.4em\relax IEEE, 2019.

\bibitem{wolf2019path}
P.~Wolf, R.~Hess, and K.~Schilling, ``Path planning for multiple uavs covering
  disjoint non-convex areas,'' in \emph{2019 IEEE International Symposium on
  Safety, Security, and Rescue Robotics (SSRR)}.\hskip 1em plus 0.5em minus
  0.4em\relax IEEE, 2019, pp. 151--157.

\bibitem{li2018multi}
J.~Li, X.~Li, and L.~Yu, ``Multi-uav cooperative coverage path planning in
  plateau and mountain environment,'' in \emph{2018 33rd Youth Academic Annual
  Conference of Chinese Association of Automation (YAC)}.\hskip 1em plus 0.5em
  minus 0.4em\relax IEEE, 2018, pp. 820--824.

\bibitem{ricciardi2019improved}
L.~A. Ricciardi and M.~Vasile, ``Improved archiving and search strategies for
  multi agent collaborative search,'' in \emph{Advances in Evolutionary and
  Deterministic Methods for Design, Optimization and Control in Engineering and
  Sciences}.\hskip 1em plus 0.5em minus 0.4em\relax Springer, 2019, pp.
  435--455.

\bibitem{vasile2011multiagentcollaborativesearch}
M.~Vasile and F.~Zuiani, ``Multi-agent collaborative search: an agent-based
  memetic multi-objective optimization algorithm applied to space trajectory
  design,'' \emph{Proceedings of the Institution of Mechanical Engineers, Part
  G: Journal of Aerospace Engineering}, vol. 225, no.~11, pp. 1211--1227, 2011.

\bibitem{narzisi2006multi}
G.~Narzisi, V.~Mysore, and B.~Mishra, ``Multi-objective evolutionary
  optimization of agent-based models: An application to emergency response
  planning.'' \emph{Computational Intelligence}, vol. 2006, pp. 224--230, 2006.

\bibitem{muecke2011distributed}
K.~Muecke and B.~Powell, ``A distributed, heterogeneous, target-optimized
  operating system for a multi-robot search and rescue application,'' in
  \emph{International Conference on Industrial, Engineering and Other
  Applications of Applied Intelligent Systems}.\hskip 1em plus 0.5em minus
  0.4em\relax Springer, 2011, pp. 266--275.

\bibitem{abbasi2016coverage}
F.~Abbasi~Doustvatan, ``Coverage control for heterogeneous multi-agent
  systems,'' Ph.D. dissertation, University of Georgia, 2016.

\bibitem{rizk2019cooperative}
Y.~Rizk, M.~Awad, and E.~W. Tunstel, ``Cooperative heterogeneous multi-robot
  systems: A survey,'' \emph{ACM Computing Surveys (CSUR)}, vol.~52, no.~2, pp.
  1--31, 2019.

\bibitem{yolov3}
J.~Redmon and A.~Farhadi, ``Yolov3: An incremental improvement,'' \emph{arXiv},
  2018.

\bibitem{Taipalmaa-MLSP}
J.~Taipalmaa, N.~Passalis, H.~Zhang, M.~Gabbouj, and J.~Raitoharju,
  ``\BIBforeignlanguage{English}{High-resolution water segmentation for
  autonomous unmanned surface vehicles: a novel dataset and evaluation},'' in
  \emph{\BIBforeignlanguage{English}{2019 IEEE 29th International Workshop on
  Machine Learning for Signal Processing (MLSP)}}, ser. IEEE International
  Workshop on Machine Learning for Signal Processing.\hskip 1em plus 0.5em
  minus 0.4em\relax IEEE, 10 2019.

\bibitem{CV-preview}
L.~Lopez-Fuentes, J.~Weijer, M.~Gonz\'{a}lez-Hidalgo, H.~Skinnemoen, and A.~D.
  Bagdanov, ``Review on computer vision techniques in emergency situations,''
  \emph{Multimedia Tools Appl.}, vol.~77, no.~13, Jul. 2018.

\bibitem{semantic-Survey}
F.~Lateef and Y.~Ruichek, ``Survey on semantic segmentation using deep learning
  techniques,'' \emph{Neurocomputing}, 02 2019.

\bibitem{Driving-Survey}
M.~{Siam}, M.~{Gamal}, M.~{Abdel-Razek}, S.~{Yogamani}, M.~{Jagersand}, and
  H.~{Zhang}, ``A comparative study of real-time semantic segmentation for
  autonomous driving,'' in \emph{2018 IEEE/CVF Conference on Computer Vision
  and Pattern Recognition Workshops}, 2018, pp. 700--70\,010.

\bibitem{Semantic-Benchmark}
B.~Bovcon and M.~Kristan, ``Benchmarking semantic segmentation methods for
  obstacle detection on a marine environment,'' in \emph{24th Computer Vision
  Winter Workshop}, 2019.

\bibitem{U-Net}
O.~Ronneberger, P.~Fischer, and T.~Brox, ``U-net: Convolutional networks for
  biomedical image segmentation,'' in \emph{Medical Image Computing and
  Computer-Assisted Intervention -- MICCAI 2015}, N.~Navab, J.~Hornegger, W.~M.
  Wells, and A.~F. Frangi, Eds.\hskip 1em plus 0.5em minus 0.4em\relax Cham:
  Springer International Publishing, 2015, pp. 234--241.

\bibitem{MarineSS-13}
H.~Zhao, J.~Shi, X.~Qi, X.~Wang, and J.~Jia, ``Pyramid scene parsing network,''
  in \emph{CVPR}, 2017.

\bibitem{MarineSS-6}
L.-C. Chen, G.~Papandreou, I.~Kokkinos, K.~Murphy, and A.~L. Yuille, ``Deeplab:
  Semantic image segmentation with deep convolutional nets, atrous convolution,
  and fully connected crfs,'' \emph{IEEE Transactions on Pattern Analysis and
  Machine Intelligence}, vol.~40, pp. 834--848, 2018.

\bibitem{Modd2}
B.~Bovcon, J.~Muhovi{\v{c}}, J.~Per{\v{s}}, and M.~Kristan, ``Stereo obstacle
  detection for unmanned surface vehicles by {IMU}-assisted semantic
  segmentation,'' \emph{Robotics and Autonomous Systems}, vol. 104, pp. 1--13,
  2018.

\bibitem{MarineSS-1}
V.~{Badrinarayanan}, A.~{Kendall}, and R.~{Cipolla}, ``Segnet: A deep
  convolutional encoder-decoder architecture for image segmentation,''
  \emph{IEEE Transactions on Pattern Analysis and Machine Intelligence},
  vol.~39, no.~12, pp. 2481--2495, 2017.

\bibitem{MarineSS-2}
B.~{Bovcon}, R.~{Mandeljc}, J.~{Perš}, and M.~{Kristan}, ``Improving
  vision-based obstacle detection on usv using inertial sensor,'' in
  \emph{Proceedings of the 10th International Symposium on Image and Signal
  Processing and Analysis}, 2017, pp. 1--6.

\bibitem{MarineSS-3}
B.~Bovcon, R.~Mandeljc, J.~Pers, and M.~Kristan, ``Stereo obstacle detection
  for unmanned surface vehicles by imu-assisted semantic segmentation,''
  \emph{Robotics and Autonomous Systems}, vol. 104, 02 2018.

\bibitem{MarineSS-4}
B.~Bovcon and M.~Kristan, ``Obstacle detection for usvs by joint stereo-view
  semantic segmentation,'' 10 2018, pp. 5807--5812.

\bibitem{MarineSS-7}
L.-C. Chen, Y.~Zhu, G.~Papandreou, F.~Schroff, and H.~Adam,
  \emph{Encoder-Decoder with Atrous Separable Convolution for Semantic Image
  Segmentation: 15th European Conference, Munich, Germany, September 8–14,
  2018, Proceedings, Part VII}, 09 2018, pp. 833--851.

\bibitem{MarineSS-12}
C.~Yu, J.~Wang, C.~Peng, C.~Gao, G.~Yu, and N.~Sang, \emph{BiSeNet: Bilateral
  Segmentation Network for Real-Time Semantic Segmentation: 15th European
  Conference, Munich, Germany, September 8-14, 2018, Proceedings, Part XIII},
  09 2018, pp. 334--349.

\bibitem{Taipalmaa-ICIP}
J.~Taipalmaa, N.~Passalis, and J.~Raitoharju,
  ``\BIBforeignlanguage{English}{Different color spaces in deep leraning-based
  water segmentation for autonomous marine operations},'' in
  \emph{\BIBforeignlanguage{English}{2020 IEEE 27th Conference on Image
  Processing (ICIP)}}, ser. IEEE International Conference on Image
  Processing.\hskip 1em plus 0.5em minus 0.4em\relax IEEE, 10 2020.

\bibitem{zhang2017adaptation}
Y.~Zhang, P.~David, and B.~Gong, ``Curriculum domain adaptation for semantic
  segmentation of urban scenes,'' in \emph{The IEEE International Conference on
  Computer Vision (ICCV)}, Oct 2017.

\bibitem{DetectionSurvey}
L.~Jiao, F.~Zhang, F.~Liu, S.~Yang, L.~Li, Z.~Feng, and R.~Qu, ``A survey of
  deep learning-based object detection,'' \emph{IEEE Access}, vol.~PP, pp.
  1--1, 09 2019.

\bibitem{real-time}
``{Real-time, cloud-based object detection for unmanned aerial
  vehicles}.''\hskip 1em plus 0.5em minus 0.4em\relax Institute of Electrical
  and Electronics Engineers Inc., may 2017, pp. 36--43.

\bibitem{slimyolov3}
P.~Zhang, Y.~Zhong, and X.~Li, ``{SlimYOLOv3: Narrower, faster and better for
  real-time UAV applications},'' \emph{Proceedings - 2019 International
  Conference on Computer Vision Workshop, ICCVW 2019}, pp. 37--45, 2019.

\bibitem{skynet}
X.~Zhang, C.~Hao, H.~Lu, J.~Li, Y.~Li, Y.~Fan, K.~Rupnow, J.~Xiong, T.~Huang,
  H.~Shi, W.-m. Hwu, and D.~Chen, ``{SkyNet: A Champion Model for DAC-SDC on
  Low Power Object Detection},'' vol.~6, 2019.

\bibitem{dronet}
C.~Kyrkou, G.~Plastiras, T.~Theocharides, S.~I. Venieris, and C.~S. Bouganis,
  ``{DroNet: Efficient convolutional neural network detector for real-time UAV
  applications},'' \emph{Proceedings of the 2018 Design, Automation and Test in
  Europe Conference and Exhibition, DATE 2018}, vol. 2018-Janua, pp. 967--972,
  2018.

\bibitem{Vaddi2019}
S.~Vaddi, C.~Kumar, and A.~Jannesari, ``{Efficient Object Detection Model for
  Real-Time UAV Applications},'' 2019.

\bibitem{tsaiSpatialSearch2019}
Y.~{Tsai}, B.~{Lu}, and K.~{Tseng}, ``Spatial search via adaptive submodularity
  and deep learning,'' in \emph{2019 IEEE International Symposium on Safety,
  Security, and Rescue Robotics (SSRR)}, 2019, pp. 112--113.

\bibitem{girshick14CVPR}
R.~Girshick, J.~Donahue, T.~Darrell, and J.~Malik, ``Rich feature hierarchies
  for accurate object detection and semantic segmentation,'' in \emph{Computer
  Vision and Pattern Recognition}, 2014.

\bibitem{girshickICCV15fastrcnn}
R.~Girshick, ``Fast r-cnn,'' in \emph{International Conference on Computer
  Vision ({ICCV})}, 2015.

\bibitem{tung2019distillation}
F.~Tung and G.~Mori, ``Similarity-preserving knowledge distillation,'' in
  \emph{The IEEE International Conference on Computer Vision (ICCV)}, October
  2019.

\bibitem{kiranyaz2019ONN}
S.~Kiranyaz, T.~Ince, A.~Iosifidis, and M.~Gabbouj, ``Operational neural
  networks,'' \emph{Neural Computing and Applications}, vol.~32, 2020.

\bibitem{khan2019paradox}
M.~N. Khan and S.~Anwar, ``Paradox elimination in dempster--shafer combination
  rule with novel entropy function: Application in decision-level multi-sensor
  fusion,'' \emph{Sensors}, vol.~19, no.~21, p. 4810, 2019.

\bibitem{lahat2015multimodal}
D.~Lahat, T.~Adali, and C.~Jutten, ``Multimodal data fusion: an overview of
  methods, challenges, and prospects,'' \emph{Proceedings of the IEEE}, vol.
  103, no.~9, pp. 1449--1477, 2015.

\bibitem{meng2020fusion}
T.~Meng, X.~Jing, Z.~Yan, and W.~Pedrycz, ``A survey on machine learning for
  data fusion,'' \emph{Information Fusion}, vol.~57, pp. 115 -- 129, 2020.

\bibitem{baltrusaitis2019multimodal}
T.~{Baltrušaitis}, C.~{Ahuja}, and L.~{Morency}, ``Multimodal machine
  learning: A survey and taxonomy,'' \emph{IEEE Transactions on Pattern
  Analysis and Machine Intelligence}, vol.~41, no.~2, pp. 423--443, 2019.

\bibitem{liu2020urban}
J.~Liu, T.~Li, P.~Xie, S.~Du, F.~Teng, and X.~Yang, ``Urban big data fusion
  based on deep learning: An overview,'' \emph{Information Fusion}, vol.~53,
  pp. 123 -- 133, 2020.

\bibitem{chen2020rgbd}
H.~Chen, Y.~Li, and D.~Su, ``Multi-modal fusion network with multi-scale
  multi-path and cross-modal interactions for rgb-d salient object detection,''
  \emph{Pattern Recognition}, vol.~86, pp. 376 -- 385, 2019.

\bibitem{wang2020environmental}
D.~Wang, W.~Li, X.~Liu, N.~Li, and C.~Zhang, ``{UAV environmental perception
  and autonomous obstacle avoidance : A deep learning and depth camera combined
  solution},'' \emph{Computers and Electronics in Agriculture}, vol. 175, no.
  February, p. 105523, 2020.

\bibitem{wang2018slam}
H.~Wang, C.~Zhang, Y.~Song, and B.~Pang, ``Information-fusion methods based
  simultaneous localization and mapping for robot adapting to search and rescue
  postdisaster environments,'' \emph{Journal of Robotics}, 2018.

\bibitem{katsamenis2020manoverboard}
I.~Katsamenis, E.~Protopapadakis, A.~Voulodimos, D.~Dres, and D.~Drakoulis,
  ``Man overboard event detection from rgb and thermal imagery: Possibilities
  and limitations,'' in \emph{Proceedings of the 13th ACM International
  Conference on PErvasive Technologies Related to Assistive
  Environments}.\hskip 1em plus 0.5em minus 0.4em\relax New York, NY, USA:
  Association for Computing Machinery, 2020.

\bibitem{herrmann2018thermal}
C.~Herrmann, M.~Ruf, and J.~Beyerer, ``{CNN-based thermal infrared person
  detection by domain adaptation},'' in \emph{Autonomous Systems: Sensors,
  Vehicles, Security, and the Internet of Everything}, M.~C. Dudzik and J.~C.
  Ricklin, Eds., vol. 10643, International Society for Optics and
  Photonics.\hskip 1em plus 0.5em minus 0.4em\relax SPIE, 2018, pp. 38 -- 43.

\bibitem{khan2018cooperative}
A.~{Khan}, B.~{Rinner}, and A.~{Cavallaro}, ``Cooperative robots to observe
  moving targets: Review,'' \emph{IEEE Transactions on Cybernetics}, vol.~48,
  no.~1, pp. 187--198, 2018.

\bibitem{silva2019cooperative}
``{Cooperative unmanned aerial vehicles with privacy preserving deep vision for
  real-time object identification and tracking},'' \emph{Journal of Parallel
  and Distributed Computing}, vol. 131, pp. 147 -- 160, 2019.

\bibitem{zadorozhny2013collective}
V.~{Zadorozhny} and M.~{Lewis}, ``Information fusion based on collective
  intelligence for multi-robot search and rescue missions,'' in
  \emph{International Conference on Mobile Data Management}, vol.~1, 2013, pp.
  275--278.

\bibitem{bajcsy1988active}
R.~Bajcsy, ``Active perception,'' \emph{Proceedings of the IEEE}, vol.~76,
  no.~8, pp. 966--1005, 1988.

\bibitem{gallos2019active}
D.~{Gallos} and F.~{Ferrie}, ``Active vision in the era of convolutional neural
  networks,'' in \emph{2019 16th Conference on Computer and Robot Vision
  (CRV)}, 2019, pp. 81--88.

\bibitem{bajcsy2018active}
R.~Bajcsy, Y.~Aloimonos, and J.~Tsotsos, ``Revisiting active perception,''
  \emph{Autonomous Robots}, vol.~42, pp. 177–--196, 2018.

\bibitem{falanga2017gaps}
D.~{Falanga}, E.~{Mueggler}, M.~{Faessler}, and D.~{Scaramuzza}, ``Aggressive
  quadrotor flight through narrow gaps with onboard sensing and computing using
  active vision,'' in \emph{2017 IEEE International Conference on Robotics and
  Automation (ICRA)}, 2017, pp. 5774--5781.

\bibitem{chessa2014obstacle}
M.~{Chessa}, S.~{Murgia}, L.~{Nardelli}, S.~P. {Sabatini}, and F.~{Solari},
  ``Bio-inspired active vision for obstacle avoidance,'' in \emph{2014
  International Conference on Computer Graphics Theory and Applications
  (GRAPP)}, 2014, pp. 1--8.

\bibitem{sandino2020target}
J.~Sandino, F.~Vanegas, F.~Gonz{\'a}lez, and F.~Maire, ``Autonomous uav
  navigation for active perception of targets in uncertain and cluttered
  environments,'' in \emph{2020 IEEE Aerospace Conference}, 2020.

\bibitem{zhong2018tracking}
F.~Zhong, P.~Sun, W.~Luo, T.~Yan, and Y.~Wang, ``{AD}-{VAT}: An asymmetric
  dueling mechanism for learning visual active tracking,'' in
  \emph{International Conference on Learning Representations}, 2019.

\bibitem{ammirato2017data}
P.~{Ammirato}, P.~{Poirson}, E.~{Park}, J.~{Košecká}, and A.~C. {Berg}, ``A
  dataset for developing and benchmarking active vision,'' in \emph{2017 IEEE
  International Conference on Robotics and Automation (ICRA)}, 2017, pp.
  1378--1385.

\bibitem{tzimas2020simulator}
A.~{Tzimas}, N.~{Passalis}, and A.~{Tefas}, ``Leveraging deep reinforcement
  learning for active shooting under open-world setting,'' in \emph{2020 IEEE
  International Conference on Multimedia and Expo}, 2020, pp. 1--6.

\bibitem{sadeghi2018sim2real}
F.~Sadeghi, A.~Toshev, E.~Jang, and S.~Levine, ``Sim2real viewpoint invariant
  visual servoing by recurrent control,'' in \emph{The IEEE Conference on
  Computer Vision and Pattern Recognition (CVPR)}, June 2018.

\bibitem{calli2018realtraining}
B.~{Calli}, W.~{Caarls}, M.~{Wisse}, and P.~P. {Jonker}, ``Active vision via
  extremum seeking for robots in unstructured environments: Applications in
  object recognition and manipulation,'' \emph{IEEE Transactions on Automation
  Science and Engineering}, vol.~15, no.~4, pp. 1810--1822, 2018.

\bibitem{mathe2016detection}
S.~Mathe, A.~Pirinen, and C.~Sminchisescu, ``Reinforcement learning for visual
  object detection,'' in \emph{The IEEE Conference on Computer Vision and
  Pattern Recognition (CVPR)}, June 2016.

\bibitem{andersson2017pid}
P.~D. Olov~Andersson, Mariusz~Wzorek, ``Deep learning quadcopter control via
  risk-aware active learning,'' in \emph{Proceedings of the Thirty-First AAAI
  Conference on Artificial Intelligence}, 2016.

\bibitem{zhou2011multirobot}
K.~Zhou and S.~I. Roumeliotis, ``Multirobot active target tracking with
  combinations of relative observations,'' \emph{IEEE Transactions on
  Robotics}, vol.~27, no.~4, pp. 678--695, 2011.

\bibitem{ahmad2013perception}
A.~Ahmad, T.~Nascimento, A.~G. Concei{\c{c}}ao, A.~P. Moreira, and P.~Lima,
  ``Perception-driven multi-robot formation control,'' in \emph{2013 IEEE
  International Conference on Robotics and Automation}.\hskip 1em plus 0.5em
  minus 0.4em\relax IEEE, 2013, pp. 1851--1856.

\bibitem{tallamraju2019active}
R.~Tallamraju, E.~Price, R.~Ludwig, K.~Karlapalem, H.~H. B{\"u}lthoff, M.~J.
  Black, and A.~Ahmad, ``Active perception based formation control for multiple
  aerial vehicles,'' \emph{IEEE Robotics and Automation Letters}, vol.~4,
  no.~4, pp. 4491--4498, 2019.

\bibitem{kiciroglu2020activemocap}
S.~Kiciroglu, H.~Rhodin, S.~N. Sinha, M.~Salzmann, and P.~Fua, ``Activemocap:
  Optimized viewpoint selection for active human motion capture,'' in
  \emph{Proceedings of the IEEE/CVF Conference on Computer Vision and Pattern
  Recognition}, 2020, pp. 103--112.

\bibitem{chen2015adaptive}
X.~Chen and Y.~Jia, ``Adaptive leader-follower formation control of
  non-holonomic mobile robots using active vision,'' \emph{IET Control Theory
  \& Applications}, vol.~9, no.~8, pp. 1302--1311, 2015.

\bibitem{zhang2019optimized}
L.~Zhang, S.~Chowdhury, R.~Siegwart, and J.~J. Chung, ``Optimized motion
  strategy for active target localization of mobile robots with time-varying
  connectivity,'' in \emph{2019 International Symposium on Multi-Robot and
  Multi-Agent Systems (MRS)}.\hskip 1em plus 0.5em minus 0.4em\relax IEEE,
  2019, pp. 185--187.

\bibitem{morbidi2012active}
F.~Morbidi and G.~L. Mariottini, ``Active target tracking and cooperative
  localization for teams of aerial vehicles,'' \emph{IEEE transactions on
  control systems technology}, vol.~21, no.~5, pp. 1694--1707, 2012.

\bibitem{gurcuoglu2013hierarchical}
U.~G{\"u}rc{\"u}oglu, G.~A. Puerto-Souza, F.~Morbidi, and G.~L. Mariottini,
  ``Hierarchical control of a team of quadrotors for cooperative active target
  tracking,'' in \emph{2013 IEEE/RSJ International Conference on Intelligent
  Robots and Systems}.\hskip 1em plus 0.5em minus 0.4em\relax IEEE, 2013, pp.
  5730--5735.

\bibitem{vander2015active}
J.~Vander~Hook, ``Active target localization and tracking with application to
  robotic environmental monitoring,'' 2015.

\bibitem{acevedo2020dynamic}
J.~J. Acevedo, J.~Messias, J.~Capit{\'a}n, R.~Ventura, L.~Merino, and P.~U.
  Lima, ``A dynamic weighted area assignment based on a particle filter for
  active cooperative perception,'' \emph{IEEE Robotics and Automation Letters},
  vol.~5, no.~2, pp. 736--743, 2020.

\bibitem{tokekar2012coverage}
P.~Tokekar, E.~Branson, J.~Vander~Hook, and V.~Isler, ``Coverage and active
  localization for monitoring invasive fish with an autonomous boat,''
  \emph{IEEE Robotics and Automation Magazine}, 2012.

\bibitem{vander2013local}
J.~Vander~Hook, P.~Tokekar, E.~Branson, P.~G. Bajer, P.~W. Sorensen, and
  V.~Isler, ``Local-search strategy for active localization of multiple
  invasive fish,'' in \emph{Experimental Robotics}.\hskip 1em plus 0.5em minus
  0.4em\relax Springer, 2013, pp. 859--873.

\bibitem{atanasov2015decentralized}
N.~Atanasov, J.~Le~Ny, K.~Daniilidis, and G.~J. Pappas, ``Decentralized active
  information acquisition: Theory and application to multi-robot slam,'' in
  \emph{2015 IEEE International Conference on Robotics and Automation
  (ICRA)}.\hskip 1em plus 0.5em minus 0.4em\relax IEEE, 2015, pp. 4775--4782.

\bibitem{schlotfeldt2019maximum}
B.~Schlotfeldt, N.~Atanasov, and G.~J. Pappas, ``Maximum information bounds for
  planning active sensing trajectories,'' in \emph{2019 IEEE/RSJ International
  Conference on Intelligent Robots and Systems (IROS)}.\hskip 1em plus 0.5em
  minus 0.4em\relax IEEE, 2019, pp. 4913--4920.

\bibitem{escusol2017autonomous}
J.~V. Escusol, J.~Aaltonen, and K.~T. Koskinen, ``Autonomous and collaborative
  offshore robotics,'' in \emph{the 2 nd Annual SMACC Research Seminar 2017},
  2017.

\bibitem{arndt2019meta}
K.~Arndt, M.~Hazara, A.~Ghadirzadeh, and V.~Kyrki, ``Meta reinforcement
  learning for sim-to-real domain adaptation,'' \emph{arXiv preprint
  arXiv:1909.12906}, 2019.

\bibitem{sampedro2019fully}
C.~Sampedro, A.~Rodriguez-Ramos, H.~Bavle, A.~Carrio, P.~de~la Puente, and
  P.~Campoy, ``A fully-autonomous aerial robot for search and rescue
  applications in indoor environments using learning-based techniques,''
  \emph{Journal of Intelligent \& Robotic Systems}, vol.~95, no.~2, pp.
  601--627, 2019.

\bibitem{niroui2019deep}
F.~Niroui, K.~Zhang, Z.~Kashino, and G.~Nejat, ``Deep reinforcement learning
  robot for search and rescue applications: Exploration in unknown cluttered
  environments,'' \emph{IEEE Robotics and Automation Letters}, vol.~4, no.~2,
  pp. 610--617, 2019.

\bibitem{li2020deep}
H.~Li, Q.~Zhang, and D.~Zhao, ``Deep reinforcement learning-based automatic
  exploration for navigation in unknown environment,'' \emph{IEEE transactions
  on neural networks and learning systems}, 2020.

\bibitem{maruyama2016exploring}
Y.~Maruyama, S.~Kato, and T.~Azumi, ``Exploring the performance of ros2,'' in
  \emph{Proceedings of the 13th International Conference on Embedded Software},
  2016, pp. 1--10.

\end{thebibliography}
